%% file: neurips_2026.tex
\documentclass{article}

\PassOptionsToPackage{numbers, compress}{natbib}
 \usepackage[eandd,final]{neurips_2026}

\usepackage[utf8]{inputenc} 
\usepackage[T1]{fontenc}    
\usepackage{hyperref}       
\usepackage{url}            
\usepackage{booktabs}       
\usepackage{longtable}
\usepackage{amsfonts}       
\usepackage{nicefrac}       
\usepackage{microtype}      
\usepackage{xcolor}         
\usepackage{amsmath}
\usepackage{spverbatim}
\usepackage{graphicx}
\usepackage{wrapfig}
\usepackage{subcaption} 
\usepackage{siunitx}    
\usepackage{float}
\usepackage{enumitem,amssymb}
\newlist{todolist}{itemize}{2}
\setlist[todolist]{label=$\square$}

\definecolor{indiagreen}{rgb}{0.0, 0.5, 0.0}

\usepackage{multirow}
\usepackage[table]{xcolor}
\usepackage{fontawesome}
\usepackage{caption}

\usepackage{booktabs}
\usepackage{tabularx}
\usepackage{array}
\usepackage{makecell}
\usepackage{pifont}
\usepackage{bm}

\newcolumntype{L}[1]{>{\raggedright\arraybackslash}p{#1\textwidth}}
\newcolumntype{P}[1]{>{\centering\arraybackslash}p{#1\textwidth}}
\newcolumntype{Y}{>{\centering\arraybackslash}X}
\newcolumntype{L}{>{\raggedright\arraybackslash}X}
\colorlet{shadecolor}{gray!30}

\usepackage{enumitem}

\usepackage[dvipsnames]{xcolor}   

\usepackage{tikz}
\newcommand{\legenddot}[1]{%
\tikz[baseline=-0.5ex]\draw[#1,fill=#1] (0,0.025) circle (3pt);%
}
\newcommand{\legendddot}[1]{%
\tikz[baseline=-0.5ex]\filldraw[color=black,fill=white,thin] (0,0.025) circle (3pt);}
 \newcommand{\imgset}[2]{%
   \begin{minipage}[c]{0.50\linewidth}%
     \begin{minipage}[c]{0.56\linewidth}%
       \includegraphics[width=\linewidth]{#1}%
     \end{minipage}%
     \hspace{5pt}%
     \begin{minipage}[c]{0.40\linewidth}%
       \renewcommand{\arraystretch}{0.75}%
       \begin{tabular}{@{}cl@{}}%
         \toprule
         & Object(s) \\
         \midrule
         #2
         \bottomrule
       \end{tabular}%
     \end{minipage}%
   \end{minipage}%
 }

\title{VLA-REPLICA: A Low-Cost, Reproducible Benchmark for Real-World Evaluation of Vision-Language-Action Models}

\author{
Alex S. Huang$^{1*}$ \quad
Jiahui Zhang$^{1*}$ \quad
Shiqing Tang$^{2}$ \quad
Yu Xiang$^{1}$ \\
\\
$^{1}$Intelligent Robotics and Vision Lab, University of Texas at Dallas \\
$^{2}$Allen High School \\ \\
\texttt{\{alex.huang, jiahui.zhang, yu.xiang\}@utdallas.edu} \\
\texttt{shiqing.tang@student.allenisd.org} \\
$^{*}$Equal contribution
}

\begin{document}

\maketitle

\begin{abstract}
Vision-Language-Action (VLA) models have shown strong promise for general-purpose robotic manipulation, but their real-world evaluation remains limited by a lack of accessible, reproducible, and consistent benchmarks. Simulation benchmarks fail to capture real-world complexity, while existing real-world benchmarks often require expensive hardware, centralized evaluation, or are limited in task diversity. We introduce \emph{VLA-REPLICA}, a low-cost, easily reproducible real-world benchmark for evaluating VLA models. Built from off-the-shelf components, our system can be quickly assembled and replicated across laboratories, providing a consistent environment for policy evaluation anywhere in the world. \emph{VLA-REPLICA} includes a diverse suite of manipulation tasks and a small-scale demonstration dataset for target-domain adaptation, with real-world evaluation protocols for both in-distribution and out-of-distribution settings. Experiments with imitation learning and state-of-the-art VLA models reveal model strengths and limitations, while consistent results across independently constructed setups demonstrate the reproducibility of our benchmark.\footnote{Data, code, and videos for the project are available at \url{https://irvlutd.github.io/VLAReplica/}.}
\end{abstract}

\input{sections/intro}

\input{sections/related}

\input{sections/benchmark}

\input{sections/experiments}
\input{sections/conclusion}

{
    \bibliographystyle{abbrvnat}
    \bibliography{refs}
}

\newpage

\appendix

\input{sections/appendix}


\newpage
\input{checklist.tex}

\end{document}

%% file: sections/intro.tex
\section{Introduction}

Vision-Language-Action (VLA) models~\cite{kim2024openvla,black2024pi0visionlanguageactionflowmodel,intelligence2025pi05} have recently emerged as a promising paradigm for building general-purpose robotic systems that can interpret natural language instructions and execute corresponding manipulation behaviors. By leveraging large-scale multimodal data~\cite{o2024open,khazatsky2024droid}, these models demonstrate impressive generalization across objects, scenes, and tasks, bringing robotics closer to the goal of flexible, human-level interaction. However, despite their success in controlled settings, a fundamental challenge remains: \emph{how to reliably evaluate VLA models in the real world}.

Existing evaluation protocols fall short in addressing this challenge. Simulation-based benchmarks such as Meta-World~\cite{mclean2025metaworld} and LIBERO~\cite{liu2023libero} provide scalable and standardized environments, but they suffer from the well-known sim-to-real gap, often leading to overly optimistic estimates of real-world performance. In contrast, real-world benchmarks offer more faithful evaluation but introduce new limitations. Some benchmarks, such as REPLAB~\cite{yang2019replab} and SceneReplica~\cite{khargonkar2024scenereplica}, focus primarily on object grasping, limiting task diversity. Others, such as FurnitureBench~\cite{heo2025furniturebench} and ManipulationNet~\cite{chen2026manipulationnet}, require expensive hardware setups, carefully engineered environments, and substantial human effort to reproduce. Additionally, systems like RoboArena~\cite{atreya2025roboarena} rely on centralized or remote evaluation, where users must submit policies for external execution, restricting accessibility, slowing iteration, and reducing transparency. As a result, there is currently \emph{no widely adopted benchmark that is both real-world grounded and easily reproducible across independent laboratories}.

\begin{figure}[t]
  \centering
  \includegraphics[width=\linewidth]{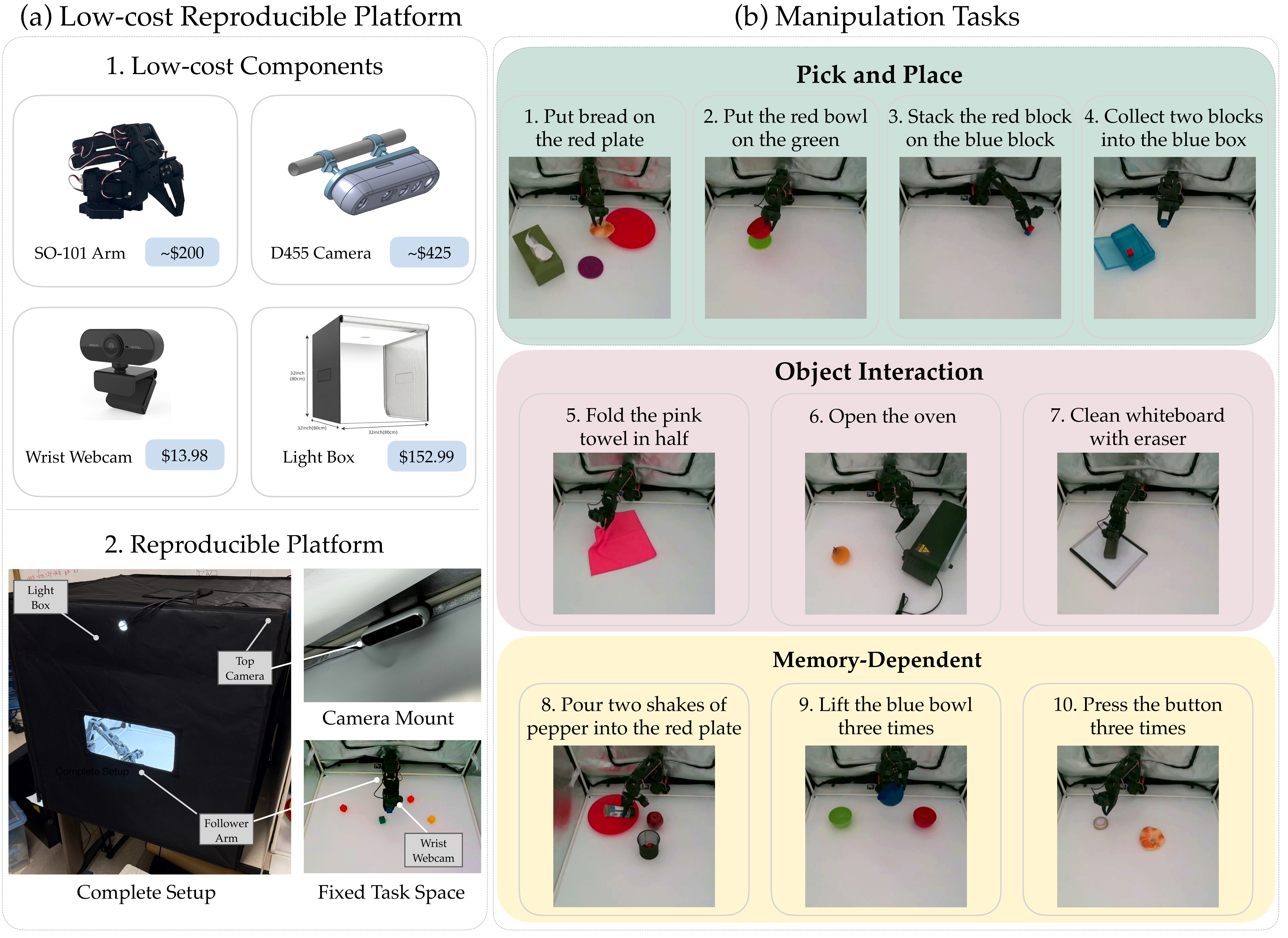}
  \caption{
  Overview of the VLA-Replica benchmark. (a)1. Hardware components. 
  (a)2. Our assembled platform with the leader arm, the follower arm, the light box, the cameras, and the manipulation workspace. 
  (b) 10 manipulation tasks in the benchmark.                           }
  \label{fig:overview}
  \vspace{-4mm}
\end{figure}

To address these limitations, we introduce \textbf{VLA-REPLICA}, a low-cost, reproducible real-world benchmark for evaluating VLA models. Our key design principle is to make real-world evaluation \emph{accessible, standardized, and locally executable}. Specifically, VLA-REPLICA is built on an affordable robotic platform using off-the-shelf components, including a low-cost SO-101 manipulator, RGB-D cameras, and a controlled light-box environment. The hardware setup is illustrated in Fig.~\ref{fig:overview}(a). The full setup can be assembled in under an hour by non-expert users, and detailed instructions ensure consistent replication across different sites.

Beyond hardware accessibility, VLA-REPLICA provides a structured evaluation protocol tailored to modern VLA models. The benchmark includes a suite of real-world manipulation tasks with diverse interaction patterns (e.g., pick-and-place, tool use, and object manipulation), along with a curated dataset of human demonstrations for target-domain adaptation. Fig.~\ref{fig:overview}(b) shows the 10 manipulation tasks in our benchmark. Crucially, the evaluation is designed to measure both \emph{in-distribution performance}, i.e. how well models learn from limited real-world data, and \emph{out-of-distribution generalization}, including variations in object properties and task requirements. This setup reflects practical deployment scenarios, where pretrained VLA models must adapt to new environments with minimal data.

A central contribution of our work is \emph{reproducibility at scale}. We introduce a standardized environment construction pipeline, including precise workspace configuration, camera placement, background lighting, and object placement protocols, enabling consistent evaluation across independently built setups. This design allows different users to recreate nearly identical physical environments, making it possible to compare results across laboratories without centralized infrastructure.


We validate VLA-REPLICA by benchmarking both imitation learning methods~\cite{zhao2023learning,chi2025diffusion} and state-of-the-art VLA models~\cite{black2024pi0visionlanguageactionflowmodel,intelligence2025pi05,shukor2025smolvla} under a unified training and evaluation protocol. Our experiments reveal key insights into the performance of current VLA systems, particularly their ability to adapt to new environments and generalize beyond training distributions. We further demonstrate that evaluation results are consistent across independently constructed setups, highlighting the robustness and reproducibility of our benchmark.

In summary, this paper makes the following contributions:
\begin{itemize}
    \item We introduce \emph{VLA-REPLICA}, a low-cost ($\sim$\$1050 USD), fully reproducible real-world benchmark for evaluating VLA models.
    \item We design a \emph{standardized hardware and environment setup} that can be replicated across laboratories with minimal effort and cost.
    \item We propose a \emph{comprehensive evaluation protocol} covering both in-distribution learning and out-of-distribution generalization.
    \item We provide \emph{extensive empirical evaluation} of state-of-the-art VLA models, offering insights into their real-world performance and limitations.
\end{itemize}

%% file: sections/related.tex
\vspace{-2mm}
\section{Related Work}
\vspace{-2mm}

\begin{table}[t]
    \centering
    \caption{Comparison of robot manipulation benchmarks across environment, reproducibility, hardware cost, task diversity, dataset availability, and evaluation protocol.}
    \tiny
    \setlength{\tabcolsep}{3pt}
    \renewcommand{\arraystretch}{1.05}
    \rowcolors{3}{gray!15}{white}

\begin{tabularx}{\textwidth}{@{}l Y Y Y Y Y Y Y@{}}
    \toprule
    \rowcolor{gray!35}
    \textbf{Benchmark} &
    \textbf{Year} &
    \textbf{Environment} &
    \textbf{\makecell{Repro-\\ducibility}} &
    \textbf{\makecell{Hardware\\Cost}} &
    \textbf{\makecell{Task\\Diversity}} &
    \textbf{\makecell{Expert\\Dataset}} &
    \textbf{\makecell{Distributed\\Evaluation}} \\
    \midrule

    Meta-World~\cite{yu2020meta} & 2020 & Simulation & High & -- & High (50 tasks) & \ding{51} & \ding{51} \\
    
    LIBERO~\cite{liu2023libero} & 2023 & Simulation & High & -- & High (100 tasks) & \ding{51} & \ding{51} \\ 

    RoboCasa~\cite{robocasa2024} & 2024 & Simulation & High & -- & High (100 tasks) &  \ding{51} & \ding{51} \\

    RoboLab~\cite{yang2026robolab} & 2026 & Simulation & High & -- & High (120 tasks) & \ding{55} & \ding{51} \\

    \midrule
    
    REPLAB~\cite{yang2019replab} & 2019 & Real & Low &  Low & Low (Grasping) & \ding{51} & \ding{51} \\
    
    RAMP~\cite{collins2023ramp} & 2023 & Real & High (April Tag) & High & Low (Assembly) & \ding{55} & \ding{51} \\
    
    SceneReplica~\cite{khargonkar2024scenereplica} & 2024 & Real & High (Overlay) & High & Low (Grasping) & \ding{55} & \ding{51} \\ 
    
    FurnitureBench~\cite{heo2025furniturebench} & 2025 & Real & High (April Tag) & High & Low (Assembly) & \ding{51} & \ding{51} \\
    
    FMB~\cite{luo2025fmb} & 2025 & Real & Low & High & Low (Assembly) & \ding{51} & \ding{51} \\
    
    AutoEval~\cite{zhou2025autoeval} & 2025 & Real & Low & None (Service) & Low (4 tasks) & \ding{51} & \ding{55} \\
    
    RoboArena~\cite{atreya2025roboarena} & 2025 & Real & Low & High & High (User-defined) & \ding{55} & \ding{51} \\

    RoboChallenge~\cite{yakefu2025robochallenge} & 2025 & Real & High (Overlay) & None (Service) & High (30 tasks) &  \ding{51} & \ding{55} \\

    \hline
    
    \textbf{VLA-Replica (Ours)} & 2026 & Real & High (Overlay+Lighting) & Low ($\sim$\$1050 USD) & High (10 Tasks) & \ding{51} & \ding{51} \\
    \bottomrule
\end{tabularx}

    \label{tab:benchmark_properties}
    \vspace{-5mm}
\end{table}

\textbf{Vision-Language-Action Models} 
Recent years have seen the emergence of VLA models~\citep{brohan2022rt1, brohan2023rt2visionlanguageactionmodelstransfer,kim2024openvla,black2024pi0visionlanguageactionflowmodel, intelligence2025pi05,gr00tn1_2025, intelligence2025pi06vla}, which are general-purpose robotic manipulation policies trained on diverse robot datasets for language-conditioned tasks. While these models show promising generalization across objects, scenes, and tasks, they often struggle when deployed in new environments or on new robot embodiments. Prior work therefore commonly fine-tunes pretrained VLA policies using a small amount of target-domain data, enabling adaptation to the target environment, embodiment, and tasks.

\textbf{Simulation Benchmarks for Robot Manipulation} 
Robotic evaluation benchmarks have been widely used to measure the performance and generalization of robot policies. Table~\ref{tab:benchmark_properties} provides a comparison of representative robot manipulation benchmarks. A large body of benchmarks are built in simulation~\citep{mees2022calvin, liu2023libero, robocasa2024, li24simpler, geng2025roboverse, mclean2025metaworld,yang2026robolab, robocasa365}. These benchmarks provide scalable and repeatable environments for policy evaluations. However, policies that perform well in simulation may not transfer directly to the real world due to the physical gap between simulation and real world.

\textbf{Real-World Benchmarks for Robot Manipulation} Another line of work evaluates robot policies in the real world~\citep{yang2019replab, khargonkar2024scenereplica, yakefu2025robochallenge, heo2025furniturebench, atreya2025roboarena,chen2026manipulationnet}. These benchmarks provide realistic measurements of robot performance in physical environments. However, only a few benchmarks offer reproducible test scenes. RAMP~\cite{collins2023ramp} and FurnitureBench~\cite{heo2025furniturebench} use April Tags on objects. SceneReplica~\cite{khargonkar2024scenereplica} and RobotChallenge~\cite{yakefu2025robochallenge} provide reference images for test scenes, where users can check the overlay of real objects with a reference image to place objects. In addition, some recent real-robot evaluation benchmarks rely on remote or centralized evaluation~\citep{atreya2025roboarena, yakefu2025robochallenge}, where users submit policy checkpoints or inference APIs and wait for external evaluators to run the experiments. While this improves standardization, it limits immediate, on-site, and reproducible evaluation by individual labs.

As shown in Table~\ref{tab:benchmark_properties}, VLA-REPLICA is a low-cost, reproducible real-world benchmark with diverse manipulation tasks. Built on the SO-101 and commodity objects, it enables reproducible test scenes via reference images and controlled lighting, supports distributed evaluation across sites, and provides in-domain demonstrations for fine-tuning pretrained VLA policies.



%% file: sections/benchmark.tex
\begin{table}
  \vspace{-1mm}
  \caption{Bill of materials for the VLA-Replica benchmark}
  \label{table:cost-table}
  \centering
  \scalebox{0.9}{
  \begin{tabular}{lccc}
    \toprule
    Item     & Unit Cost & Quantity & Description \\
    \midrule
    SO-101 Follower Arm & $\sim$\$200& 1 &  multiple vendors   \\
    Intel RealSense D455 Camera & $\sim$\$425   & 1 & multiple vendors    \\
    Vinmooog Webcam & \$13.98 & 1 & from Amazon \\
    Light Box (see Appendix~\ref{app:setup}) & \$152.99 & 1  & from Amazon \\
    Object Set (see Appendix~\ref{app:objects}) & \$215.99 & 1  & from Amazon \\
    \midrule
 \textbf{Total Cost} & \textbf{$\sim$\$1050} & &\\
    \bottomrule
  \end{tabular}
  }
  \vspace{-2mm}
\end{table}

\section{The VLA-Replica Benchmark}

\subsection{Physical Platform and Hardware Configuration}

We now describe the hardware setup of VLA-REPLICA. Detailed documentation for constructing the complete platform is available at: \url{https://irvlutd.github.io/VLAReplica}.

\textbf{Low-cost reproducible hardware design.} 
To make the benchmark reproducible, realistic, and consistent across laboratories, we design a low-cost hardware platform for less than \textbf{\$1100 USD}. As shown in Figure~\ref{fig:overview}(a), the platform consists of a 6-DoF low-cost SO-101 follower arm~\cite{so101arm}, an RGB web camera, and an Intel RealSense D455 RGB-D camera. For the workspace, we use a 32~inch $\times$ 32~inch $\times$ 32~inch photography light box to reduce environmental variation and provide consistent illumination, at around 5600K. We 3D-print a custom mount to attach the Intel RealSense D455 RGB-D camera to the rear top frame of the light box, where it serves as the top-view camera for observing the workspace. Similarly, we 3D-print a wrist camera mount~\cite{so101camera} and attach it onto the end-effector of the SO-101 follower arm. Table~\ref{table:cost-table} shows the bill of materials of our benchmark.

\begin{figure}
  \centering
  \includegraphics[width=\linewidth]{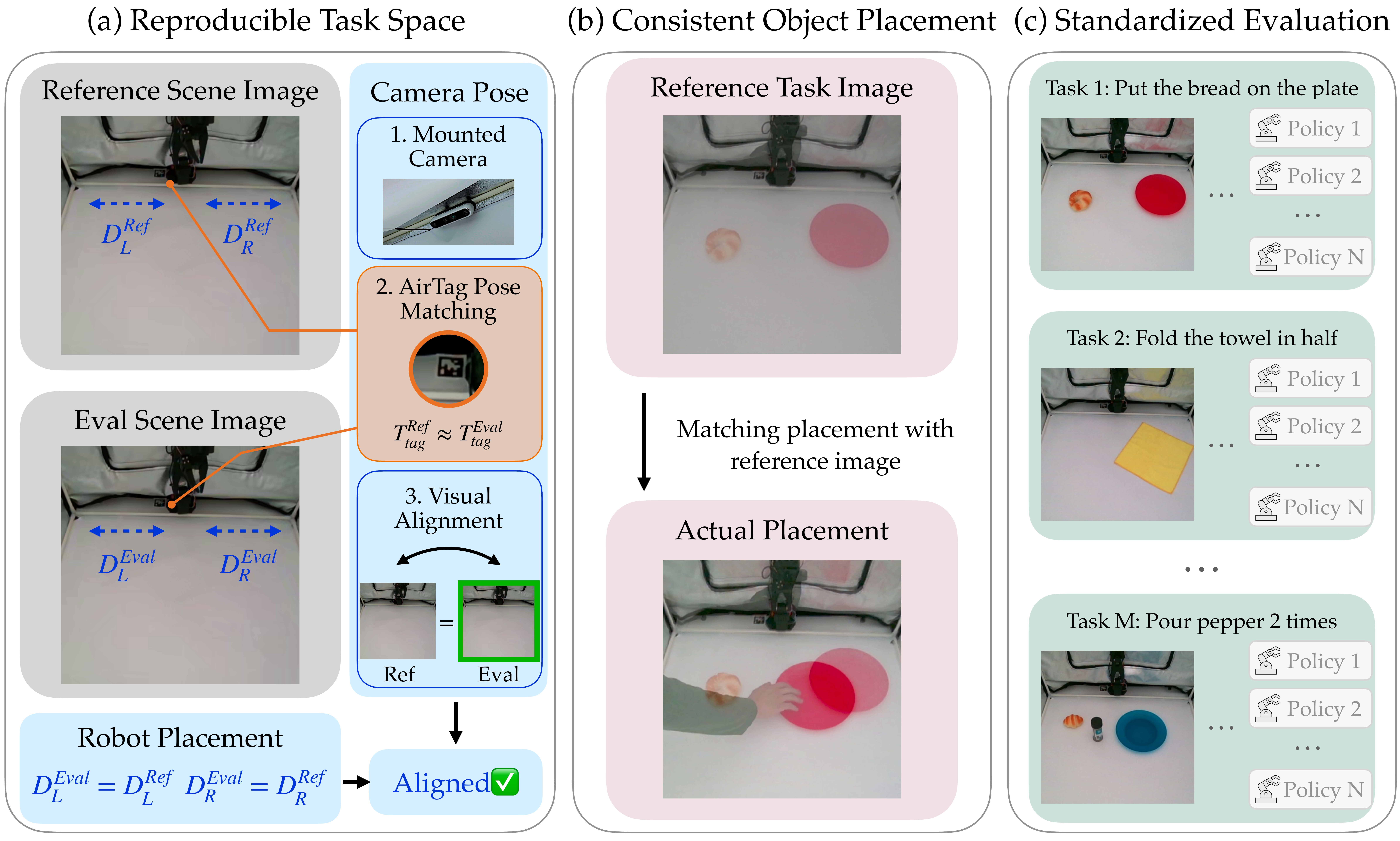}
  \caption{
VLA-REPLICA standardized method ensuring reproducibility. (a) Align the task space, robot position, and camera poses via AprilTag calibration and video overlay matching.
(b) Use task-specific reference images to ensure consistent object placement during setup.
(c) Evaluate different policies on the VLA-REPLICA task suite.                           
  }
  \label{fig:method}
  \vspace{-2mm}
\end{figure}

\textbf{Hardware configuration.}  To make both camera setups reproducible across laboratories,  we implement a camera overlay program to ensure the camera view at any laboratory matches our original setup's view (for both top and wrist cameras). Additionally, to fine-tune the camera position, we place an AprilTag at a fixed position next to the SO-101 arm and record its 6-DoF pose. Combined with the visual overlay program, this AprilTag provides an additional reference metric for aligning the camera with respect to the robot base, helping different users reproduce a consistent camera viewpoint to the workspace. Fig.~\ref{fig:method}(a) illustrates the process of reproducing the physical platform. 

We provide detailed system setup instructions in Appendix~\ref{app:setup}, enabling users to build the benchmark environment. As an initial reproducibility check, a user with no prior knowledge of the benchmark was able to build the setup within one hour.

\subsection{SO-101 Action Reproducibility}
\label{sec:reproduce}
One advantage of LeRobot's SO-101 robotic arm is its low cost and ease of self-assembly, comprised entirely of 3D-printed parts and Feetech STS3215 servo motors. However, this introduces the problem of non-standardized manufacturing, specifically, the differing initial offset of the servo motors across differently constructed SO-101 arms, which constrains each arm to its own specific action space. The LeRobot SO-101 Python library~\cite{cadene2024lerobot} contains useful tools, such as individual robot calibration, to solve this. We ensure action reproducibility by defining a \emph{universal action space} that all SO-101 arms must share.

\textbf{Arm Calibration.} Each action $\mathbf{\hat{a}}$ that controls the SO-101 arm is a 6-dimensional vector $\{a_1,\dots,a_6\}$, where $a_i$ is a raw encoder value in range $[0,4095]$ sent to servo motor $i$ on the arm. Every SO-101 arm requires calibration during setup with the \texttt{lerobot-setup-motors} package command. During calibration, the arm is extended at a standard ``L'' shape, which defines the universal ``zero position'' of all SO-101 arms. Three actions $\{\mathbf{a}^{\text{offset}} ,\mathbf{a}^{\text{min}} ,\mathbf{a}^{\text{max}}\}$  are captured during calibration: the servos' offset values at the ``zero position'', and the servos' minimum and maximum values swept throughout the joint limits, respectively.

\textbf{Action Normalization.} Let us define a $6\times6$ diagonal calibration matrix $\mathbf{D}$ of the arm so that its entries $r_i$ are the linear scaling factors calculated from $\mathbf{a}^{\text{min}}$ and $\mathbf{a}^{\text{max}}$ (every arm should have very similar physical joint limits):
\begin{equation}
\mathbf{D} =
\frac{1}{180}\begin{bmatrix}
\label{eq1}
r_1 & 0 & 0 \\
0 & \ddots  & 0 \\
 0& 0 & r_6
\end{bmatrix}~,~r_i=\frac{a_i^{\text{max}}-a_i^{\text{min}}}{2}~,~i=1,\ldots,6.
\end{equation}
Now, given an action $\mathbf{\hat{a}}$, we can define the \emph{SO-101 universal action} $\mathbf{\hat{u}}$, the joint angle (in degrees) of each servo joint as
\begin{equation}
\label{eq2}
    \hat{\mathbf{u}}=\mathbf{D}^{-1}(\mathbf{\hat{a}}- \mathbf{a}^{\text{offset}}) ~,~ \hat{\mathbf{u}}\in [-180,180],
\end{equation}
where $\mathbf{D}$ and $\mathbf{a}^{\text{offset}}$ are obtained from the arm calibration. $\mathbf{\hat{u}}$ is the action that we save during demonstration collection. Consequently, a policy is also trained to output $\mathbf{\hat{u}}$. Inversely, to convert $\mathbf{\hat{u}}$ into a specific action for the arm, we have
\begin{equation}
\label{eq3}
    \mathbf{\hat{a}} = \mathbf{D}\hat{\mathbf{u}}+\mathbf{a}^{\text{offset}}
~,~ \mathbf{\hat{a}} \in [0,4095].
\end{equation}
Therefore, as long as an SO-101 arm is calibrated, i.e., $\{\mathbf{a}^{\text{offset}} ,\mathbf{a}^{\text{min}} ,\mathbf{a}^{\text{max}}\}$ are available, we can use Eq.~\eqref{eq3} to convert the policy output $\hat{\mathbf{u}}$ to the actual action $\hat{\mathbf{a}}$ for the arm. 



\subsection{Task Suite and Evaluation Protocols}

\label{sec:task}

Our \emph{VLA-REPLICA} task suite consists of ten tasks as shown in Table~\ref{tab:task_split_with_id}. The task suite covers a diverse set of robot behaviors, including pick-and-place, pulling, wiping, pouring, other object-centric interactions, and memory capabilities. The full task list, including task success criterion, is provided in Appendix~\ref{app:task_variant_list}.

\begin{table}
    \centering
    \small
    \setlength{\tabcolsep}{5pt}
    \renewcommand{\arraystretch}{1.18}
\caption{Task definitions and examples for training/ID and OOD evaluation. ID tasks follow the same distribution as training data with varied object configurations, while OOD tasks introduce distribution shifts in object attributes (e.g., color, count) and task requirements to evaluate generalization.}
    \label{tab:task_split_with_id}
    \scalebox{0.9}{
    \begin{tabular}{c|c|p{0.34\linewidth} p{0.34\linewidth}}
        \toprule
        \textbf{Task \#} & \textbf{Task Type} & \centering\textbf{Training \& ID Eval Task Example}& \centering\arraybackslash\textbf{OOD Eval Task Example}\\
        \midrule

        1  & \multirow{4}{*}{\makecell[c]{Pick-and-\\Place}}
           & Put bread in \textbf{\textcolor{red}{red}/\textcolor{blue}{blue}} plate& Put bread on \textcolor{orange}{\textbf{yellow}} plate\\
           
        2  & & Put \textcolor{red}{\textbf{red}} bowl on \textcolor{purple}{\textbf{purple}} coaster& Put \textcolor{green}{\textbf{green}} bowl on \textcolor{orange}{\textbf{yellow}} coaster \\
        
        3  & & Stack \textcolor{blue}{\textbf{blue}} cube on \textcolor{red}{\textbf{red}} cube & Stack \textcolor{green}{\textbf{green}} cube on \textcolor{green}{\textbf{green}} cube\\
        
        4  & & Put all \textbf{2 or 3} blocks in \textcolor{blue}{\textbf{blue}} box & Put all \textbf{3 or 4} blocks in \textcolor{pink}{\textbf{pink}} box \\

        \midrule

        5  & \multirow{3}{*}{\makecell[c]{Object\\Interaction}}
           & Fold \textbf{\textcolor{pink}{pink}/\textcolor{orange}{yellow}} towel in half& Fold \textcolor{blue}{\textbf{blue}} towel in half\\
        6  & & Open oven & N/A \\
        7  & & Clean whiteboard with eraser & N/A \\

        \midrule

        8  & \multirow{3}{*}{\makecell[c]{Memory}}
           & Pour pepper \textbf{1,2 or 3} times to \textcolor{red}{\textbf{red}} plate  & Pour pepper \textbf{4 or 5} times to \textcolor{blue}{\textbf{blue}} plate \\ 
        9  & & Lift \textbf{\textcolor{green}{green}/\textcolor{red}{red}/\textcolor{blue}{blue}} bowl \textbf{1 or 3} times & Lift \textcolor{orange}{\textbf{yellow}} bowl \textbf{2, 4 or 5} times \\
        10 & & Press  button \textbf{1 or 3} times & Press button \textbf{2, 4 or 5} times \\

        \bottomrule
    \end{tabular}
    }
\end{table}

\begin{figure}
\vspace{-2mm}
  \centering
  \includegraphics[width=0.9\linewidth]{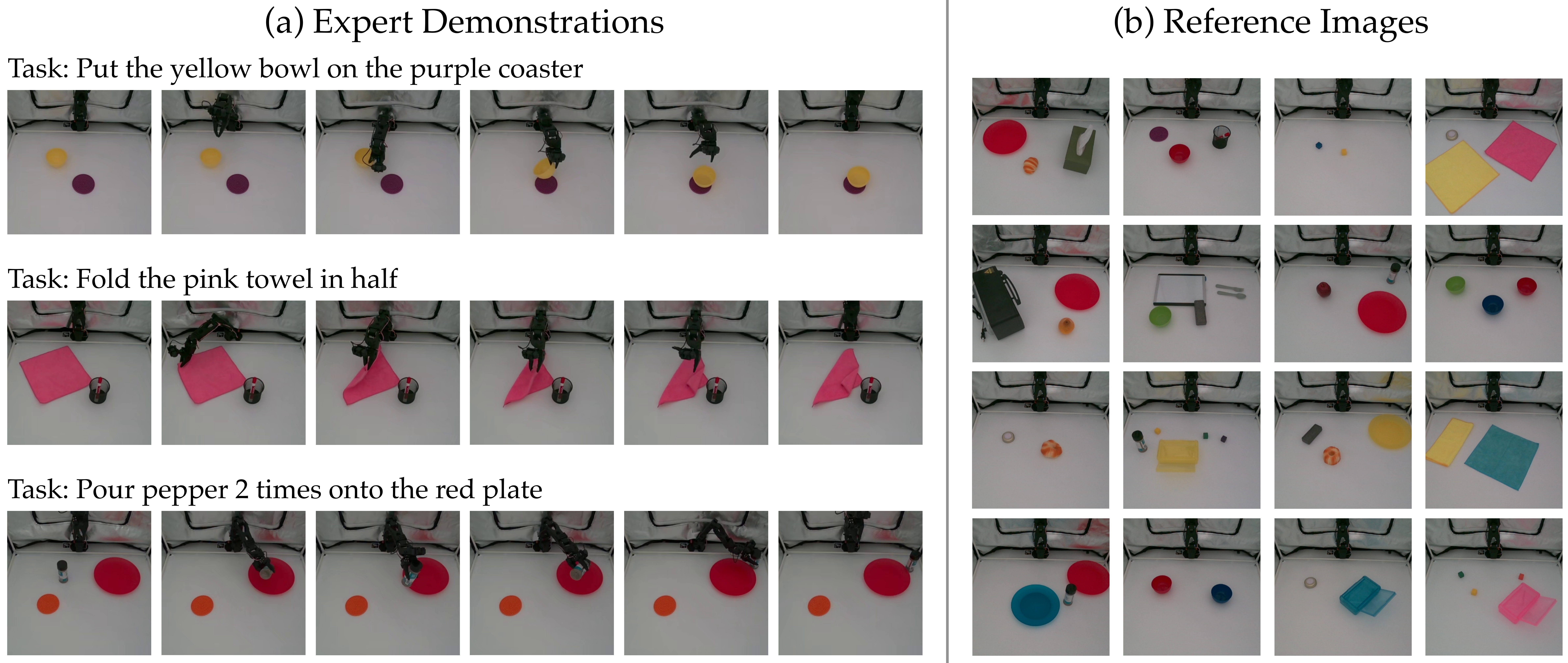}
  \caption{
(a) Examples of expert demonstrations collected in our dataset. (b) Examples of reference images for setting up test scenes.            
  }
  \label{fig:expert_flow}
  \vspace{-5mm}
\end{figure}

\textbf{Demonstration and Training Data Collection.}
We collect demonstrations for the suite of ten real-world robot manipulation tasks by using an SO-101 leader arm to control the SO-101 follower arm. For each task, we collect 50 demonstrations with different scene configurations through an expert teleoperator, resulting in 500 demonstrations total. We ensured consistent robot actions to reduce action divergence, including establishing a ``home position'' for the robot to return to at the beginning and end of an episode. Fig.~\ref{fig:expert_flow}(a) shows some examples of our collected demonstrations. 

\textbf{Evaluation Scenarios and Test Scene Design.}
A key challenge in evaluating VLA models is to disentangle \emph{adaptation} from \emph{generalization}. In practical deployments, a policy must both learn effectively from limited target-domain data and generalize to novel variations of the environment. To capture these complementary aspects, we evaluate policies under both \emph{in-distribution} (ID) and \emph{out-of-distribution} (OOD) settings. The ID setting measures how well a model can fit and execute tasks within the target environment after fine-tuning, while OOD setting evaluates robustness to variations that are not observed during training, such as changes in object properties or task configurations.

\begin{wrapfigure}{r}{0.50\linewidth}
    \vspace{-6mm}
    \centering
    \includegraphics[width=\linewidth]{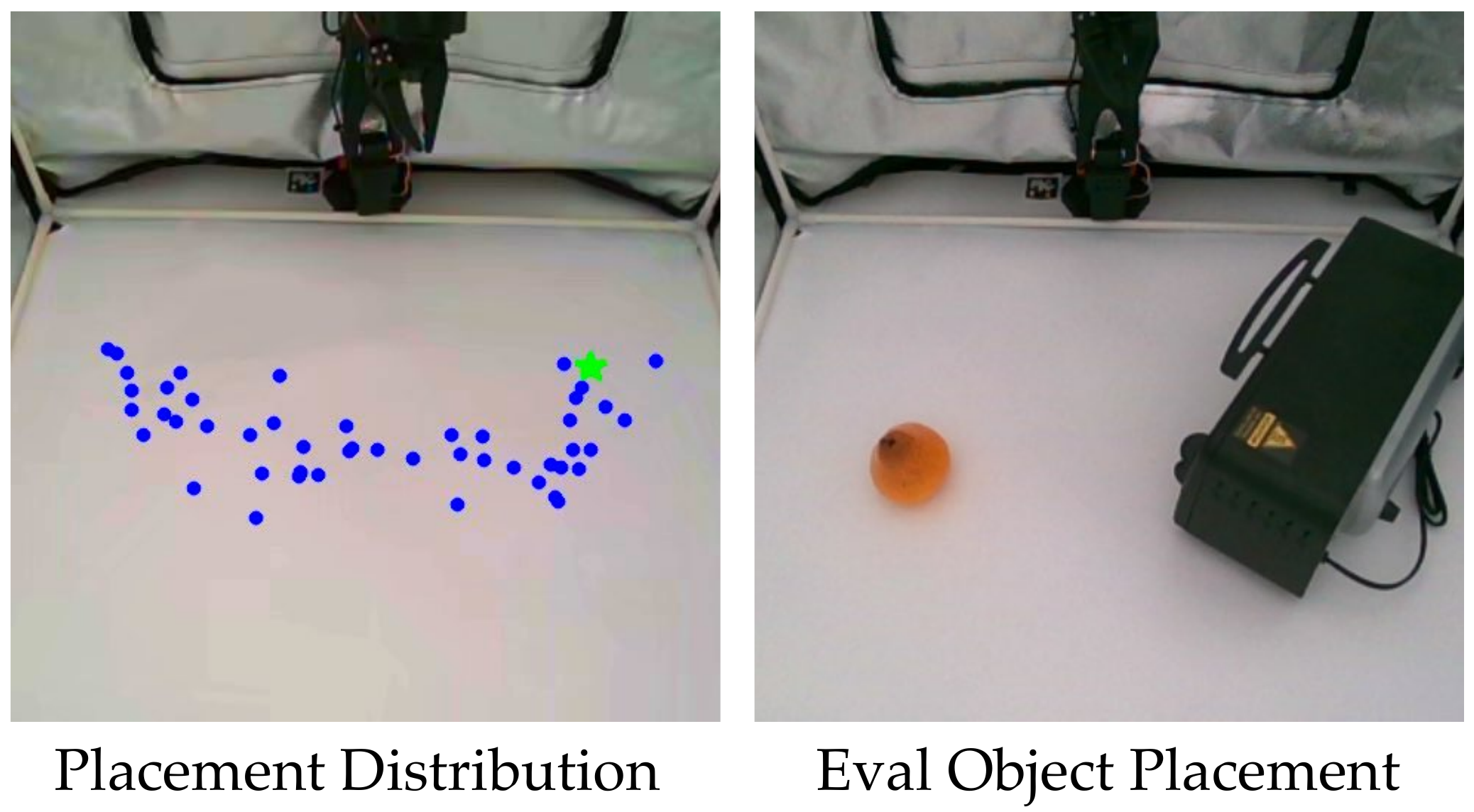}
    \vspace{-5mm}
    \caption{For the task \textit{Open the Oven}, the blue dots indicate the oven center locations in the training set. The green star indicates the selected oven location for a test scene.}
    \vspace{-5mm}
    \label{fig:demo_reference}
    
\end{wrapfigure}

To enable standardized and reproducible evaluation, we define a \emph{test scene} as the initial configuration of all objects in the workspace, including both target and distractor objects. The \emph{VLA-REPLICA} benchmark provides a total of 90 test scenes (50 ID and 40 OOD), ensuring that every evaluation is performed under identical initial conditions. Fig.~\ref{fig:expert_flow}(b) shows some examples of these test scenes. The design of these test scenes is guided by the need to avoid overfitting to training configurations. During expert data collection, object placements are manually randomized while ensuring diversity across demonstrations. We then analyze the spatial distribution of object positions by extracting object centers from all 50 demonstrations for each task (Appendix~\ref{app:demonstration_dataset_object_placements}). This reveals regions with varying densities of object placements (Fig.~\ref{fig:demo_reference}). Based on this distribution, we construct test scenes with controlled deviations from the training data, introducing varying levels of difficulty depending on how far object placements depart from previously seen configurations.

For ID evaluation, we define five test scenes per task that remain within the support of the training distribution while still covering diverse configurations. Across 10 tasks, this results in 50 ID test scenes. Each scene is evaluated once per round in three evaluation rounds per policy, yielding 150 ID trials per policy. For OOD evaluation, we further design test scenes that introduce systematic variations beyond the training distribution. Specifically, we modify object attributes (e.g., color and shape) and task requirements (e.g., memory or sequencing), and evaluate on eight of the ten tasks, resulting in 40 OOD test scenes and 120 OOD trials per policy across three rounds. Each task therefore has 15 trials in either evaluation split; the aggregation and uncertainty convention is defined in Sec.~\ref{sec:algorithm}. These OOD scenarios are designed to probe whether policies can generalize beyond memorized behaviors and adapt to novel conditions within the same physical environment. Table~\ref{tab:task_split_with_id} illustrates the differences between ID and OOD test scenes.

\vspace{-0.1cm}

\subsection{Real-World Scene Reconstruction and Reproducibility}

Ensuring consistent scene setup across different environments is critical for reliable real-world evaluation. Prior work has explored the use of \emph{image overlay} as a practical tool for improving reproducibility~\citep{khargonkar2024scenereplica}. Building on this idea, we adopt a similar overlay-based approach in \emph{VLA-REPLICA} to standardize scene reconstruction. During setup, a live video stream from the top-view camera is overlaid with predefined test scene images, allowing users to directly match object placements to the target configuration. Fig.~\ref{fig:method}(b) illustrates the scene setup process for a single evaluation trial. This design enables fast and intuitive alignment of both target and distractor objects, reducing human effort and minimizing deviations across setups. Refer to Appendix~\ref{app:test_scene_reference_images} for all scenes.

While image-overlay-based methods have been explored in prior real-world benchmarks (e.g., SceneReplica~\citep{khargonkar2024scenereplica} and RoboChallenge~\cite{yakefu2025robochallenge}), we integrate this mechanism into a unified evaluation pipeline tailored for VLA models. In combination with our standardized hardware setup and predefined test scenes, this approach provides a simple yet effective solution for achieving consistent and reproducible real-world evaluation across independent laboratories.

\vspace{-0.15cm}




%% file: sections/experiments.tex
\section{Benchmarking Experiments}
\subsection{Algorithms}
\label{sec:algorithm}
We evaluate three imitation learning baselines and six pretrained VLA policies. All methods use the same 500 expert demonstrations and are trained or fine-tuned for 40K steps. The imitation learning baselines are ACT~\citep{zhao2023learning}, a multi-task Diffusion Transformer (DiT-D), and its flow-matching variant (DiT-F)~\citep{jones2025multitaskditpolicy}. The VLA policies are SmolVLA~\citep{shukor2025smolvla}, X-VLA~\citep{zheng2025x}, $\pi_0$~\citep{black2024pi0visionlanguageactionflowmodel}, $\pi_{0.5}$~\citep{intelligence2025pi05}, MolmoAct2-SO100\_101~\citep{fang2026molmoact2}, and GR00T N1.7~\citep{gr00tn1_2025}. We abbreviate the last two policies as MolmoAct2 and GR00T in the result tables.

ACT learns a transformer-based action-chunking policy from demonstrations. DiT-D uses a text- and vision-conditioned diffusion transformer, while DiT-F replaces the diffusion objective with flow matching. SmolVLA is a lightweight VLA policy; X-VLA supports cross-embodiment control; $\pi_0$ uses a flow-based action model; and $\pi_{0.5}$ extends it toward open-world generalization. MolmoAct2 couples a vision-language backbone with a continuous-action expert; we evaluate its SO100/101 variant. GR00T is a generalist robot foundation model; the evaluated checkpoint is N1.7.

For X-VLA, we fine-tune the vision encoder, language encoder, and policy transformer. For $\pi_0$, we fine-tune the VLM while keeping the vision encoder frozen. All implementations use LeRobot~\citep{cadene2024lerobot}; training configurations are given in Appendix~\ref{app:training}. Before each rollout, objects are reset to their reference-image placements.

\paragraph{Repeated evaluation and uncertainty.}
For each policy, we evaluate each task in three rounds, with one rollout per reference scene in each round. Five scenes per task yield 15 trials per task, or 150 ID and 120 OOD trials per policy. For task $t$, let $s_{t,r}$ be the fraction of successful binary trials over the five scenes in round $r$. We report the mean $\bar{s}_t=\frac{1}{3}\sum_{r=1}^3s_{t,r}$ and sample standard deviation $\sqrt{\frac{1}{2}\sum_{r=1}^3(s_{t,r}-\bar{s}_t)^2}$. For the overall score, we first average the task scores within each round and then compute the mean and standard deviation of these three round-level averages. These deviations describe variability across evaluation rounds, rather than across independently trained models.

\subsection{In-Distribution Policy Evaluation}
\label{indomain_eval}
We evaluate policies in the same physical setting used for demonstration collection, using the ID tasks in Table~\ref{tab:task_split_with_id} with held-out initial object placements. This measures adaptation to the target environment with limited demonstration data.

Table~\ref{tab:ID_success_rate} reports the updated repeated-evaluation results. $\pi_{0.5}$ has the highest mean overall ID score ($0.53\pm0.05$), followed by GR00T ($0.42\pm0.06$) and MolmoAct2 ($0.39\pm0.06$). All fine-tuned VLA policies (except X-VLA) have higher mean overall ID scores than the three baselines trained from scratch. Performance remains task-dependent: bread placement and towel folding are relatively successful for the strongest policies, while block manipulation and counting tasks remain challenging. These rankings compare observed means; the three evaluation rounds do not establish statistically significant differences between policies.

\providecommand{\replicascore}[2]{%
  \ensuremath{#1\,{\scriptstyle\pm #2}}%
}
\begin{table}[t]
\caption{ID success rates on ten tasks. Each cell shows the mean above $\pm$ one sample standard deviation across three rounds of five scenes (15 trials per task). The average is computed from round-level task averages. MolmoAct2 denotes SO100\_101 and GR00T denotes N1.7. Task numbering follows Table~\ref{tab:task_split_with_id}.}
\label{tab:ID_success_rate}
\centering\footnotesize
\setlength{\tabcolsep}{3pt}
\renewcommand{\arraystretch}{1.05}
\resizebox{\linewidth}{!}{\begin{tabular}{l*{9}{c}}
\toprule
\textbf{Task}
& \textbf{ACT}~\citep{zhao2023learning}
& \textbf{DiT-D}~\citep{jones2025multitaskditpolicy}
& \textbf{DiT-F}~\citep{jones2025multitaskditpolicy}
& \textbf{SmolVLA}~\citep{shukor2025smolvla}
& \textbf{X-VLA}~\citep{zheng2025x}
& $\bm{\pi}_0$~\citep{black2024pi0visionlanguageactionflowmodel}
& $\bm{\pi}_{0.5}$~\citep{intelligence2025pi05}
& \textbf{MolmoAct2}~\citep{fang2026molmoact2}
& \textbf{GR00T}~\citep{gr00tn1_2025} \\
\midrule
1 Bread on plate & \replicascore{0.33}{0.12}&\replicascore{0.20}{0.20}&\replicascore{0.13}{0.23}&\replicascore{0.80}{0.20}&\replicascore{0.27}{0.12}&\replicascore{0.67}{0.12}&\replicascore{0.80}{0.20}&\replicascore{\mathbf{0.93}}{0.12}&\replicascore{0.80}{0.00}\\
2 Bowl on coaster & \replicascore{0.07}{0.12}&\replicascore{0.13}{0.12}&\replicascore{0.00}{0.00}&\replicascore{0.20}{0.20}&\replicascore{0.13}{0.12}&\replicascore{0.67}{0.31}&\replicascore{\mathbf{0.87}}{0.12}&\replicascore{0.73}{0.12}&\replicascore{\mathbf{0.87}}{0.12}\\
3 Stack blocks & \replicascore{0.00}{0.00}&\replicascore{0.00}{0.00}&\replicascore{0.00}{0.00}&\replicascore{0.27}{0.12}&\replicascore{0.00}{0.00}&\replicascore{0.00}{0.00}&\replicascore{\mathbf{0.33}}{0.12}&\replicascore{0.27}{0.12}&\replicascore{0.00}{0.00}\\
4 Blocks into box & \replicascore{0.00}{0.00}&\replicascore{0.07}{0.12}&\replicascore{0.00}{0.00}&\replicascore{0.00}{0.00}&\replicascore{0.00}{0.00}&\replicascore{0.00}{0.00}&\replicascore{\mathbf{0.47}}{0.12}&\replicascore{0.27}{0.12}&\replicascore{0.29}{0.08}\\
5 Fold towel & \replicascore{0.60}{0.20}&\replicascore{0.20}{0.00}&\replicascore{0.27}{0.12}&\replicascore{0.73}{0.12}&\replicascore{0.53}{0.12}&\replicascore{0.67}{0.23}&\replicascore{\mathbf{1.00}}{0.00}&\replicascore{0.73}{0.12}&\replicascore{\mathbf{1.00}}{0.00}\\
6 Open oven & \replicascore{0.33}{0.12}&\replicascore{0.20}{0.35}&\replicascore{0.13}{0.23}&\replicascore{0.33}{0.12}&\replicascore{0.07}{0.12}&\replicascore{0.20}{0.00}&\replicascore{\mathbf{0.53}}{0.12}&\replicascore{0.13}{0.23}&\replicascore{0.47}{0.31}\\
7 Erase whiteboard & \replicascore{0.20}{0.00}&\replicascore{0.20}{0.00}&\replicascore{0.13}{0.12}&\replicascore{0.20}{0.00}&\replicascore{0.00}{0.00}&\replicascore{0.33}{0.12}&\replicascore{0.33}{0.12}&\replicascore{\mathbf{0.53}}{0.12}&\replicascore{0.40}{0.20}\\
8 Shake pepper $n$ times & \replicascore{0.27}{0.12}&\replicascore{0.07}{0.12}&\replicascore{0.00}{0.00}&\replicascore{0.00}{0.00}&\replicascore{0.20}{0.20}&\replicascore{0.27}{0.12}&\replicascore{\mathbf{0.33}}{0.12}&\replicascore{0.13}{0.12}&\replicascore{0.20}{0.00}\\
9 Lift bowl $n$ times & \replicascore{0.33}{0.23}&\replicascore{0.00}{0.00}&\replicascore{0.00}{0.00}&\replicascore{0.20}{0.00}&\replicascore{0.07}{0.12}&\replicascore{0.20}{0.00}&\replicascore{\mathbf{0.40}}{0.00}&\replicascore{0.13}{0.12}&\replicascore{0.07}{0.12}\\
10 Press button $n$ times & \replicascore{0.00}{0.00}&\replicascore{0.00}{0.00}&\replicascore{0.00}{0.00}&\replicascore{0.20}{0.00}&\replicascore{0.00}{0.00}&\replicascore{0.20}{0.00}&\replicascore{\mathbf{0.27}}{0.31}&\replicascore{0.00}{0.00}&\replicascore{0.13}{0.12}\\
\midrule
Average & \replicascore{0.21}{0.08}&\replicascore{0.11}{0.05}&\replicascore{0.07}{0.05}&\replicascore{0.29}{0.04}&\replicascore{0.13}{0.02}&\replicascore{0.32}{0.03}&\replicascore{\mathbf{0.53}}{0.05}&\replicascore{0.39}{0.06}&\replicascore{0.42}{0.06}\\
\bottomrule
\end{tabular}}
\end{table}

\subsection{Out-of-distribution Policy Evaluation}
\label{ood_eval}
We evaluate eight OOD tasks that change object attributes or task requirements beyond the training distribution (Table~\ref{tab:task_split_with_id}). Open Oven and Erase Whiteboard have no OOD variants. For counting tasks, held-out repetition counts test whether policies can track repeated actions; for example, the pepper task includes one, two, or three shakes in ID and four or five shakes in OOD, as well as a change in target plate color.

Table~\ref{tab:OOD_success_rate} shows that MolmoAct2 achieves the highest mean overall OOD score ($0.47\pm0.05$), followed by $\pi_{0.5}$ ($0.39\pm0.10$) and GR00T ($0.35\pm0.00$). MolmoAct2's mean OOD score exceeds its ID score, so an overall OOD performance drop is not universal. MolmoAct2 performs particularly well on OOD bread placement, bowl placement, stacking, and towel folding. Counting remains difficult for every policy: no model exceeds a mean success rate of 0.27 on any OOD counting task. The results support visual generalization on selected tasks while exposing the limited capabilities of VLAs on memory-related tasks, such as counting.

\begin{table}[t]
\caption{OOD success rates on eight tasks, with the same mean and standard-deviation convention as Table~\ref{tab:ID_success_rate}. Task numbers match the main task suite; Tasks 6 and 7 have no OOD variants.}
\label{tab:OOD_success_rate}
\centering\footnotesize
\setlength{\tabcolsep}{3pt}
\renewcommand{\arraystretch}{1.05}
\resizebox{\linewidth}{!}{\begin{tabular}{l*{9}{c}}
\toprule
\textbf{Task}
& \textbf{ACT}~\citep{zhao2023learning}
& \textbf{DiT-D}~\citep{jones2025multitaskditpolicy}
& \textbf{DiT-F}~\citep{jones2025multitaskditpolicy}
& \textbf{SmolVLA}~\citep{shukor2025smolvla}
& \textbf{X-VLA}~\citep{zheng2025x}
& $\bm{\pi}_0$~\citep{black2024pi0visionlanguageactionflowmodel}
& $\bm{\pi}_{0.5}$~\citep{intelligence2025pi05}
& \textbf{MolmoAct2}~\citep{fang2026molmoact2}
& \textbf{GR00T}~\citep{gr00tn1_2025} \\
\midrule
1 Bread on plate & \replicascore{0.40}{0.20}&\replicascore{0.07}{0.12}&\replicascore{0.07}{0.12}&\replicascore{0.73}{0.12}&\replicascore{0.33}{0.31}&\replicascore{0.73}{0.12}&\replicascore{0.93}{0.12}&\replicascore{0.87}{0.12}&\replicascore{\mathbf{0.93}}{0.12}\\
2 Bowl on coaster & \replicascore{0.27}{0.12}&\replicascore{0.33}{0.23}&\replicascore{0.13}{0.23}&\replicascore{0.33}{0.12}&\replicascore{0.00}{0.00}&\replicascore{0.67}{0.12}&\replicascore{0.60}{0.20}&\replicascore{\mathbf{0.87}}{0.12}&\replicascore{0.67}{0.12}\\
3 Stack blocks & \replicascore{0.00}{0.00}&\replicascore{0.00}{0.00}&\replicascore{0.00}{0.00}&\replicascore{0.20}{0.20}&\replicascore{0.00}{0.00}&\replicascore{0.20}{0.20}&\replicascore{0.20}{0.20}&\replicascore{\mathbf{0.60}}{0.00}&\replicascore{0.00}{0.00}\\
4 Blocks into box & \replicascore{0.00}{0.00}&\replicascore{0.00}{0.00}&\replicascore{0.00}{0.00}&\replicascore{0.13}{0.12}&\replicascore{0.00}{0.00}&\replicascore{0.00}{0.00}&\replicascore{\mathbf{0.27}}{0.12}&\replicascore{0.20}{0.20}&\replicascore{0.13}{0.12}\\
5 Fold towel & \replicascore{0.00}{0.00}&\replicascore{0.07}{0.12}&\replicascore{0.07}{0.12}&\replicascore{0.47}{0.12}&\replicascore{0.07}{0.12}&\replicascore{0.53}{0.06}&\replicascore{0.80}{0.00}&\replicascore{0.87}{0.12}&\replicascore{\mathbf{1.00}}{0.00}\\
8 Shake pepper $n$ times & \replicascore{0.00}{0.00}&\replicascore{0.07}{0.12}&\replicascore{0.00}{0.00}&\replicascore{0.00}{0.00}&\replicascore{0.00}{0.00}&\replicascore{0.22}{0.03}&\replicascore{\mathbf{0.27}}{0.23}&\replicascore{0.13}{0.12}&\replicascore{0.00}{0.00}\\
9 Lift bowl $n$ times & \replicascore{0.00}{0.00}&\replicascore{0.00}{0.00}&\replicascore{0.00}{0.00}&\replicascore{\mathbf{0.20}}{0.00}&\replicascore{0.00}{0.00}&\replicascore{0.00}{0.00}&\replicascore{0.00}{0.00}&\replicascore{0.13}{0.12}&\replicascore{0.07}{0.12}\\
10 Press button $n$ times & \replicascore{0.00}{0.00}&\replicascore{\mathbf{0.07}}{0.12}&\replicascore{0.00}{0.00}&\replicascore{\mathbf{0.07}}{0.12}&\replicascore{0.00}{0.00}&\replicascore{\mathbf{0.07}}{0.12}&\replicascore{\mathbf{0.07}}{0.12}&\replicascore{\mathbf{0.07}}{0.12}&\replicascore{0.00}{0.00}\\
\midrule
Average & \replicascore{0.08}{0.01}&\replicascore{0.08}{0.04}&\replicascore{0.04}{0.01}&\replicascore{0.27}{0.04}&\replicascore{0.05}{0.04}&\replicascore{0.30}{0.02}&\replicascore{0.39}{0.10}&\replicascore{\mathbf{0.47}}{0.05}&\replicascore{0.35}{0.00}\\
\bottomrule
\end{tabular}}
\end{table}

\vspace{-3mm}

\subsection{Reproducibility Analysis}
\label{sec:reproducibility}
\begin{figure}[!htbp]
\centering
\includegraphics[width=0.75\linewidth]{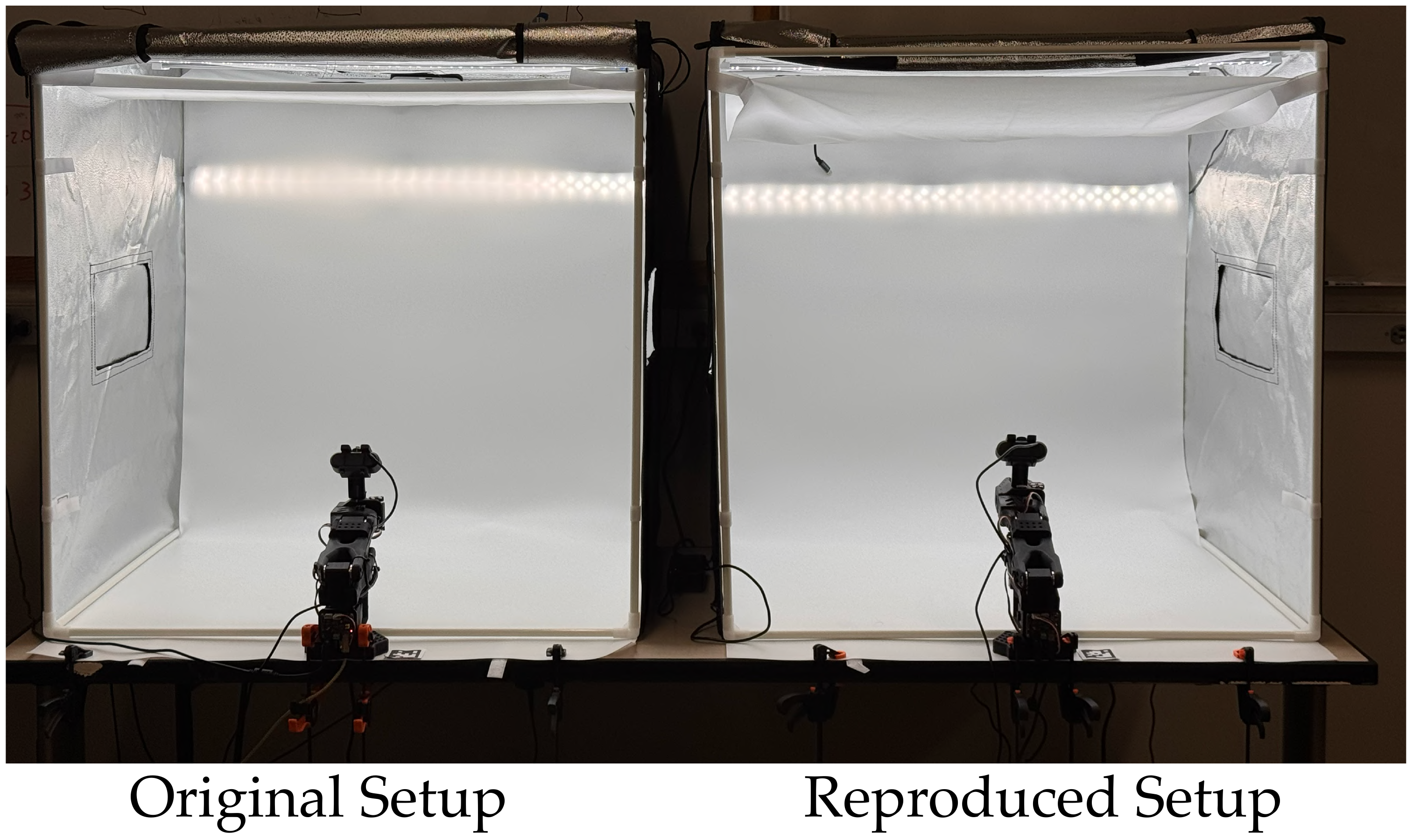}
\caption{Original and independently reproduced benchmark setups.}
\label{fig:reproduced_setup}
\end{figure}

We built a second instance of the benchmark at a different location
(Fig.~\ref{fig:reproduced_setup}), using a separate but identical light box,
cameras, and SO-101 follower arm. A user without prior benchmark experience
assembled it following Appendix~\ref{app:setup}. Cameras were calibrated to
match the original views, and policy outputs were mapped to the new arm using
Eq.~\eqref{eq3} in Sec.~\ref{sec:reproduce}.

To strengthen the reproducibility analysis, we evaluated ACT, SmolVLA, $\pi_0$, and $\pi_{0.5}$ on all 10 ID
tasks and all 8 OOD tasks, with one rollout for each of five scenes per task.
This yields 50 ID and 40 OOD evaluations per policy (360 rollouts in total).
Table~\ref{tab:id_ood_success_rate} reports per-task and overall success rates
for the reproduced setup. 

Across the four evaluated policies, the reproduced setup generally achieves task-level success rates comparable to those observed in the original environment. Many reproduced scores lie within the original evaluation's reported round-to-round variability, including results for challenging tasks such as opening the oven and erasing the whiteboard. These findings support the practical reproducibility of VLA-Replica.

\vspace{-4mm}
\begin{table}[!htbp]
\centering
\caption{Task-level success rates in the reproduced environment across all
10 ID and 8 OOD tasks. Each entry is an average over five scene evaluations (one rollout per scene). The average success from the original evaluation environment is listed below as comparison.}
\label{tab:id_ood_success_rate}
\begingroup
\scriptsize
\renewcommand{\arraystretch}{1.14}
\setlength{\tabcolsep}{2pt}
\begin{minipage}[t]{0.49\linewidth}
\vspace{5pt}
\centering
\textbf{(a) In-distribution (ID)}\par\smallskip
\begin{tabular}{@{}>{\raggedright\arraybackslash}p{0.43\linewidth}cccc@{}}
\toprule
\textbf{Task} & \textbf{ACT} & \textbf{SmolVLA} & \textbf{$\pi_0$} & \textbf{$\pi_{0.5}$} \\
\midrule
Place Bread on Plate & 0.40 & 0.60 & \textbf{0.80} & \textbf{0.80} \\
Place Bowl on Coaster & 0.00 & 0.40 & 0.60 & \textbf{1.00} \\
Stack Blocks & 0.00 & 0.00 & 0.20 & \textbf{0.60} \\
Put All Blocks into Box & 0.00 & 0.00 & 0.00 & \textbf{0.40} \\
Fold Towel & 0.60 & \textbf{0.80} & 0.60 & \textbf{0.80} \\
Open Oven Door & \textbf{0.40} & \textbf{0.40} & 0.20 & \textbf{0.40} \\
Erase Whiteboard & 0.20 & 0.20 & 0.20 & \textbf{0.40} \\
Shake Pepper $N$ Times & \textbf{0.40} & 0.00 & 0.20 & \textbf{0.40} \\
Lift Bowl $N$ Times & 0.20 & 0.20 & 0.20 & \textbf{0.40} \\
Press Button $N$ Times & 0.00 & \textbf{0.20} & \textbf{0.20} & \textbf{0.20} \\
\midrule
\textbf{Avg. success (reproduced)} & \textbf{0.22} & \textbf{0.28} & \textbf{0.32} & \textbf{0.54} \\
\midrule
\textbf{Avg. success (original)} & \textbf{0.21} & \textbf{0.29} & \textbf{0.32} & \textbf{0.53} \\
\bottomrule

\end{tabular}
\end{minipage}\hfill%
\begin{minipage}[t]{0.49\linewidth}
\vspace{5pt}
\centering
\textbf{(b) Out-of-distribution (OOD)}\par\smallskip
\begin{tabular}{@{}>{\raggedright\arraybackslash}p{0.43\linewidth}cccc@{}}
\toprule
\textbf{Task} & \textbf{ACT} & \textbf{SmolVLA} & \textbf{$\pi_0$} & \textbf{$\pi_{0.5}$} \\
\midrule
Place Bread on Plate & 0.40 & 0.60 & \textbf{0.80} & \textbf{0.80} \\
Place Bowl on Coaster & \textbf{0.40} & \textbf{0.40} & \textbf{0.40} & \textbf{0.40} \\
Stack Blocks & 0.00 & 0.00 & \textbf{0.20} & \textbf{0.20} \\
Put All Blocks into Box & 0.00 & 0.20 & 0.00 & \textbf{0.40} \\
Fold Towel & 0.00 & 0.60 & 0.60 & \textbf{0.80} \\
Oven Oven Door & -- & -- & -- & -- \\
Erase Whiteboard & -- & -- & -- & -- \\
Shake Pepper $N$ Times & 0.00 & 0.00 & \textbf{0.20} & 0.00 \\
Lift Bowl $N$ Times & 0.00 & \textbf{0.20} & \textbf{0.20} & \textbf{0.20} \\
Press Button $N$ Times & 0.00 & 0.00 & \textbf{0.20} & \textbf{0.20} \\
\midrule
\textbf{Avg. success (reproduced)} & \textbf{0.10} & \textbf{0.25} & \textbf{0.33} & \textbf{0.38} \\
\midrule
\textbf{Avg. success (original)} & \textbf{0.08} & \textbf{0.27} & \textbf{0.30} & \textbf{0.39} \\
\bottomrule
\end{tabular}
\end{minipage}
\endgroup
\end{table}

\subsection{Ablation Study}
\label{sec:ablation_study}

To assess the contribution of standardized evaluation conditions, we
ablated two components of the VLA-Replica setup: (i) the uniform lighting
inside the enclosure and (ii) the wrist-camera placement. We evaluated
ACT and $\pi_{0.5}$ on three ID tasks spanning different levels of difficulty: folding
a towel, placing all blocks into a box, and pressing a button $N$ times.
For the lighting ablation, we turned off the enclosure lighting and used
ambient overhead room lighting, resulting in low illumination within enclosure. For the camera ablation, we displaced the
wrist camera 2\,cm to the left of its standardized position. Each condition
was evaluated in three runs, with five scenes per task in each run;
Table~\ref{tab:standardization_ablation} reports the mean and standard
deviation of the task-level success rate across runs.

\vspace{-3mm}
\begin{table}[!htbp]
\centering
\caption{Ablation of standardized lighting and wrist-camera placement on
three in-distribution (ID) tasks. Values are mean $\pm$ standard deviation
of task success rates over three evaluation runs (five scenes per task per
run). The control uses the default standardized setup.}
\vspace{0.3mm}
\label{tab:standardization_ablation}
\footnotesize
\renewcommand{\arraystretch}{1.15}
\setlength{\tabcolsep}{3.5pt}
\resizebox{0.65\linewidth}{!}{%
\begin{tabular}{llccc}
\toprule
\textbf{Policy} & \textbf{Task} & \textbf{Control (default)} &
\textbf{Box lighting off} & \textbf{Wrist camera left 2\,cm} \\
\midrule
ACT~\cite{zhao2023learning} & Fold Towel & $0.60 \pm 0.20$ & $0.07 \pm 0.12$ & $0.40 \pm 0.20$ \\
& Put All Blocks into Box & $0.00 \pm 0.00$ & $0.00 \pm 0.00$ & $0.00 \pm 0.00$ \\
& Press Button $N$ Times & $0.00 \pm 0.00$ & $0.00 \pm 0.00$ & $0.00 \pm 0.00$ \\
\midrule
$\pi_{0.5}$~\cite{intelligence2025pi05} & Fold Towel & $1.00 \pm 0.00$ & $0.73 \pm 0.12$ & $0.40 \pm 0.20$ \\
& Put All Blocks into Box & $0.47 \pm 0.12$ & $0.13 \pm 0.12$ & $0.00 \pm 0.00$ \\
& Press Button $N$ Times & $0.27 \pm 0.31$ & $0.07 \pm 0.12$ & $0.07 \pm 0.12$ \\
\bottomrule
\end{tabular}%
}
\end{table}

Removing the enclosure lighting substantially reduced ACT's fold-towel
success rate (from $0.60$ to $0.07$). Although $\pi_{0.5}$ was more robust
to this change, its success rates decreased on all three tasks. Shifting
the wrist camera by 2\,cm also reduced performance, particularly for
$\pi_{0.5}$ on folding the towel (from $1.00$ to $0.40$) and placing blocks
in the box (from $0.47$ to $0.00$). These results suggest that consistent
illumination and camera placement matter for reliable cross-policy
comparisons, even when the physical robot and task instructions are
unchanged.

We additionally performed a preliminary calibration perturbation test by
adding Gaussian angular noise ($\mu=0^{\circ}$,
$\sigma=5^{\circ}$) to the SO-101 arm calibration parameters. In the
observed $\pi_{0.5}$ rollouts on the same three tasks, the perturbed arm
failed to align its end effector with the target objects and did not
complete the tasks. Because this test was not conducted as a full repeated
quantitative evaluation, we report it qualitatively rather than including
it in Table~\ref{tab:standardization_ablation}. Together, these ablations
motivate the benchmark's standardized lighting, camera geometry, and arm
calibration procedures.

\vspace{-3mm}

%% file: sections/conclusion.tex
\section{Conclusion and Limitations}
We introduced \textbf{VLA-REPLICA}, a low-cost and reproducible real-world benchmark for vision-language-action models. The benchmark combines an affordable hardware setup, standardized scene construction, and evaluation of both in-distribution adaptation and out-of-distribution generalization.

Repeated evaluation of nine policies reveals complementary strengths: $\pi_{0.5}$ has the highest mean overall ID score, while MolmoAct2 has the highest mean overall OOD score. Counting tasks remain difficult across policies. A separate evaluation on an independently reconstructed setup across the complete ID and OOD task suites provides further evidence of cross-site reproducibility for four representative policies. We hope VLA-REPLICA supports accessible and consistent evaluation across laboratories.

\textbf{Limitations.} The benchmark focuses on tabletop manipulation with one low-cost robot embodiment, and its task suite remains smaller than large-scale simulation benchmarks. Three evaluation rounds provide a limited estimate of rollout variability. The cross-setup experiment covers four policies in a single round, so variability may still exist. Small hardware, lighting, and placement differences can still affect results. Further evaluations across additional laboratories, hardware instances, and repeated trials are needed to quantify between-site variability.

\section*{Acknowledgments}
This work was supported in part by the National Science Foundation (NSF) under Grant Nos. 2346528 and 2520553, the NVIDIA Academic Grant Program Award, and a gift funding from XPeng.


%% file: sections/appendix.tex
%
%

\newcommand{\TODO}[1]{\textcolor{red}{\textbf{[TODO: #1]}}}
\appendix
\counterwithin{figure}{section}
\counterwithin{table}{section}


\section{VLA-REPLICA Benchmark Setup Instructions}
\label{app:setup}


This section provides step-by-step instructions for reliably reproducing our benchmark
environment across different laboratories (\href{https://irvlutd.github.io/VLAReplica/}{online version}).
  
\textbf{Please read and follow these instructions carefully.
Careful inspection of the photographs and text is crucial for creating a
reproducible and precise environment. }

\subsection{Environment Overview}
\label{subsec:environment_overview}

\begin{figure}[h]
  \centering
  \includegraphics[width=\linewidth]{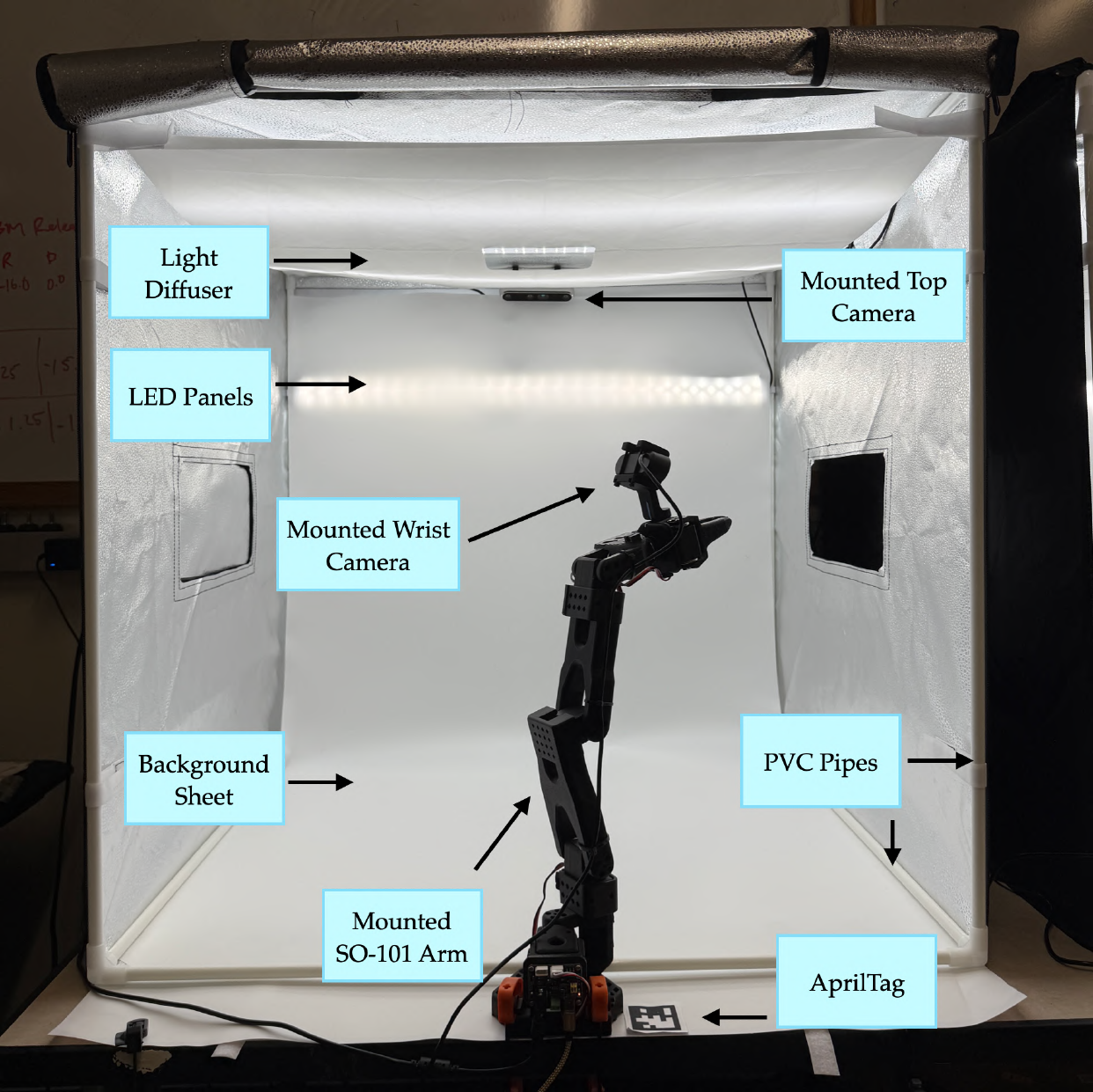}
  \caption{\textbf{Benchmark environment.} Physical workspace showing the SO-101 follower arm, 32$\times$32\,in light box, LED panel, white background sheet, and
    AprilTag.
  }
  \label{fig:env_overview}
\end{figure}

The benchmark comprises of the components listed in Table~\ref{tab:parts}. The cameras, after calibration, handle RGB observation, while the SO-101 follower arm executes manipulation tasks.

\begin{table}[h]
  \centering
  \caption{Parts list for the benchmark setup.}
  \label{tab:parts}
  \small
  \scalebox{0.9}{
  \begin{tabularx}{\textwidth}{lX} 
    \toprule
    \textbf{Qty} & \textbf{Item} \\
    \midrule
    1 & Glendan 32$\times$32\,in box set (box tarp, 12$\times$ PVC pipes, 
        8$\times$ PVC edge connectors, white PP background sheet, 
        white light diffuser sheet, 3$\times$LED panel set, power cables) \href{https://a.co/d/0iGoGvFo}{\color{blue}(link)} \\
    \addlinespace
    1 & Intel RealSense D455 \\
    \addlinespace
    1 set & 3-D printed camera mount 
        (1$\times$ backplate, 2$\times$ snap-hooks, 
         2$\times$ M3$\times$6\,mm screws, 1$\times$ M3$\times$12\,mm screws, 1$\times$ M3 nut, 2$\times$ M4$\times$6\,mm screws) \href{https://utdallas.box.com/s/us6poai696dsp02hwfdcjaed1dey2v6p}{\color{blue}(link)}\\
    \addlinespace
    1 & SO-101 follower arm \href{https://shop.wowrobo.com/products/so-arm101-diy-kit-assembled-version-1}{\color{blue}(link)}+ 3-D printed wrist camera mount \href{https://github.com/TheRobotStudio/SO-ARM100/blob/main/Optional/Wrist_Cam_Mount_Vinmooog_Webcam/stl/Webcam_Mount_Wrist.stl}{\color{blue}(link)}
        + Vinmooog RGB camera \href{https://www.amazon.com.be/-/en/Vinmooog-Streaming-Equipment-Microphone-Conferences/dp/B0BG1YJWFN}{\color{blue}(link)}+ 12\,V power adapter \\
    \addlinespace
    1 & 4\,cm tag size \emph{tag36h11} AprilTag \href{https://chaitanyantr.github.io/apriltag.html}{\color{blue}(generate it here)} \\
    \addlinespace
    4 & Clamps \\
    \addlinespace
    1 roll & Rubber grip tape \href{https://a.co/d/0cYfPnRP}{\color{blue}(link)}\\
    \addlinespace
    1 (opt.) & SO-101 leader arm + 6\,V power adaptor \\
    \bottomrule
  \end{tabularx}
  }
\end{table}

\subsection{Assemble the Camera Mounts}
\label{subsec:camera_mount_assembly}
\noindent\textbf{Top Camera:} Before mounting hardware, 3-D print the two required parts (PLA/PETG, standard 15\% infill, 0.4mm) and assemble the
top camera mount (for the D455 Realsense) as follows.

\begin{figure}[h]
  \centering
  \begin{subfigure}[b]{0.30\linewidth}
    \includegraphics[width=\linewidth]{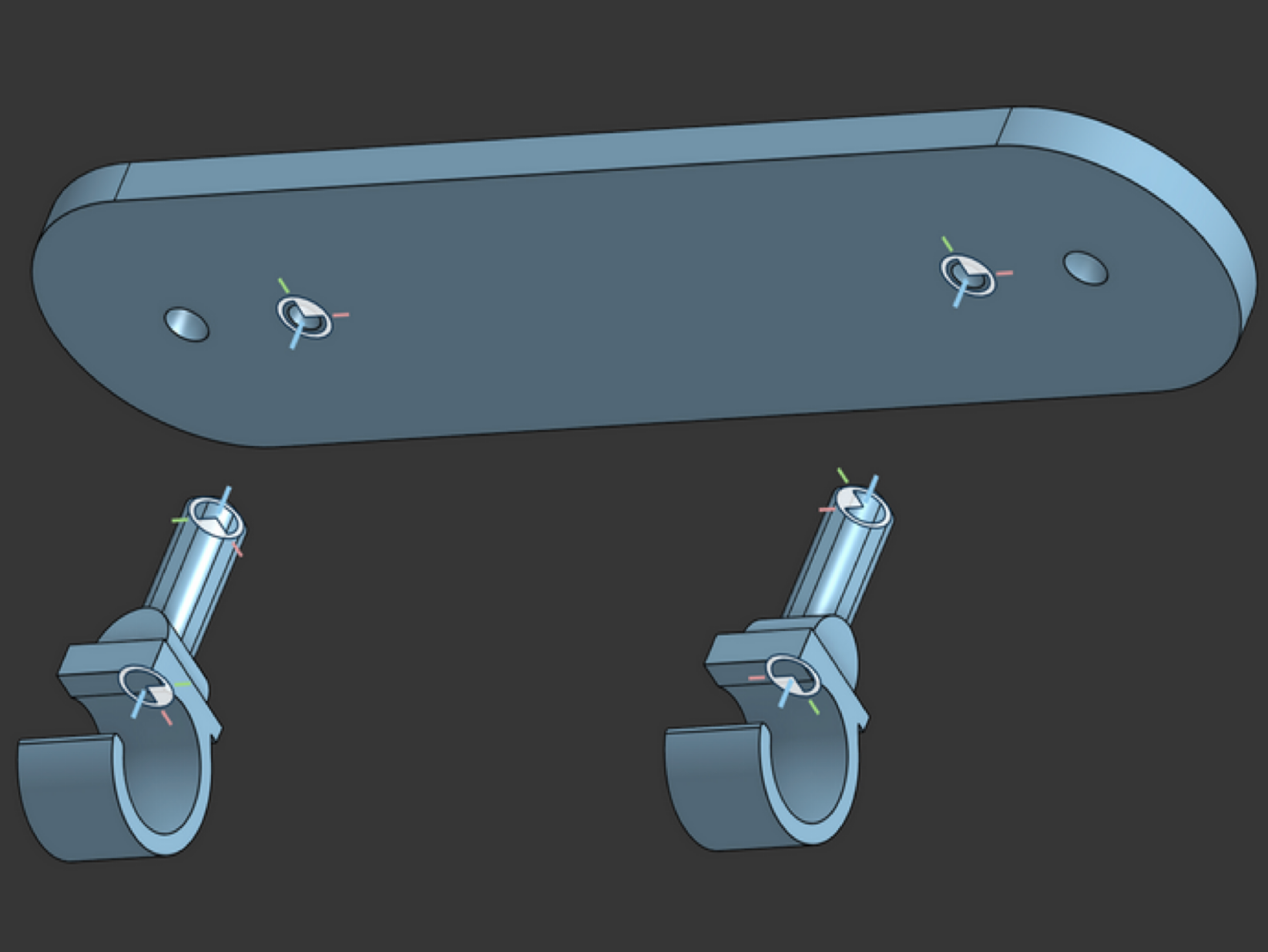}
    \caption{Snap-hooks attached to backplate (Part~3).}
  \end{subfigure}\hfill
  \begin{subfigure}[b]{0.30\linewidth}
    \includegraphics[width=\linewidth]{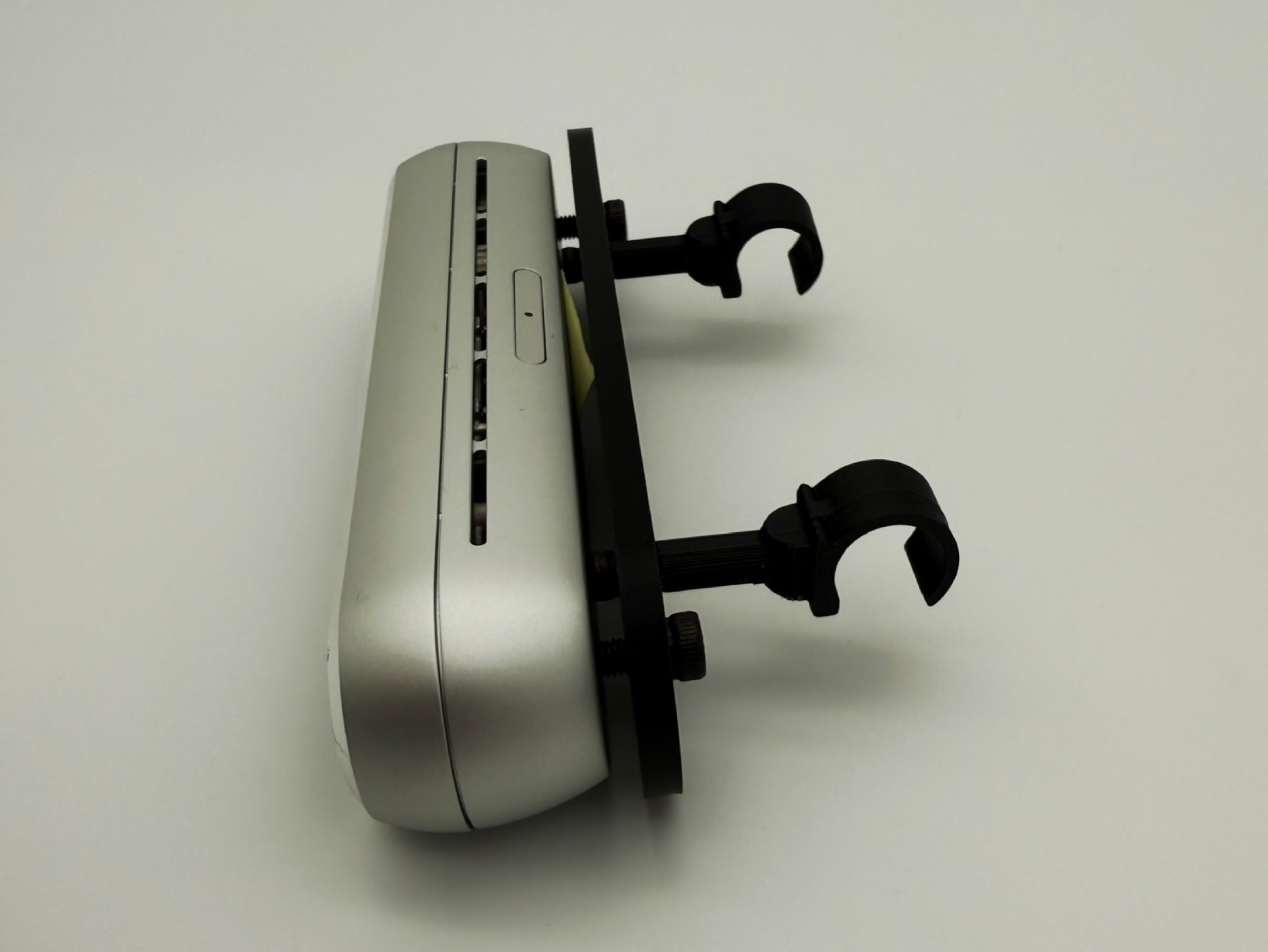}
    \caption{Part~3 screwed to D455 camera.}
  \end{subfigure}\hfill
  \begin{subfigure}[b]{0.30\linewidth}
    \includegraphics[width=\linewidth]{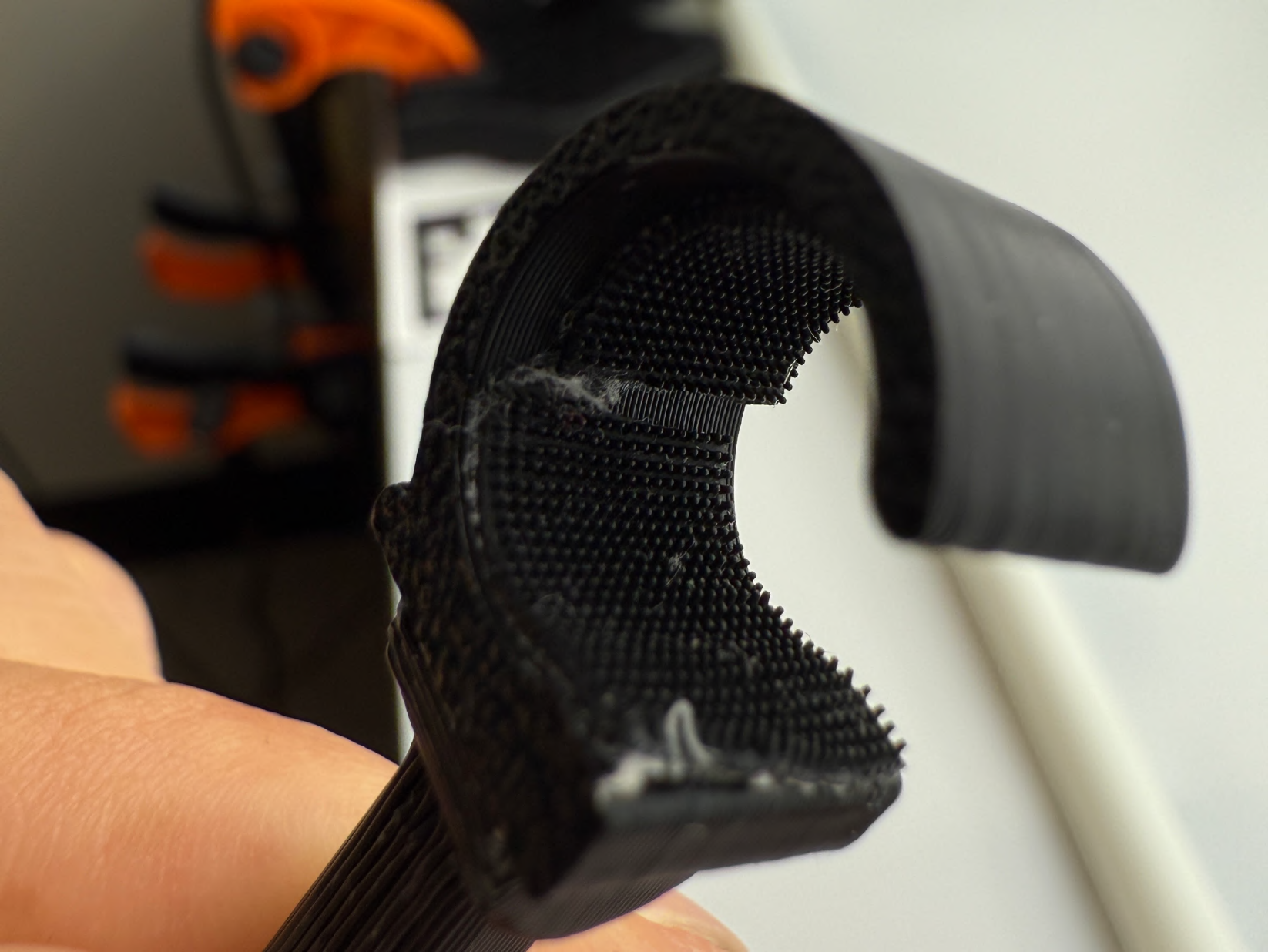}
    \caption{Rubber grip tape on inside of hooks.}
  \end{subfigure}
  \caption{\textbf{Camera mount assembly.}}
  \label{fig:cam_mount}
\end{figure}

\begin{enumerate}
  \item Print \textbf{two copies} of the snap-hook part (\texttt{Part1.stl}).

  \item Print \textbf{one copy} of the camera backplate (\texttt{Part2.stl}).

  \item Attach one snap-hook (Part~1) to the backplate (Part~2) using one
    M3$\times$6\,mm screw.
    Repeat for the second hook.
    The assembled unit is referred to as \emph{Part~3}
    (Fig.~\ref{fig:cam_mount}(a)).

  \item Screw Part~3 tightly to the \textbf{rear mounting holes} of the D455
    camera using two M4$\times$6\,mm screws
    (Fig.~\ref{fig:cam_mount}(b)).

  \item To prevent the hooks from sliding on the PVC pipe, attach a small
    piece of rubber grip tape to the \emph{inside} of each hook
    (Fig.~\ref{fig:cam_mount}(c)).
    \textbf{Do not over-tighten the screws or apply excessive force to the
    hooks, as they may snap.}
\end{enumerate}

Note: the CAD file for the snap-hook has an inner diameter that matches the outer diameter of the PVC pipe for the Glendan light box. Other light boxes may have different PVC diameters and may require manual re-CAD.

\noindent\textbf{SO-101 and Wrist Camera setup:}
Before mounting hardware, 3-D print the one required part (PLA/PETG, standard 15\% infill, 0.4mm) and mount the wrist camera as follows.
\begin{enumerate}
  \item If applicable (i.e. the SO-101 did not come pre-assembled), follow LeRobot's SO-101 documentation page to assemble the SO-101 Follower arm: (\url{https://huggingface.co/docs/lerobot/so101}). \textbf{Don't calibrate the assembled SO-101 arm yet.}
  \item Follow TheRobotStudio's page to print and set up the wrist camera mount with the Vinmooog webcam: (\url{https://github.com/TheRobotStudio/SO-ARM100/tree/main/Optional/Wrist_Cam_Mount_Vinmooog_Webcam})
  \item Secure the camera mount onto the end-effector of the SO-101 with one M3$\times$12 mm screw and the M3 nut.

\end{enumerate}

\noindent\textbf{Important Checklist:}
\begin{todolist}
  \item Both snap-hooks are attached with M3 screws and sit flush against
    the backplate.
  \item The mount is fastened to the D455 with M4 screws; the camera does
    not wobble.
  \item Rubber grip tape is applied to the inside of both hooks.
  \item The SO-101 follower arm is assembled, with the wrist camera mounted in place.
\end{todolist}

\subsection{Set Up the Light Box}
\label{subsec:lightbox_setup}

A controlled environment with consistent lighting and a clutter-free background is essential for consistent visual observations across laboratory settings. To do this, we set up a 32$\times$32\,in. Glendan light box at the edge of a work table.

\textbf{Read the orientation convention below before assembly.}

\begin{figure}[h]
  \centering
  \includegraphics[width=0.55\linewidth]{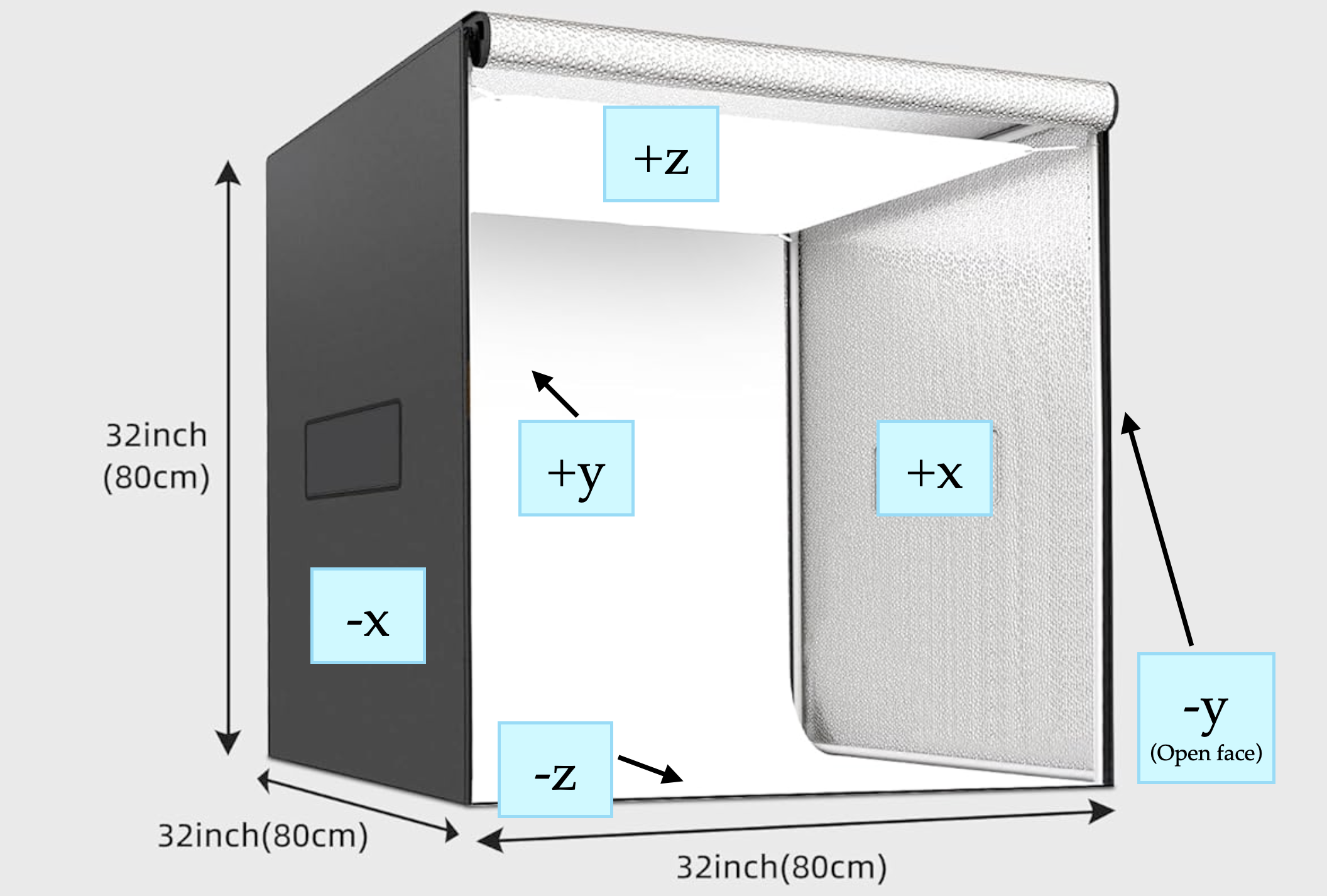}
  \caption{\textbf{Light box coordinate system.}
    $-y$ is the open face; $+z$ is the top; $-z$ is the workspace floor.}
  \label{fig:box_orientation}
\end{figure}

\paragraph{Orientation convention}
We define the box coordinate system as follows
(see Fig.~\ref{fig:box_orientation}):
\begin{itemize}
  \item $-y$: open face of the box (with full-length zippers). When unzipped, the box tarp should open upwards. The SO-101 arm is mounted near this face, pointing \emph{into} the box ($+y$ direction).
  \item $+z$: top face (round hole for the top camera); 
  \item $-z$: bottom face (workspace where objects are placed); SO-101 arm mounted on this face.
  \item $\pm x$: side faces (removable Velcro panels).
  \begin{itemize}
      \item During data capture and policy evaluation, the top camera is zoomed in by 1.5x and cropped to 680$\times$480p. so that it cannot view the panels if they are removed (The top camera view that is input to the policy is the same as in Appendix~\ref{app:test_scene_reference_images}). Additionally, the box is lit up well enough so that the outside of the panels is completely dark from the camera exposure difference.
      \item In our methods, we keep the $-x$ face on the box and remove the $+x$ face Velcro panel in order to switch out objects during data capture and policy evaluation.
  \end{itemize}
  \item The \textbf{front-view camera} is mounted on the internal edge where
    the $+z$ and $+y$ faces intersect (see Fig.~\ref{fig:env_overview}).
\end{itemize}

\paragraph{Assembly steps.}

\begin{figure}[h]
  \centering
  \begin{subfigure}[b]{0.30\linewidth}
    \includegraphics[width=\linewidth]{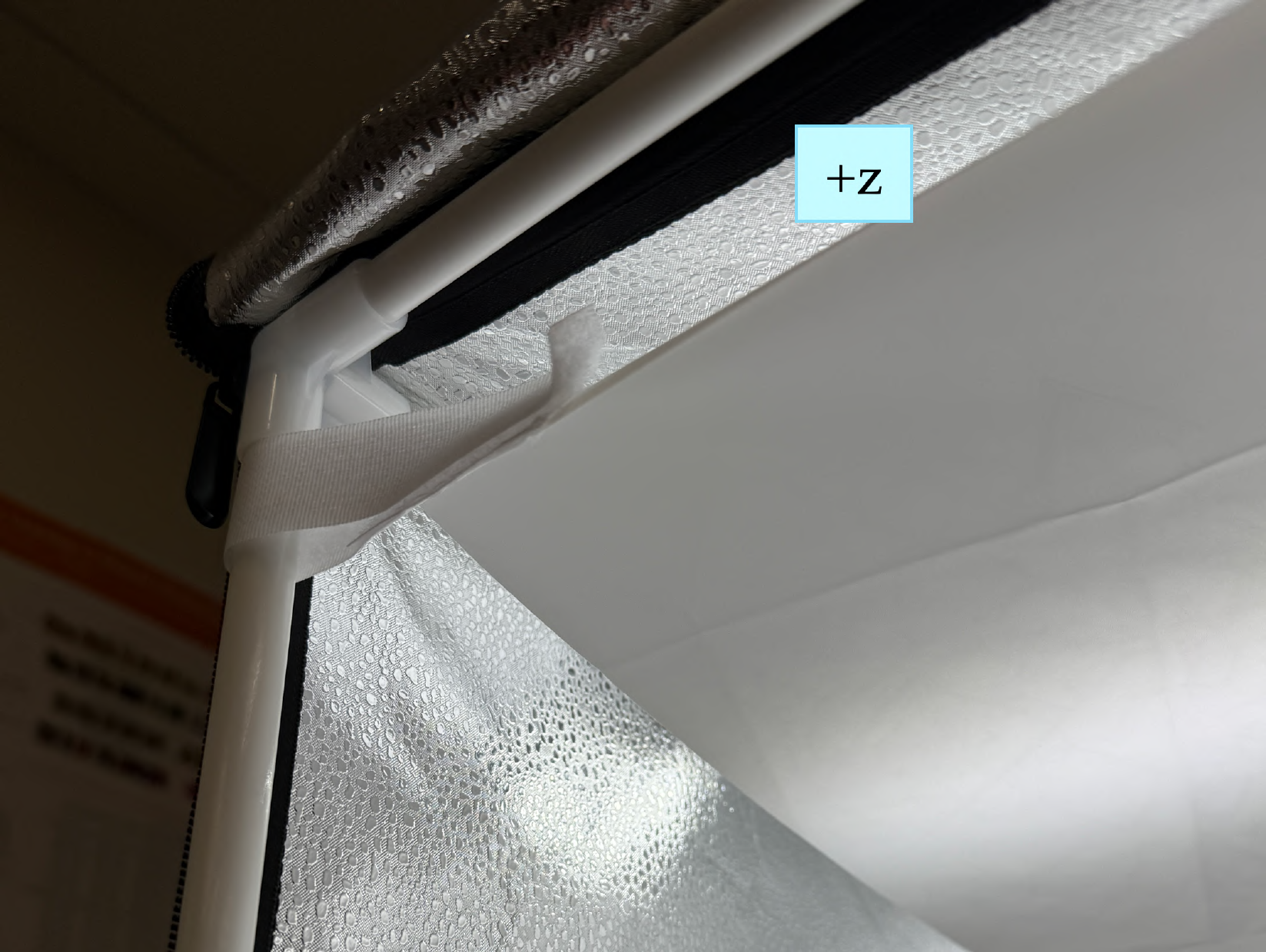}
    \caption{Diffuser sheet}
  \end{subfigure}\hfill
  \begin{subfigure}[b]{0.30\linewidth}
    \includegraphics[width=\linewidth]{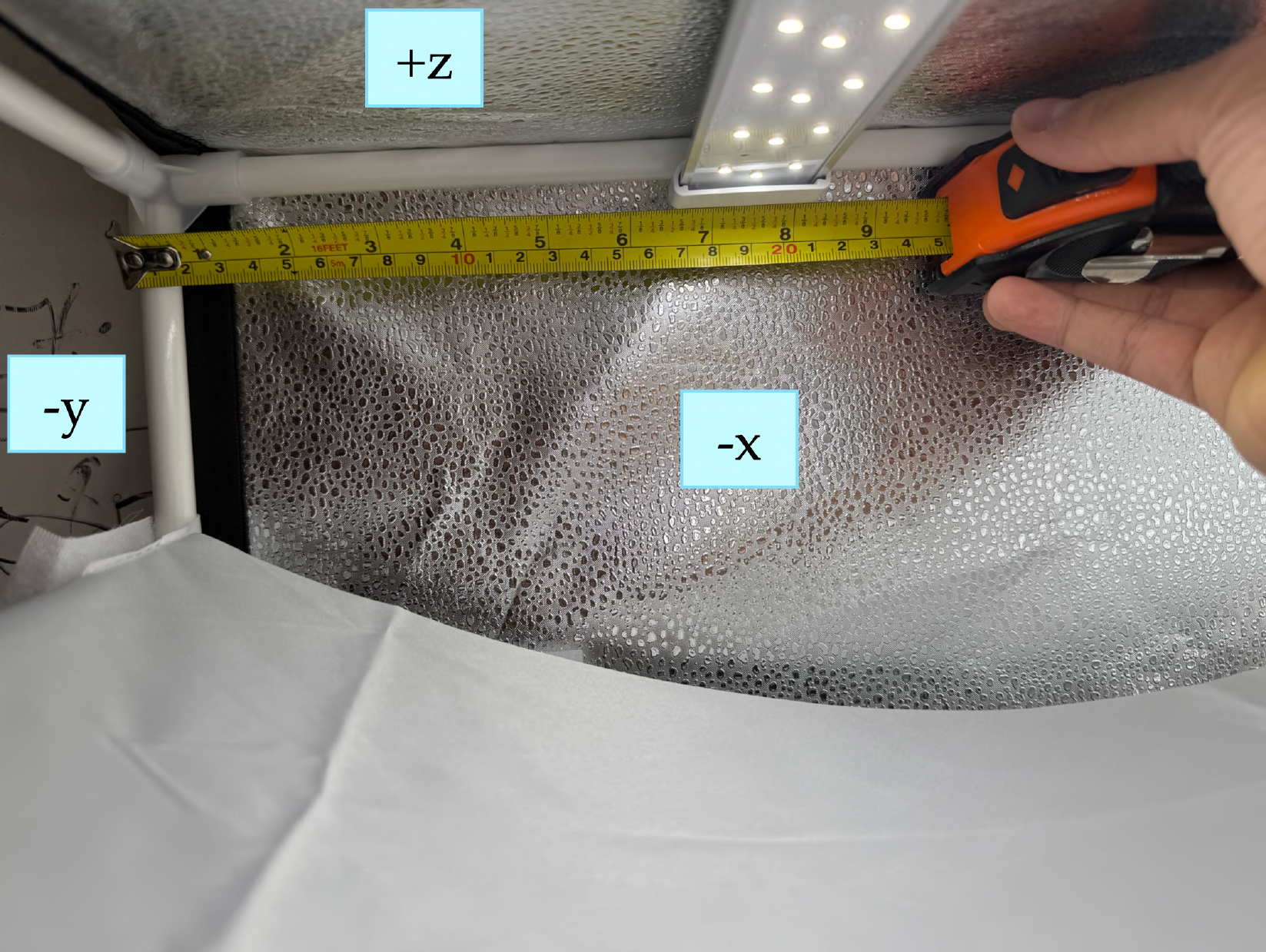}
    \caption{LED panel positions.}
  \end{subfigure}\hfill
  \begin{subfigure}[b]{0.30\linewidth}
    \includegraphics[width=\linewidth]{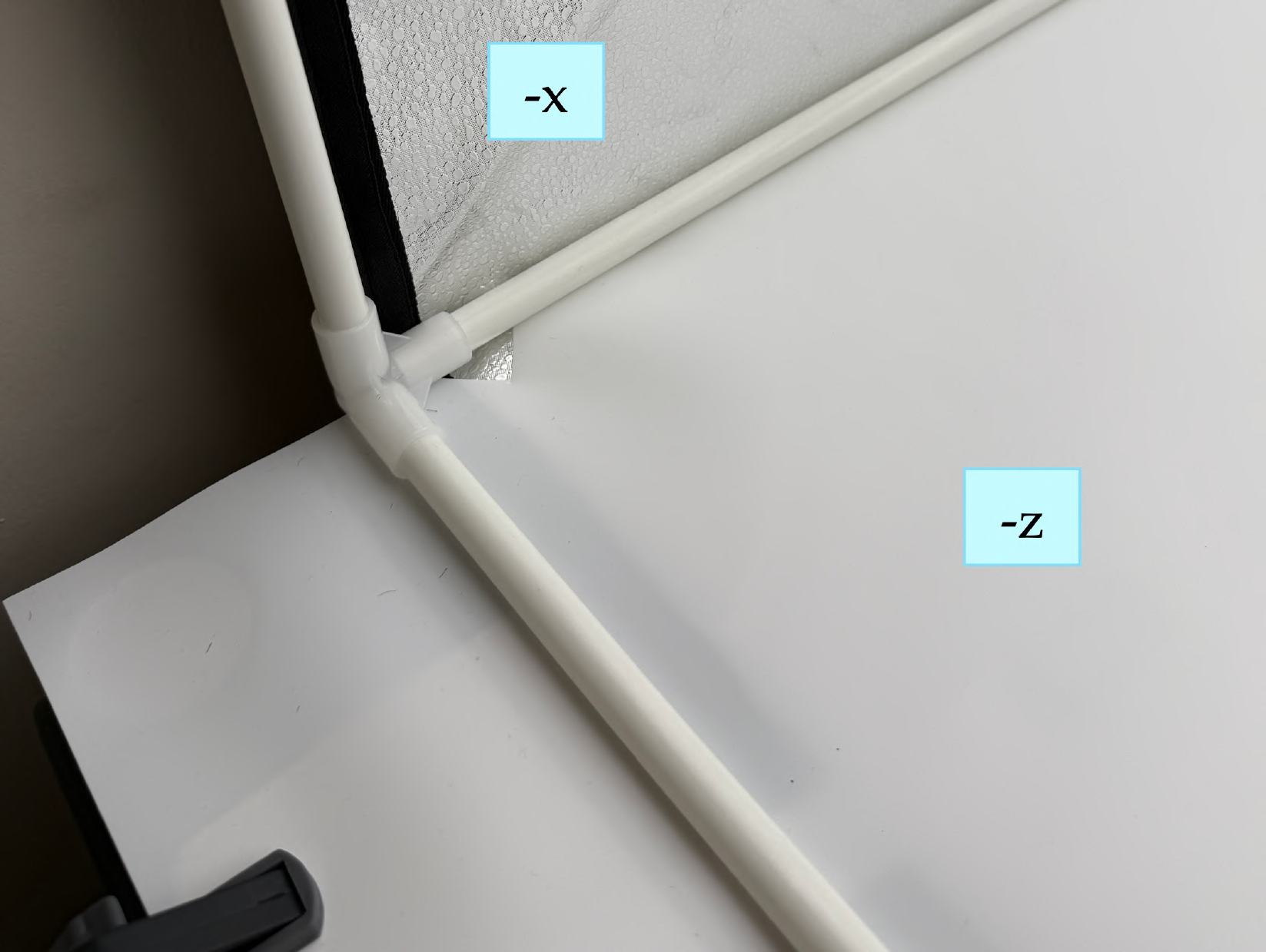}
    \caption{Background sheet.}
  \end{subfigure}
  \caption{\textbf{Light box assembly.}}
  \label{fig:box_assembly}
\end{figure}

\begin{enumerate}
  \item Construct the cube-shaped \textbf{PVC frame} using the 12 pipes and 8 edge
    connectors supplied with the Glendan kit.
    \textbf{Do not attach the zipper tarp yet}; complete all internal
    installations first.

  \item \textbf{Attach the white light diffuser sheet} to the top ($+z$) face
    of the frame using the supplied velcro strips.
    Secure the sheet using the velcro strips on both the $-y$ and $+y$ pipes, as close as possible to the $+z$ face.

    (Fig.~\ref{fig:box_assembly}(a)).

  \item \textbf{Attach the provided white hooks onto the LED panels} and then \textbf{mount the three LED panels} on the PVC frame before fitting
    the tarp (Fig.~\ref{fig:box_assembly}(b)):
    \begin{enumerate}
      \item \emph{LED Panel~1}: $+z$ face, pointed downward toward $-z$;
        center $\approx$ 7.5 inches from the $-y$ face.
      \item \emph{LED Panel~2}: $+z$ face, pointed downward toward $-z$;
        center $\approx$ 7.5 inches from the $+y$ face
        (mirror image of Strip~1).
      \item \emph{LED Panel~3}: $+y$ face, pointed toward $-y$;
        center $\approx$ 8 inches from the $+z$ face.
      \item \emph{Note}: Connect the power cables of each LED panel to the LED power supply after putting the tarp on the box frame in the next step. Feed cables through the small hole in the tarp near the top of the $-x$ or $+x$ faces.
    \end{enumerate}

  \item \textbf{Mount the front-view camera} on the PVC frame:
    \begin{enumerate}
      \item Connect the USB-C cable to the RealSense camera. (Note: Feed the computer-side of the cable through the small hole in the tarp near the top of the $-x$ or $+x$ faces after the tarp is placed over the frame in the next step).
      \item Snap the camera mount hooks onto the PVC pipe at the junction
        of the $+z$ and $+y$ inner faces.
      \item Position the camera's midpoint
        $\approx$ 17.5 inches from the $-x$ face.
        The camera angle should be approximately
        $-50$\textdegree\ to $-60$\textdegree\ from horizontal;
        fine calibration is performed in Section~\ref{subsec:calib}.
    \end{enumerate}

  \item Slide the \textbf{zipper tarp} over the completed PVC/LED frame, ensuring that the $-y$ face of the frame matches the side of the tarp with the zippers. Ensure that the $-z$ side (the object workspace) is actually on the bottom.

  \item \textbf{Attach the white PP background sheet} to the inner $+y$
    face using the velcro strips on the tarp and the sheet
    (Fig.~\ref{fig:box_assembly}(c)):
    \begin{enumerate}
      \item Tuck the sheet \emph{under} the $-z$ PVC pipes so that the
        workspace is as flat as possible.
        Fold the sides of the sheet under the pipes if necessary to flatten the $-z$ workspace.
      \item If the sheet bunches near the $-y$ face, cut a small triangular
        slit near the $-y$ edge to relieve the tension.
      \item The sheet should extend $\approx$ 3.5 inches beyond the
        $-y$ open face of the box.
      \item Clamp the overhanging portion of the sheet to the table on both sides.
    \end{enumerate}
\end{enumerate}

\noindent\textbf{Important Checklist:}
\begin{todolist}
  \item LED panels are positioned as specified; all power connections are
    secure; the lights can be turned on to the maximum brightness (around 5600K). 
  \item The diffuser sheet is taut with no visible sag.
  \item The background sheet is flat against the workspace; no curves or
    wrinkles are visible from the camera view.
  \item The zipper tarp fully covers all PVC pipes and LED wiring.
\end{todolist}

\subsection{Mount the SO-101 Follower Arm and Place AprilTag}
\label{subsec:SO-101_Setup}

\begin{figure}[h]
  \centering
  \begin{subfigure}[b]{0.45\linewidth}
    \includegraphics[width=\linewidth]{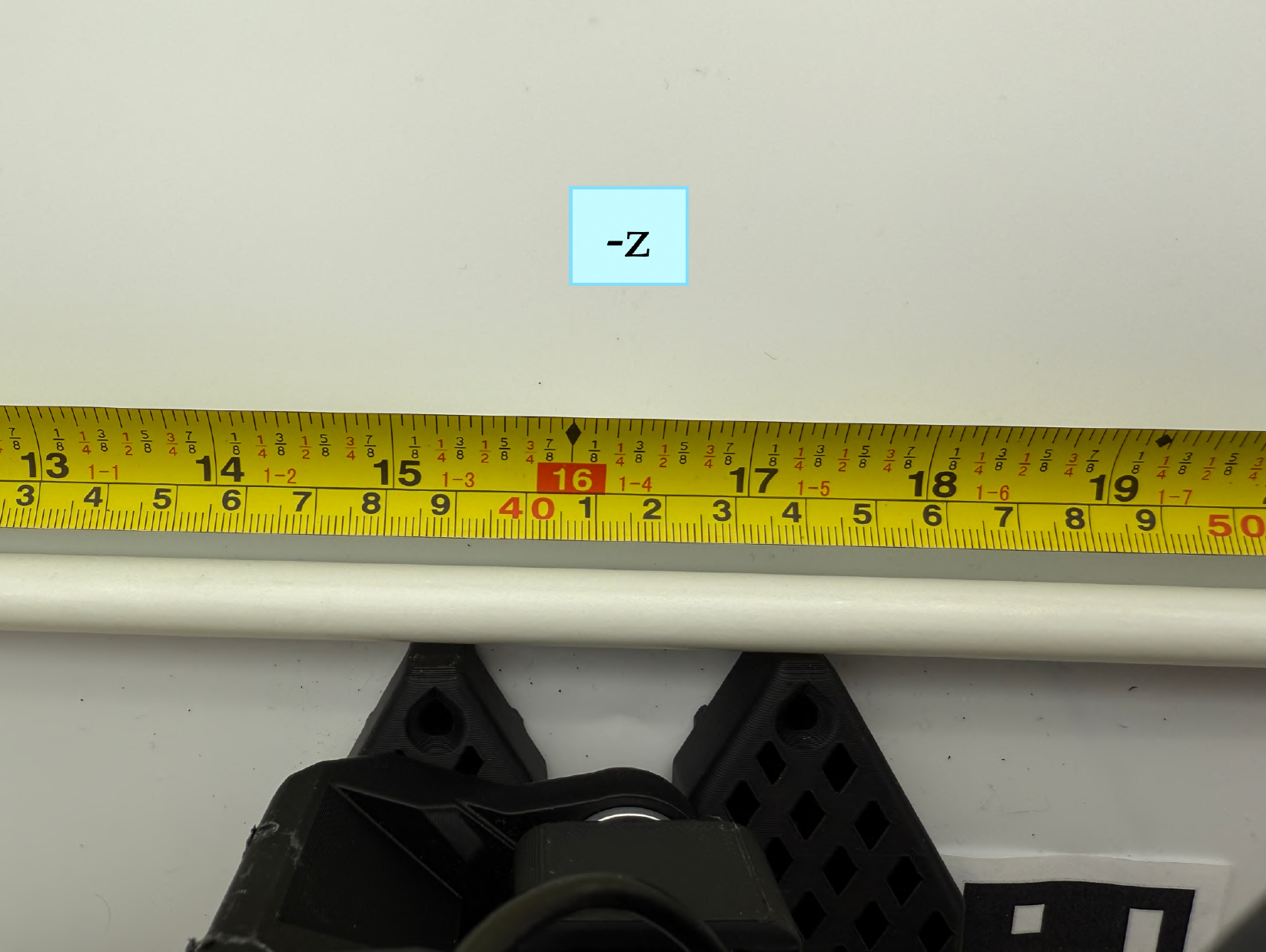}
    \caption{SO-101 arm clamped to table.
      Center of base is \SI{16.5} inch from the $-x$ face.}
  \end{subfigure}\hfill
  \begin{subfigure}[b]{0.45\linewidth}
    \includegraphics[width=\linewidth]{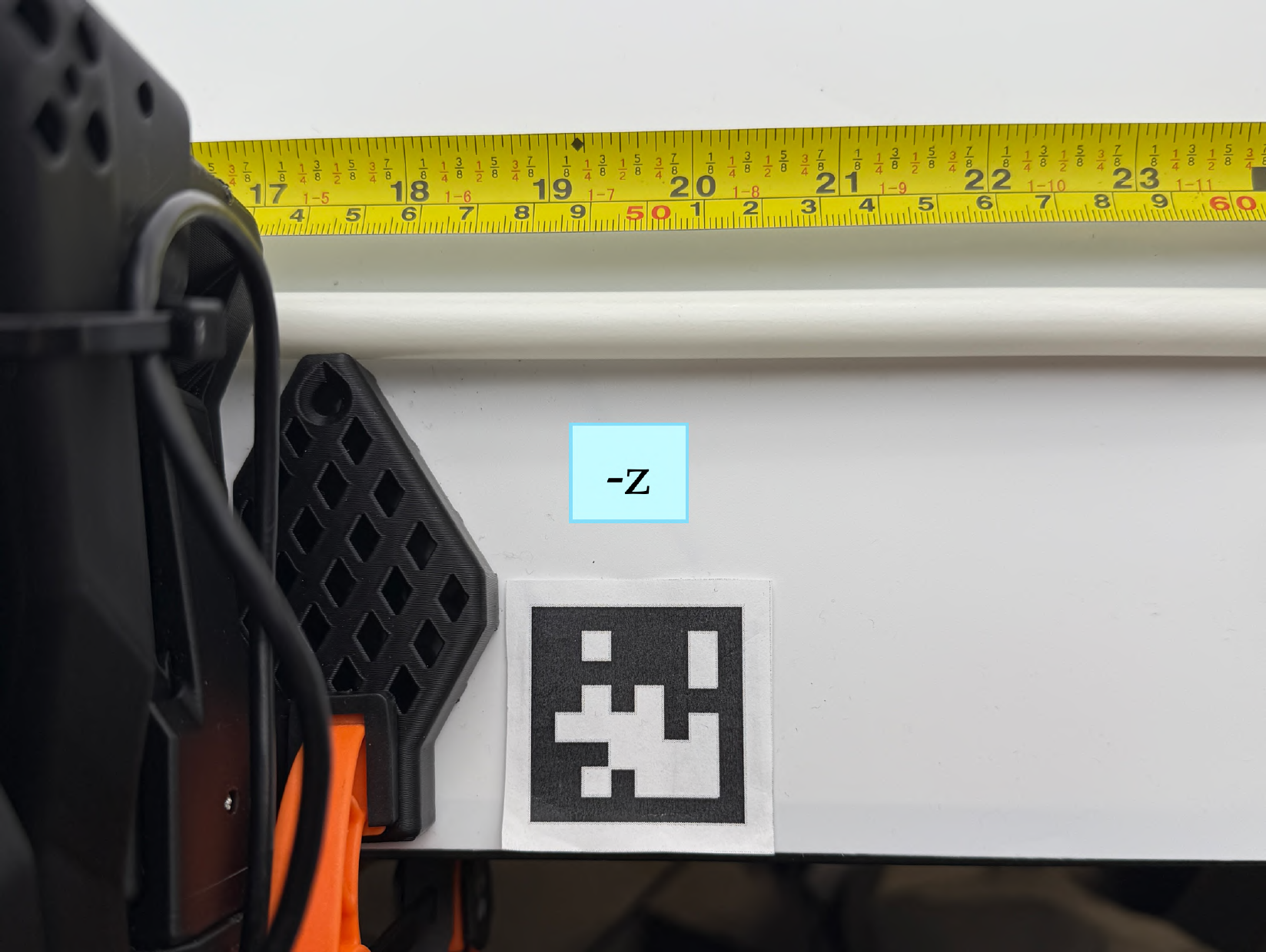}
    \caption{AprilTag aligned with the base edges.\newline}
  \end{subfigure}
  \caption{\textbf{SO-101 arm placement and AprilTag positioning.}}
  \label{fig:arm_apriltag}
\end{figure}

We use a mixture of AprilTag camera calibration and image overlay matching to ensure consistent camera viewing angles the top view camera (Fig.~\ref{fig:arm_apriltag}).

\begin{enumerate}
  \item Clamp both sides of the \textbf{SO-101 follower arm} to the edge of the table so
    that the front edge of its base touches the PVC pipe running between
    the $-z$ and $-y$ faces of the box.
    The center of the SO-101 base should be $\approx$ 16.5 inches from the $-x$ face
    (Fig.~\ref{fig:arm_apriltag}(a)).

  \item Attach the \textbf{12\,V power adaptor} to the SO-101 arm.

  \item Place the \textbf{4\,cm AprilTag} on the \emph{right} side of the
    SO-101 base (Fig.~\ref{fig:arm_apriltag}(b)):
    \begin{enumerate}
      \item The \emph{northwest corner} of the tag must touch the vertical
        edge of the SO-101 base.
      \item The \emph{south black border} of the tag must be aligned with
        the bottom edge of the SO-101 base.
    \end{enumerate}

  \item Double-check the tag orientation.
    Wrinkled paper causes unreliable detection;
    affix it flat using double-sided tape on all four corners.
\end{enumerate}

\noindent\textbf{Important Checklist:}
\begin{todolist}
  \item The SO-101 base center is \SI{16.5} inch from the $-x$ box face.
  \item The base front edge is flush against the $-z$/$-y$ PVC pipe.
  \item The AprilTag northwest corner touches the vertical edge of the base;
    the south border aligns with the base bottom edge.
  \item The tag is flat with no wrinkles or lifted corners.
\end{todolist}

\subsection{Install Software}
\label{subsec:Install_software}


Before proceeding to calibration, ensure the following software is installed
on the host computer:

\begin{enumerate}

  \item Clone the benchmark repository, create a new Conda environment, and install dependencies:
    \begin{verbatim}
git clone https://github.com/IRVLUTD/VLAReplica.git
cd VLAReplica
conda env create -f environment.yml
conda activate vlareplica
    \end{verbatim}
  \item Find available cameras indices with the command (note down the numbers):
    \begin{verbatim}
lerobot-find-cameras
    \end{verbatim}
    Record the camera indices for the two cameras.
  \item Find USB device serial ports from the following command: 
    \begin{verbatim}
lerobot-find-port
    \end{verbatim}
    Then unplug the SO-101 USB cable from the computer, and press \texttt{Enter}. The terminal will output something like \texttt{/dev/ttyACM1}.
    Record the serial port for the follower arm.
\end{enumerate}

\subsection{Calibrate the SO-101 arm}

Next, calibrate the SO-101 follower according to the LeRobot Docs (\url{https://huggingface.co/docs/lerobot/so101?setup_motors=Command#calibrate}). \textbf{Follow the video carefully, and ensure each motor is at the middle position before starting the calibration process.} During calibration, thoroughly rotate each of the six motors to their physical joint limits.

(If you can't find the USB port address, use the command \texttt{lerobot-find-port})
\begin{enumerate}
    \item Locate the calibration file that LeRobot saved to your device. It should be under: \begin{verbatim} 
~/.cache/huggingface/lerobot/calibration/robots/<your-robot-id>
    \end{verbatim} in your root folder.
    \item Copy this .json file to: \texttt{VLAReplica/calibration/robots/so101\_follower}
    \item And rename that file to: \texttt{so101\_follower\_arm.json}
\end{enumerate}

\subsection{Camera Calibration}
\label{subsec:calib}

We provide a calibration script that detects the AprilTag and reports the
camera pose in real time, allowing fine adjustment of the camera mount
before locking it in place.
The target pose values are listed in Table~\ref{tab:cam_pose}.

\begin{table}[h]
  \centering
  \caption{Target front-view camera pose relative to the AprilTag.
    Adjust until all values match within the specified tolerance.}
  \label{tab:cam_pose}
  \begin{tabular}{cccccc}
    \toprule
    $x$\,(m) & $y$\,(m) & $z$\,(m)
             & $R$\,(deg) & $P$\,(deg) & $Y$\,(deg) \\
    \midrule
    $-0.06\pm0.01$ & $-0.39\pm0.01$ & $1.25\pm0.01$ & $-18.5\pm1.0$ & $3.0\pm1.0$ & $2.5\pm1.0$ \\
    \bottomrule
  \end{tabular}
\end{table}

\begin{figure}[h]
  \centering
  \includegraphics[width=0.9\linewidth]{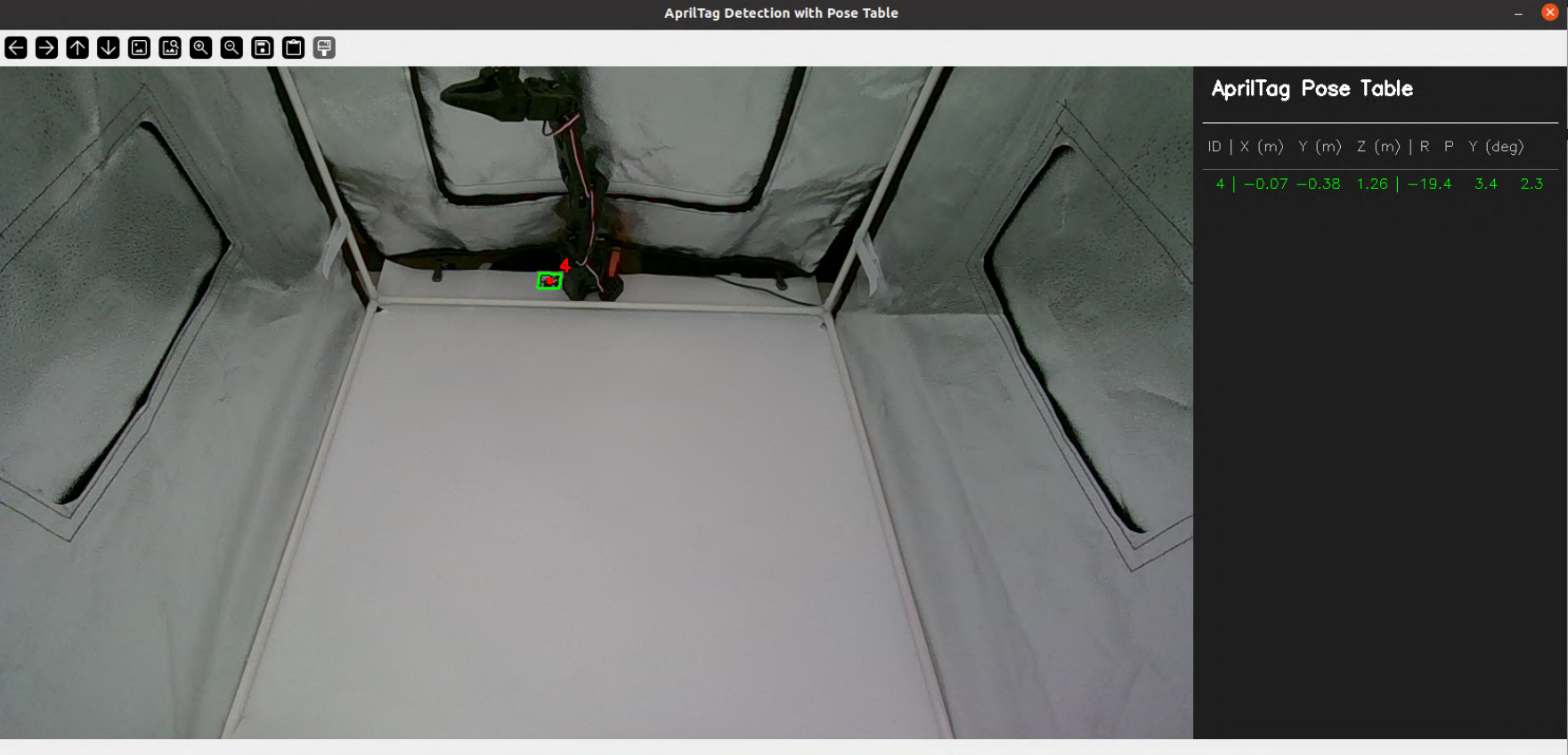}
  \caption{\textbf{AprilTag camera calibration GUI.}
    The live camera feed (\emph{left}) and the detected AprilTag pose table
    (\emph{right}) are shown simultaneously.
    Adjust the camera position until the pose values match
    Table~\ref{tab:cam_pose}.
  }
  \label{fig:calibration}
\end{figure}

\begin{figure}[h]
  \centering
  \includegraphics[width=0.9\linewidth]{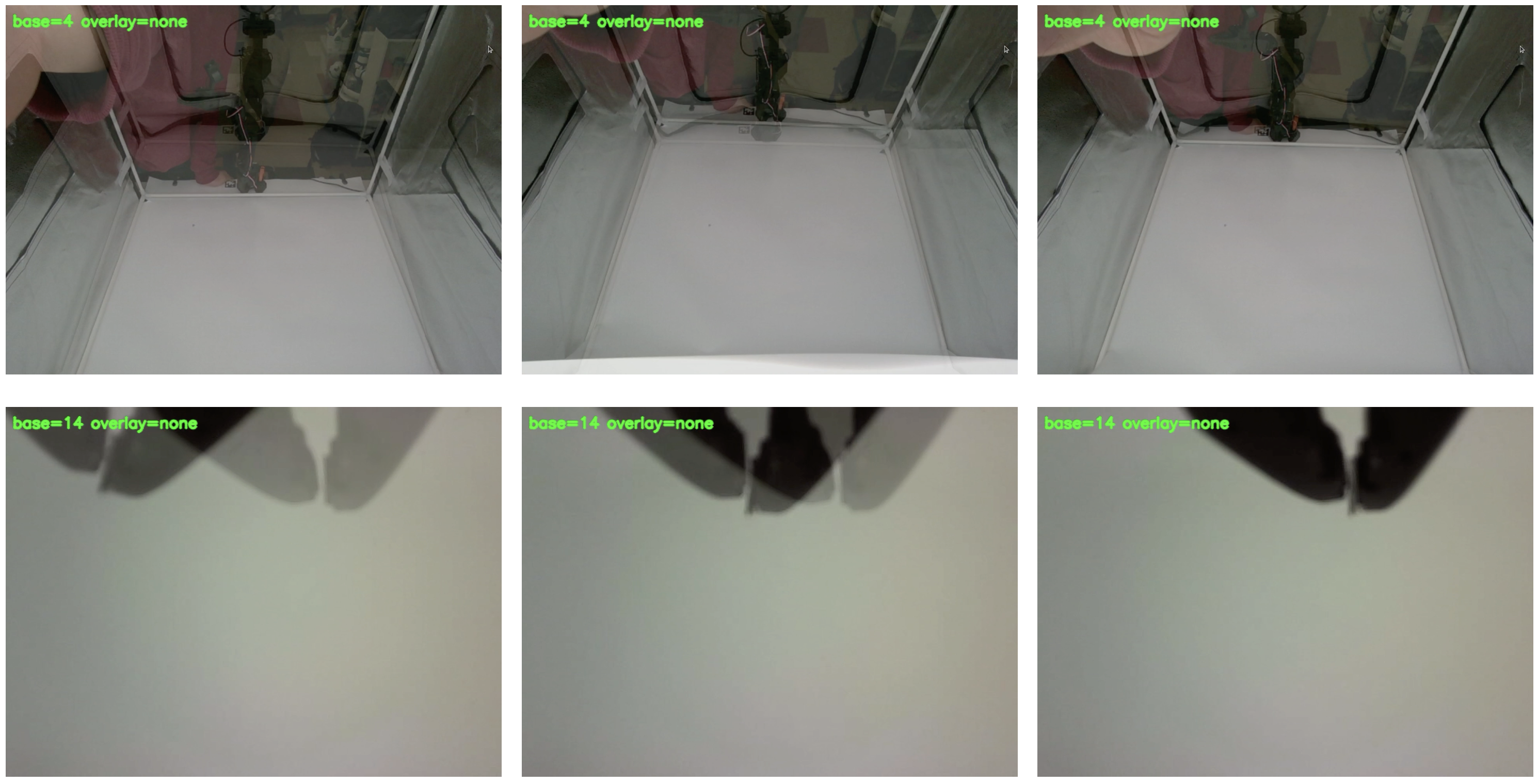}
  \caption{\textbf{Visual calibration GUI.}
    Top camera (\emph{top}) and wrist camera
    (\emph{bottom}) calibration over time.
    The cameras are adjusted physically until the overlay matches the reference image.
  }
  \label{fig:visual_calibration}
\end{figure}

\begin{enumerate}
  \item In a new terminal inside the virtual environment, run the calibration script (replace \texttt{<your-top-camera-index>} with the number you recorded in 
  Appendix~\ref{subsec:Install_software}): 
    \begin{spverbatim}
python calibration/camera/detect_apriltag.py --camera-index <your-top-camera-index>
    \end{spverbatim}

  \item A GUI window will display the live camera feed alongside the
    estimated AprilTag pose (Fig.~\ref{fig:calibration}).
    Reach into the box and physically slide or tilt the camera mount along
    the PVC pipe until all reported values match Table~\ref{tab:cam_pose}
    as closely as possible.

  \item Some error is acceptable (see Table~\ref{tab:cam_pose}). Once satisfied,
    press \texttt{q} to exit the program.

    \item Although the AprilTag pose estimator may output values close to Table~\ref{tab:cam_pose}, there may still be slight camera misalignment. To solve this, we utilize \emph{visual overlay matching} (see Fig.~\ref{fig:visual_calibration}) to ensure the camera view is as close as possible to \emph{VLA-REPLICA's} original view. 
    \begin{enumerate}
        \item First, calibrate the top camera for the second time. Run the following, replacing \texttt{your-top-camera-id} with the number you recorded in Appendix~\ref{subsec:Install_software}:
    \begin{spverbatim}
python calibration/camera/overlay.py --overlay-image-folder calibration/camera/referenceImages/top --base-cam <your-top-camera-id>
    \end{spverbatim}
    \item Use the provided GUI to match the view of your camera with the reference image by reaching into the box and  sliding or tilting the camera mount along the PVC pipe.

    \item Next, calibrate the wrist camera for the first time. Run the following, replacing \texttt{your-wrist-camera-id} with the number you recorded in Appendix~\ref{subsec:Install_software}:
    \begin{spverbatim}
python calibration/camera/overlay.py --overlay-image-folder calibration/camera/referenceImages/wrist --base-cam <your-wrist-camera-id>
    \end{spverbatim}
    \item Slightly loosen the M3 screw on the wrist camera mount on the SO-101, and use the provided GUI to match the view of your camera with the reference image by rotating the wrist camera on the end-effector.
    \end{enumerate}

\end{enumerate}

\noindent\textbf{Important Checklist:}
\begin{todolist}
  \item All six pose values ($x,y,z,R,P,Y$) match the targets in
    Table~\ref{tab:cam_pose}.
  \item Reference images and camera views match almost identically for both top and wrist cameras.
\end{todolist}


The environment setup is complete.

\subsection{Policy evaluation script \texttt{benchmark.py}}

\begin{table}[H]
\caption{Command-line flags for \texttt{benchmark.py}.}
\label{tab:benchmark_commands}
\centering
\small
\renewcommand{\arraystretch}{1.15}
\begin{tabularx}{\linewidth}{>{\raggedright\arraybackslash}p{0.35\linewidth}X}
\toprule
\textbf{Flag} & \textbf{Description} \\
\midrule
\texttt{---policy-type <model>} & Selects the policy family to evaluate. Currently supported models: \texttt{\{act,smolvla,dit,xvla,pi0,pi05\}} \\
\texttt{---policy-path <path>} & Hugging Face repo ID or local path for the policy checkpoint. \\
\texttt{---policy-from-hub} & If called, loads policy from Hugging Face Hub instead of local directory. \\
\texttt{---run-all-tasks} & Runs evaluation across all tasks from task config, instead of single task. \\
\texttt{---task-subset <ID or OOD>} & When using \texttt{---run-all-tasks}, restricts evaluation to ID or OOD task subset. \\
\texttt{---iterations <number>} & Number of evaluation iterations per task. \\
\texttt{---eval-follower-calib-dirs <path>} & Follower calibration directory. \newline (default: \texttt{calibration/robots/so101\_follower}). \\
\texttt{---eval-follower-ports <serial port>} & Serial port for the follower robot. \\
\texttt{---eval-follower-ids <id>} & Robot ID for the follower arm.
\newline (default: \texttt{so101\_follower\_arm} \\
\texttt{---eval-top-indexes <index>} & Top-camera index for the active arm. \\
\texttt{---eval-wrist-indexes <index>} & Wrist-camera index for the active arm. \\
\texttt{---reset-mode fixed} & Uses a fixed reset action instead of teleoperated leader reset. \\
\texttt{---reset-action-file <path>} & JSON file containing the normalized reset action vector required when \texttt{---reset-mode fixed} is used.
\newline (default: \texttt{arm\_reset.json}) \\
\bottomrule
\end{tabularx}
\label{tab:benchmark-flags}
\end{table}

Now you are ready to start policy evaluations. The python script \texttt{benchmark.py} allows model evaluations, with an assortment of CLI flags (Table~\ref{tab:benchmark_commands}). Example command:
\begin{verbatim}
python benchmark.py \
  --policy-type pi0 \
  --policy-path lerobot/pi0_base \
  --policy-from-hub \
  --run-all-tasks \
  --task-subset ID \
  --iterations 5 \
  --eval-follower-calib-dirs calibration/robots/so101_follower \
  --eval-follower-ports /dev/ttyACM1 \
  --eval-follower-ids so101_follower_arm \
  --eval-top-indexes 4 \
  --eval-wrist-indexes 14 \
  --reset-mode fixed \
  --reset-action-file arm_reset.json
\end{verbatim}

\begin{figure}[h]
  \centering
  \includegraphics[width=1\linewidth]{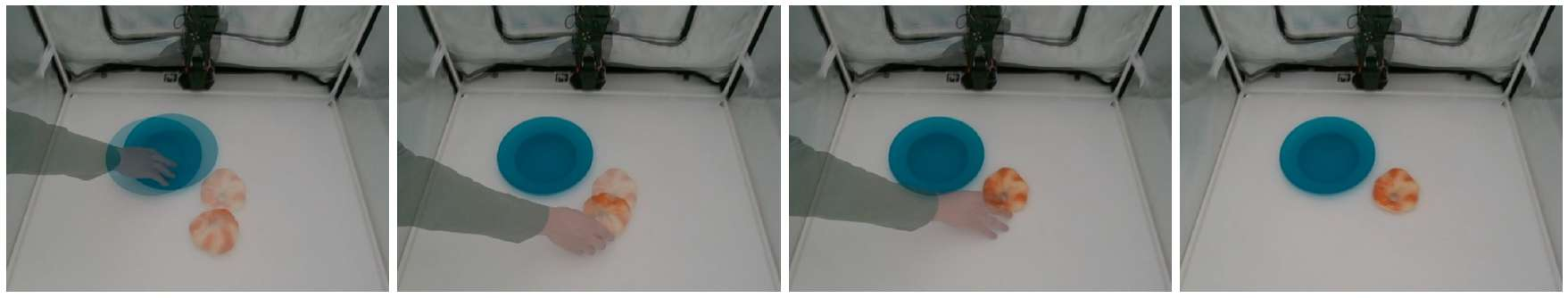}
  \caption{\textbf{\texttt{benchmark.py} live video evaluation GUI.} The user is currently setting up the scene for the \texttt{Put bread on plate} task.}
  \label{fig:benchmark_gui}
\end{figure}

After the script loads the corresponding policy and connects successfully to the followers, the follower arm will move to a start position (predefined in \texttt{arm\_reset.json}). An openCV GUI will pop up, overlaying the live video feed from the top camera with the proper test scene (i.e. object placements) for that task (Fig.~\ref{fig:benchmark_gui}).

When the live video feed and the overlay image match almost exactly, press \texttt{enter} on the keyboard to start the policy inference. The policy is given 90 seconds to complete the task before the iteration ends. If the policy completes the task before the 90 seconds, press \texttt{right arrow} to skip to the setup phase of the next iteration. The SO-101 arm will reset back to the start position. Evaluate each rollout as \textbf{success} or \textbf{failure}. Refer to success criterion in Appendix~\ref{app:task_variant_list}.

\noindent\textbf{Important Checklist:}
\begin{todolist}
  \item Ensure SO-101 arm works properly.
  \item Run a policy and ensure proper behavior.
\end{todolist}

\section{Dataset Objects}
\label{app:objects}
\vspace{-1mm}
\begin{table}[H]
\begin{center}
\caption{Bill of Materials for the objects used in policy evaluation.}
\label{}
\begin{tabular}{lllc}
\toprule
\textbf{Object}& \textbf{Price (USD)}  &\textbf{Variations of object used}& \textbf{Link} \\
\midrule
Button & 9.99  &N/A& \href{https://a.co/d/fT3YDCG}{Amazon} \\
Pen cup & 5.99  &N/A& \href{https://a.co/d/6ciSG6D}{Amazon} \\
Colored towels & 7.49  &Pink, yellow, blue& \href{https://a.co/d/8HAiPb0}{Amazon} \\
Oven & 34.98  &N/A& \href{https://a.co/d/ceA62jv}{Amazon} \\
Mug coasters & 13.98  &Green, orange, purple, yellow& \href{https://a.co/d/eOCyPa9}{Amazon} \\
Colored bowls & 12.99  &Red, blue, yellow, green& \href{https://a.co/d/cZWS2Vl}{Amazon} \\
White board \& marker & 9.75  &N/A& \href{https://a.co/d/56kRyPI}{Amazon} \\
Toy fruit set & 15.99  &Apple, yellow pear& \href{https://a.co/d/dFtjtPY}{Amazon} \\
Colored plates & 19.99  &Red, blue& \href{https://a.co/d/2PmPK7R}{Amazon} \\
Spoon \& fork set & 7.99  &N/A& \href{https://a.co/d/9bqCtz3}{Amazon} \\
Toy bread set & 27.99  &Croissant (A), small bun (B), donut (C)& \href{https://a.co/d/dKcH6hA}{Amazon} \\
Tissue boxes & 7.49  &Green box& \href{https://a.co/d/gQueIUQ}{Amazon} \\
Black pepper & 2.49  &N/A& \href{https://a.co/d/drrxkYU}{Amazon} \\
Colored pencil box & 14.99  &Blue, yellow, pink& \href{https://a.co/d/4lTryil}{Amazon} \\
Colored blocks & 16.99  &Red, blue, yellow, green, orange, purple& \href{https://a.co/d/0aG0ONjL}{Amazon} \\
Whiteboard eraser & 3.12  &N/A& \href{https://a.co/d/25n9EWn}{Amazon} \\
\midrule
\textbf{Total:} & \textbf{212.21}  && \\
\bottomrule
\end{tabular}
\end{center}
\end{table}

\newpage

\vspace{-3mm}
\section{Task Variant List \& Success Criterion}
\label{app:task_variant_list}

\noindent\textbf{Numbering convention.} The task-family numbers in Appendices~\ref{app:task_variant_list}--\ref{app:test_scene_reference_images} follow the released scene-reference order. The main-paper tables group tasks by behavior: main-paper Tasks 1--3 are scene families 1--3, main-paper Task 4 (block collection) is scene family 10, and main-paper Tasks 5--10 are scene families 4--9. Task names identify the same behaviors throughout; scene variant IDs 1--90 remain unchanged.

Each task has 5 in-distribution (ID) and 5 out-of-distribution (OOD) variants (except Task 5 \& 6, which do not have OOD tasks). All 90 task scenes are listed below. \emph{Level difficulty} refers to the number of distractor objects in frame, and the target objects' positional deviations from the training set (Appendix~\ref{app:demonstration_dataset_object_placements}). The success criteria for each task (i.e. what counts as success) is also listed.
\setlength\LTleft{0pt}
\setlength\LTright{0pt}

\setlength\LTleft{0pt}
\setlength\LTright{0pt}

\begin{longtable}[c]{@{}clp{6.8cm}c p{2.6cm}@{}}
\caption{In-Distribution (ID) and Out-of-Distribution (OOD) Task Variants} \\
\toprule
\textbf{ID\#} & \textbf{Level} & \textbf{Task Variant} & \textbf{Dist.} & \textbf{Success Criterion} \\
\midrule
\endfirsthead
\toprule
\textbf{ID\#} & \textbf{Level} & \textbf{Task Description} & \textbf{Dist.} & \textbf{Success Criterion} \\
\midrule
\endhead
\bottomrule
\endfoot

\multicolumn{5}{@{}l}{\textbf{Task 1: Put Bread on Plate}} \\*
1  & Easy   & Put A bread on the red plate                & ID  & \multirow{10}{=}{Bread is resting on the correct colored plate, and SO-101 moved back to its home position.} \\*
2  & Easy   & Put B bread on the blue plate               & ID  & \\*
3  & Medium & Put A bread on the blue plate               & ID  & \\*
4  & Medium & Put B bread on the red plate (1 distractor) & ID  & \\*
5  & Hard   & Put A bread on the red plate (2 distractors) & ID & \\*
\cellcolor{shadecolor}6  & \cellcolor{shadecolor}Easy   & \cellcolor{shadecolor}Put C bread on the red plate           & \cellcolor{shadecolor}OOD & \\*
\cellcolor{shadecolor}7  & \cellcolor{shadecolor}Easy   & \cellcolor{shadecolor}Put C bread on the blue plate          & \cellcolor{shadecolor}OOD & \\*
\cellcolor{shadecolor}8  & \cellcolor{shadecolor}Medium & \cellcolor{shadecolor}Put A bread on the yellow plate        & \cellcolor{shadecolor}OOD & \\*
\cellcolor{shadecolor}9  & \cellcolor{shadecolor}Medium & \cellcolor{shadecolor}Put B bread on the yellow plate        & \cellcolor{shadecolor}OOD & \\*
\cellcolor{shadecolor}10 & \cellcolor{shadecolor}Hard   & \cellcolor{shadecolor}Put C bread on the yellow plate        & \cellcolor{shadecolor}OOD & \\

\midrule
\multicolumn{5}{@{}l}{\textbf{Task 2: Put Bowl on Coaster}} \\*
11 & Easy   & Put red bowl on green coaster               & ID  & \multirow{10}{=}{Correct bowl is touching/resting on correct colored coaster, and SO-101 moved back to its home position.} \\*
12 & Easy   & Put blue bowl on orange coaster             & ID  & \\*
13 & Medium & Put blue bowl on green coaster              & ID  & \\*
14 & Medium & Put red bowl on purple coaster (1 distractor) & ID & \\*
15 & Hard   & Put yellow bowl on purple coaster (2 distractors) & ID & \\*
\cellcolor{shadecolor}16 & \cellcolor{shadecolor}Easy   & \cellcolor{shadecolor}Put blue bowl on purple coaster   & \cellcolor{shadecolor}OOD & \\*
\cellcolor{shadecolor}17 & \cellcolor{shadecolor}Easy   & \cellcolor{shadecolor}Put red bowl on orange coaster    & \cellcolor{shadecolor}OOD & \\*
\cellcolor{shadecolor}18 & \cellcolor{shadecolor}Medium & \cellcolor{shadecolor}Put yellow bowl on green coaster  & \cellcolor{shadecolor}OOD & \\*
\cellcolor{shadecolor}19 & \cellcolor{shadecolor}Medium & \cellcolor{shadecolor}Put yellow bowl on orange coaster & \cellcolor{shadecolor}OOD & \\*
\cellcolor{shadecolor}20 & \cellcolor{shadecolor}Hard   & \cellcolor{shadecolor}Put green bowl on yellow coaster  & \cellcolor{shadecolor}OOD & \\

\midrule
\multicolumn{5}{@{}l}{\textbf{Task 3: Stack Blocks}} \\*
21 & Easy   & Stack red block on blue block               & ID  & \multirow{10}{=}{Correctly colored top block has touched/rested on correctly colored bottom block for more than 2 seconds.} \\*
22 & Easy   & Stack yellow block on blue block            & ID  & \\*
23 & Medium & Stack blue block on yellow block            & ID  & \\*
24 & Medium & Stack blue block on red block (1 distractor) & ID & \\*
25 & Hard   & Stack red block on yellow block (2 distractors) & ID & \\*
\cellcolor{shadecolor}26 & \cellcolor{shadecolor}Easy   & \cellcolor{shadecolor}Stack yellow block on red block       & \cellcolor{shadecolor}OOD & \\*
\cellcolor{shadecolor}27 & \cellcolor{shadecolor}Easy   & \cellcolor{shadecolor}Stack blue block on blue block        & \cellcolor{shadecolor}OOD & \\*
\cellcolor{shadecolor}28 & \cellcolor{shadecolor}Medium & \cellcolor{shadecolor}Stack red block on green block        & \cellcolor{shadecolor}OOD & \\*
\cellcolor{shadecolor}29 & \cellcolor{shadecolor}Medium & \cellcolor{shadecolor}Stack green block on yellow block     & \cellcolor{shadecolor}OOD & \\*
\cellcolor{shadecolor}30 & \cellcolor{shadecolor}Hard   & \cellcolor{shadecolor}Stack green block on green block      & \cellcolor{shadecolor}OOD & \\

\midrule
\multicolumn{5}{@{}l}{\textbf{Task 4: Fold Towel}} \\*
31 & Easy   & Fold pink towel in half                     & ID  & \multirow{10}{=}{Correctly colored towel's edges are lifted and folded on itself by more than 50\%, and SO-101 moved back to its home position.} \\*
32 & Easy   & Fold yellow towel in half                   & ID  & \\*
33 & Medium & Fold pink towel in half (1 distractor)      & ID  & \\*
34 & Medium & Fold yellow towel in half (1 distractor)    & ID  & \\*
35 & Hard   & Fold pink towel in half (2 distractors)     & ID  & \\*
\cellcolor{shadecolor}36 & \cellcolor{shadecolor}Easy   & \cellcolor{shadecolor}Fold blue towel in half               & \cellcolor{shadecolor}OOD & \\*
\cellcolor{shadecolor}37 & \cellcolor{shadecolor}Easy   & \cellcolor{shadecolor}Fold blue towel in half   & \cellcolor{shadecolor}OOD & \\*
\cellcolor{shadecolor}38 & \cellcolor{shadecolor}Medium & \cellcolor{shadecolor}Fold blue towel in half (1 distractor) & \cellcolor{shadecolor}OOD & \\*
\cellcolor{shadecolor}39 & \cellcolor{shadecolor}Medium & \cellcolor{shadecolor}Fold blue towel in half (2 distractors)  & \cellcolor{shadecolor}OOD & \\*
\cellcolor{shadecolor}40 & \cellcolor{shadecolor}Hard   & \cellcolor{shadecolor}Fold blue towel in half (2 distractors) & \cellcolor{shadecolor}OOD & \\

\midrule
\multicolumn{5}{@{}l}{\textbf{Task 5: Open Oven (No OOD Variants)}} \\*
41 & Easy   & Open the oven                               & ID  & \multirow{3}{=}{Oven door open for at least 2 seconds; arm returns home.} \\*
42 & Easy   & Open the oven                    & ID  & \\*
43 & Medium & Open the oven (1 distractor)                & ID  & \\*
44 & Medium & Open the oven (2 distractors)                   & ID  & \\*
45 & Hard   & Open the oven (2 distractors)               & ID  & \\

\midrule
\multicolumn{5}{@{}l}{\textbf{Task 6: Clean Whiteboard (No OOD Variants)}} \\*
46 & Easy   & Clean whiteboard with eraser                & ID  & \multirow{5}{=}{Eraser wiped whiteboard 2 or more times, and then placed next to whiteboard.} \\*
47 & Easy   & Clean whiteboard with eraser     & ID  & \\*
48 & Medium & Clean whiteboard with eraser (1 distractor) & ID  & \\*
49 & Medium & Clean whiteboard with eraser (2 distractors)    & ID  & \\*
50 & Hard   & Clean whiteboard with eraser (2 distractors) & ID & \\

\midrule
\multicolumn{5}{@{}l}{\textbf{Task 7: Pour Pepper}} \\*
51 & Easy   & Pour 1 shake of pepper into the red plate   & ID  & \multirow{10}{=}{The exact number of pepper shakes is poured into the correct colored plate, the pepper is placed to the left of the plate, and SO-101 moved back to its home position.} \\*
52 & Easy   & Pour 3 shakes of pepper into the red plate  & ID  & \\*
53 & Medium & Pour 2 shakes of pepper (1 distractor)      & ID & \\*
54 & Medium & Pour 1 shake of pepper (1 distractor)      & ID & \\*
55 & Hard   & Pour 3 shakes of pepper (2 distractors)     & ID & \\*
\cellcolor{shadecolor}56 & \cellcolor{shadecolor}Easy   & \cellcolor{shadecolor}Pour 4 shakes of pepper into red plate & \cellcolor{shadecolor}OOD & \\*
\cellcolor{shadecolor}57 & \cellcolor{shadecolor}Easy   & \cellcolor{shadecolor}Pour 5 shakes of pepper into red plate & \cellcolor{shadecolor}OOD & \\*
\cellcolor{shadecolor}58 & \cellcolor{shadecolor}Medium & \cellcolor{shadecolor}Pour 1 shake of pepper into blue plate & \cellcolor{shadecolor}OOD & \\*
\cellcolor{shadecolor}59 & \cellcolor{shadecolor}Medium & \cellcolor{shadecolor}Pour 3 shakes of pepper into blue plate & \cellcolor{shadecolor}OOD & \\*
\cellcolor{shadecolor}60 & \cellcolor{shadecolor}Hard   & \cellcolor{shadecolor}Pour 5 shakes of pepper into blue plate & \cellcolor{shadecolor}OOD & \\

\midrule
\multicolumn{5}{@{}l}{\textbf{Task 8: Lift Bowl}} \\*
61 & Easy   & Lift green bowl one time                    & ID  & \multirow{10}{=}{Correct colored bowl lifted up and down the correct number of times, and SO-101 moved back to its home position.\newline\newline} \\*
62 & Easy   & Lift blue bowl one time                     & ID  & \\*
63 & Medium & Lift red bowl three times          & ID & \\*
64 & Medium & Lift green bowl three times (1 distractor)  & ID  & \\*
65 & Hard   & Lift blue bowl three times (2 distractors)  & ID  & \\*
\cellcolor{shadecolor}66 & \cellcolor{shadecolor}Easy   & \cellcolor{shadecolor}Lift green bowl two times              & \cellcolor{shadecolor}OOD & \\*
\cellcolor{shadecolor}67 & \cellcolor{shadecolor}Easy   & \cellcolor{shadecolor}Lift blue bowl two times               & \cellcolor{shadecolor}OOD & \\*
\cellcolor{shadecolor}68 & \cellcolor{shadecolor}Medium & \cellcolor{shadecolor}Lift red bowl two times                & \cellcolor{shadecolor}OOD & \\*
\cellcolor{shadecolor}69 & \cellcolor{shadecolor}Medium & \cellcolor{shadecolor}Lift yellow bowl two times             & \cellcolor{shadecolor}OOD & \\*
\cellcolor{shadecolor}70 & \cellcolor{shadecolor}Hard   & \cellcolor{shadecolor}Lift yellow bowl four times            & \cellcolor{shadecolor}OOD & \\

\midrule
\multicolumn{5}{@{}l}{\textbf{Task 9: Press Button}} \\*
71 & Easy   & Press button one time                       & ID  & \multirow{10}{=}{Button is touched the correct number of times, and SO-101 moved back to its home position.} \\*
72 & Easy   & Press button three times                    & ID  & \\*
73 & Medium & Press button one time (1 distractor)        & ID  & \\*
74 & Medium & Press button three times (1 distractor)     & ID  & \\*
75 & Hard   & Press button three times (2 distractors)    & ID  & \\*
\cellcolor{shadecolor}76 & \cellcolor{shadecolor}Easy   & \cellcolor{shadecolor}Press button two times                 & \cellcolor{shadecolor}OOD & \\*
\cellcolor{shadecolor}77 & \cellcolor{shadecolor}Easy   & \cellcolor{shadecolor}Press button four times (1 distractor)       & \cellcolor{shadecolor}OOD & \\*
\cellcolor{shadecolor}78 & \cellcolor{shadecolor}Medium & \cellcolor{shadecolor}Press button two times (2 distractors)        & \cellcolor{shadecolor}OOD & \\*
\cellcolor{shadecolor}79 & \cellcolor{shadecolor}Medium & \cellcolor{shadecolor}Press button four times (2 distractors)       & \cellcolor{shadecolor}OOD & \\*
\cellcolor{shadecolor}80 & \cellcolor{shadecolor}Hard   & \cellcolor{shadecolor}Press button five times (2 distractors)       & \cellcolor{shadecolor}OOD & \\

\midrule
\multicolumn{5}{@{}l}{\textbf{Task 10: Collect Blocks}} \\*
81 & Easy   & Collect 2 blocks into blue box              & ID  & \multirow{10}{=}{All the blocks on the workspace are placed into the correct colored box, and SO-101 moved back to its home position.} \\*
82 & Easy   & Collect 2 blocks into yellow box            & ID  & \\*
83 & Medium & Collect 3 blocks into blue box (1 distractor)     & ID & \\*
84 & Medium & Collect 3 blocks into yellow box & ID & \\*
85 & Hard   & Collect 4 blocks into blue box (2 distractors)     & ID & \\*
\cellcolor{shadecolor}86 & \cellcolor{shadecolor}Easy   & \cellcolor{shadecolor}Collect 2 blocks into pink box         & \cellcolor{shadecolor}OOD & \\*
\cellcolor{shadecolor}87 & \cellcolor{shadecolor}Easy   & \cellcolor{shadecolor}Collect 3 blocks into pink box         & \cellcolor{shadecolor}OOD & \\*
\cellcolor{shadecolor}88 & \cellcolor{shadecolor}Medium & \cellcolor{shadecolor}Collect 5 blocks into blue box         & \cellcolor{shadecolor}OOD & \\*
\cellcolor{shadecolor}89 & \cellcolor{shadecolor}Medium & \cellcolor{shadecolor}Collect 5 blocks into yellow box       & \cellcolor{shadecolor}OOD & \\*
\cellcolor{shadecolor}90 & \cellcolor{shadecolor}Hard   & \cellcolor{shadecolor}Collect 5 blocks into pink box         & \cellcolor{shadecolor}OOD & \\

\end{longtable}
\vspace{-4mm}

\section{Demonstration Dataset Object Placements}
\label{app:demonstration_dataset_object_placements}

For all 500 expert demonstrations of the 10 tasks, we manually segment the target object(s) and record the position of each target object in its initial position by calculating the center of its bounding box (represented by a colored dot). We superimpose all the dots together by task to create 10 \emph{placement images}, which we utilized to aid us in creating the test scene images (see Appendix~\ref{app:test_scene_reference_images}). This way, we could ensure that no initial object positions in the test scenes were too close to the initial object positions in the expert demonstration dataset. We provide a key next to each placement image to help identify the objects by color.

\begin{figure}[H]
  \centering
  \imgset{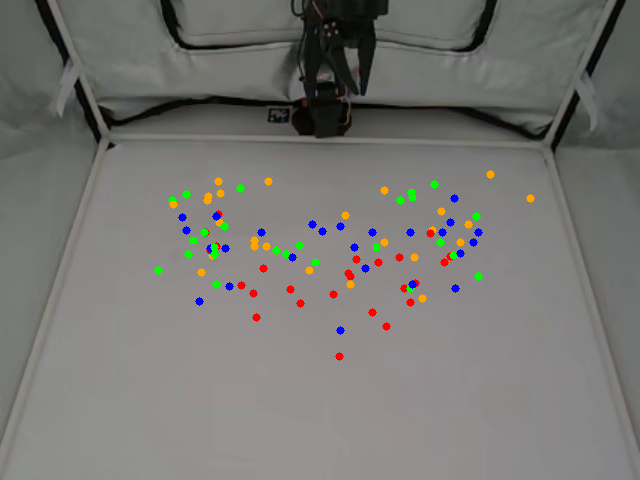}{%
    \legenddot{Red}         & Red plate \\
    \legenddot{Blue}        & Blue plate \\
    \legenddot{Green} & A bread \\
    \legenddot{Orange} & B bread \\
  }%
  \hfill
  \imgset{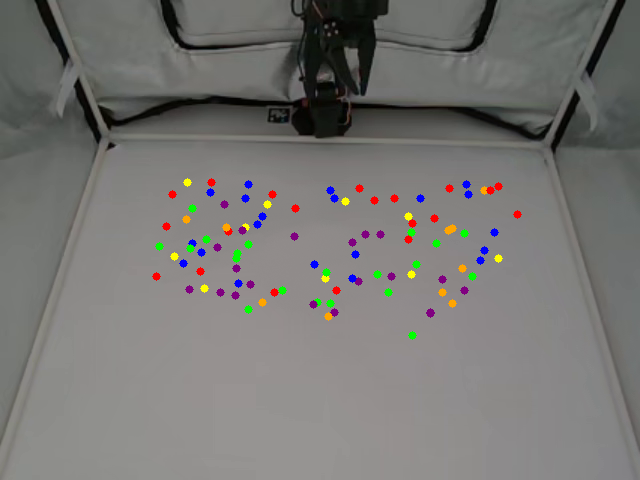}{%
    \legenddot{Red}      & Red bowl \\
    \legenddot{Blue}      & Blue bowl \\
    \legenddot{Yellow}     & Yellow bowl  \\
    \legenddot{Green}      & Green coaster \\
    \legenddot{Orange}      & Orange coaster \\

    \legenddot{Purple}        & Purple coaster \\
  }%
  \caption{Tasks 1--2: object center annotations.}
  \label{fig:centers_1_2}
\end{figure}

\begin{figure}[H]
  \centering
  \imgset{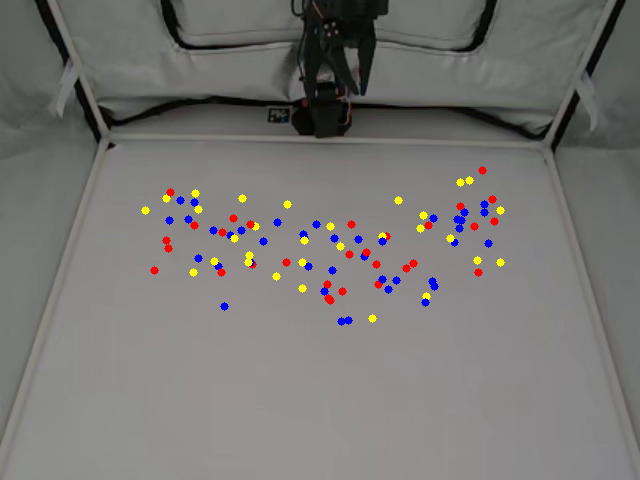}{%
    \legenddot{Red}         & Red block \\
    \legenddot{Blue}        & Blue block \\
    \legenddot{Yellow}      & Yellow block \\
  }%
  \hfill
  \imgset{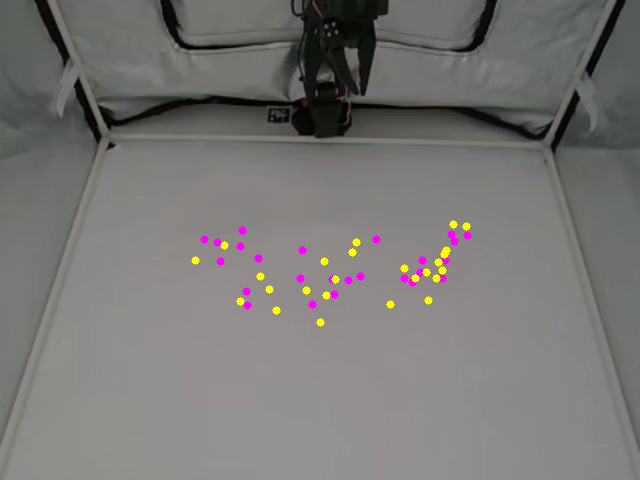}{%
    \legenddot{Magenta} & Pink towel \\
    \legenddot{Yellow}     & Yellow towel \\
  }%
  \caption{Tasks 3--4: object center annotations.}
  \label{fig:centers_3_4}
\end{figure}

\begin{figure}[H]
  \centering
  \imgset{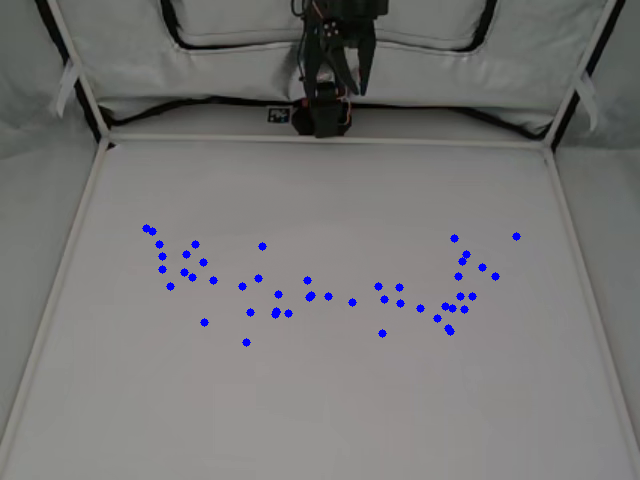}{%
    \legenddot{Blue}         & Oven \\
  }%
  \hfill
  \imgset{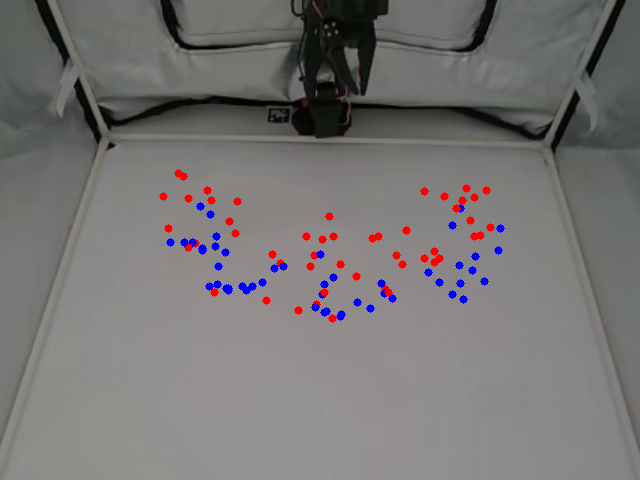}{%
    \legenddot{Red}        & Eraser \\
    \legenddot{Blue}      & Whiteboard \\
  }%
  \caption{Tasks 5--6: object center annotations.}
  \label{fig:centers_5_6}
\end{figure}
 
\begin{figure}[H]
  \centering
  \imgset{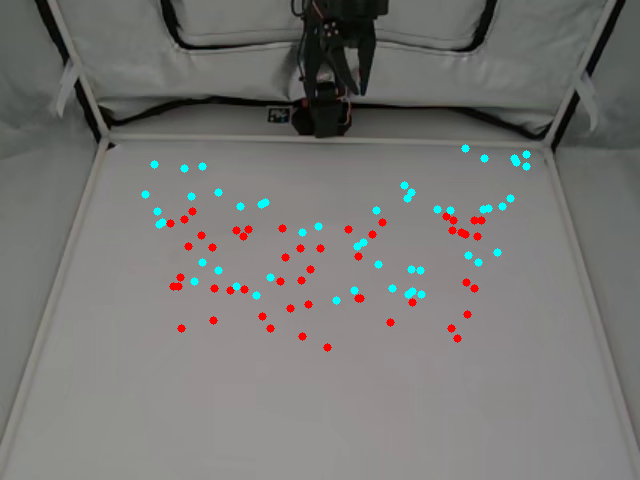}{%
    \legenddot{Red}         & Red plate \\
    \legenddot{SkyBlue}        & Pepper \\
  }%
  \hfill
  \imgset{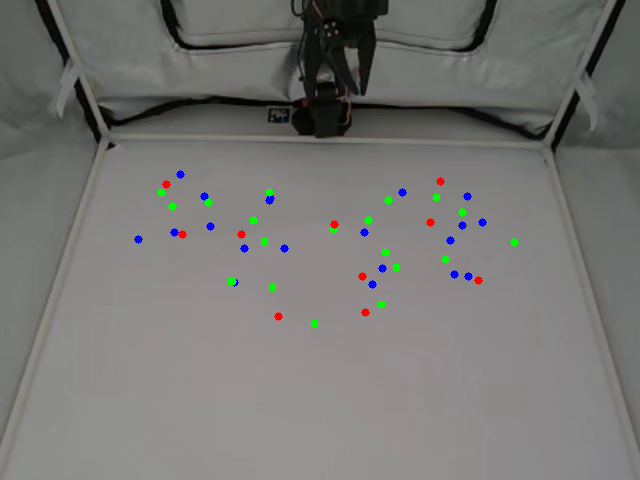}{%
    \legenddot{Red}      & Red bowl \\
    \legenddot{Blue}     & Blue bowl \\
    \legenddot{Green}      & Green bowl \\
  }%
  \caption{Tasks 7--8: object center annotations.}
  \label{fig:centers_7_8}
\end{figure}
 
\begin{figure}[H]
  \centering
  \imgset{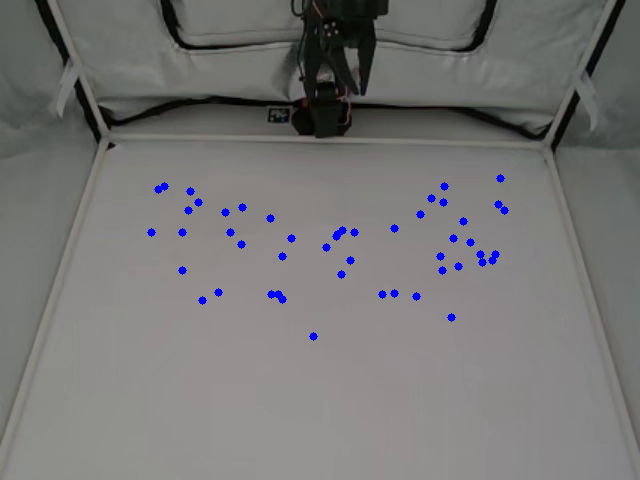}{%
    \legenddot{Blue}         & Button \\
  }%
  \hfill
  \imgset{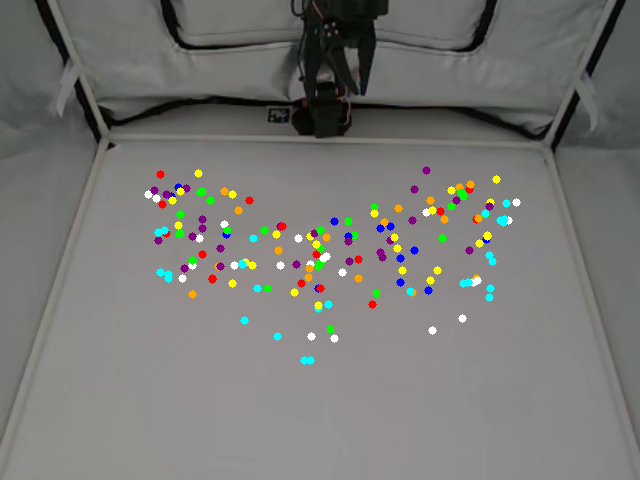}{%
    \legenddot{Red}        & Red block \\
    \legenddot{Blue}      & Blue block \\
    \legenddot{Yellow}        & Yellow block \\
    \legenddot{Green}      & Green block \\
    \legenddot{Orange}        & Orange block \\
    \legenddot{Purple}      & Purple block \\
    \legendddot{White}       & Yellow box \\
    \legenddot{SkyBlue}        & Blue box \\
  }%
  \caption{Tasks 9--10: object center annotations.}
  \label{fig:centers_9_10}
\end{figure}

\section{Test Scene Reference Images}
\label{app:test_scene_reference_images}

All 90 test scene reference images are provided below. For all tasks, the first row includes ID (in-distribution) tasks, and the second row includes OOD (out-of-distribution) tasks (except 5 \& 6).

\begin{figure}[H]
    \centering
    \includegraphics[width=\linewidth]{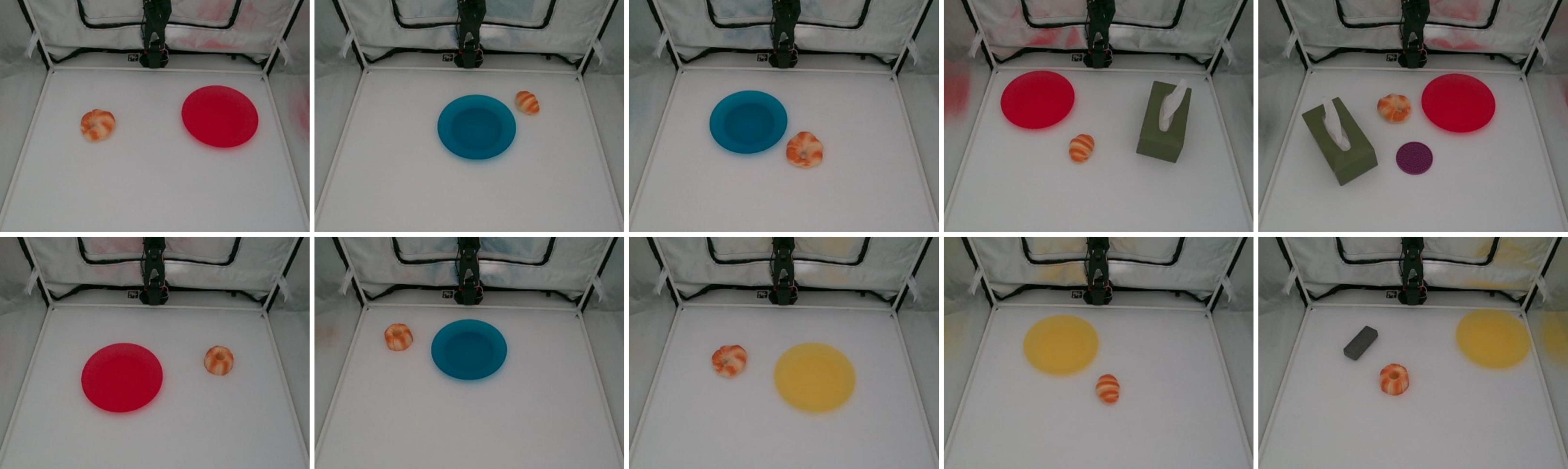}
    \caption{Reference images for Task 1: Put Bread on Plate (ID\# 1-10).}
    \label{fig:task01}
\end{figure}

\begin{figure}[H]
    \centering
    \includegraphics[width=\linewidth]{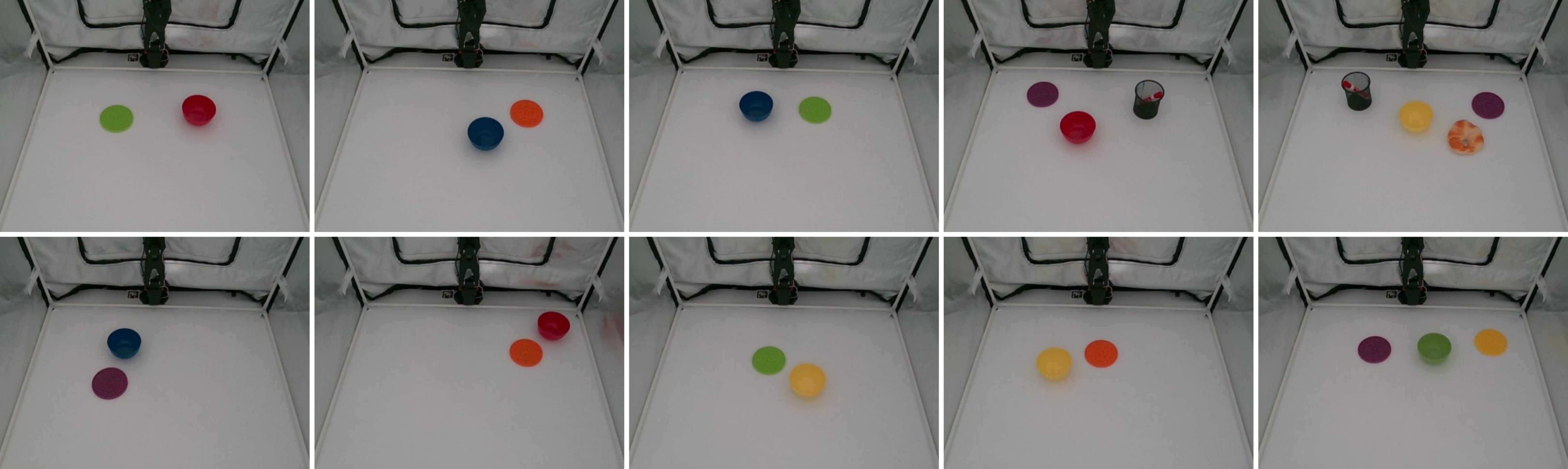}
    \caption{Reference images for Task 2: Put Bowl on Coaster (ID\# 11-20).}
    \label{fig:task02}
\end{figure}

\begin{figure}[H]
    \centering
    \includegraphics[width=\linewidth]{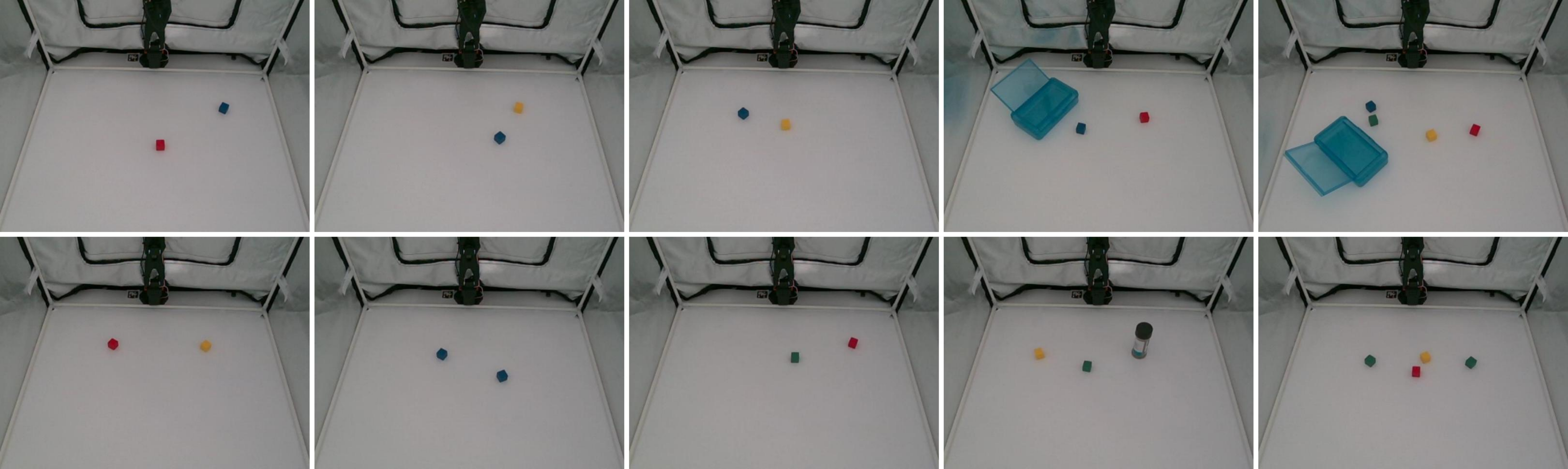}
    \caption{Reference images for Task 3: Stack Blocks (ID\# 21-30).}
    \label{fig:task03}
\end{figure}

\begin{figure}[H]
    \centering
    \includegraphics[width=\linewidth]{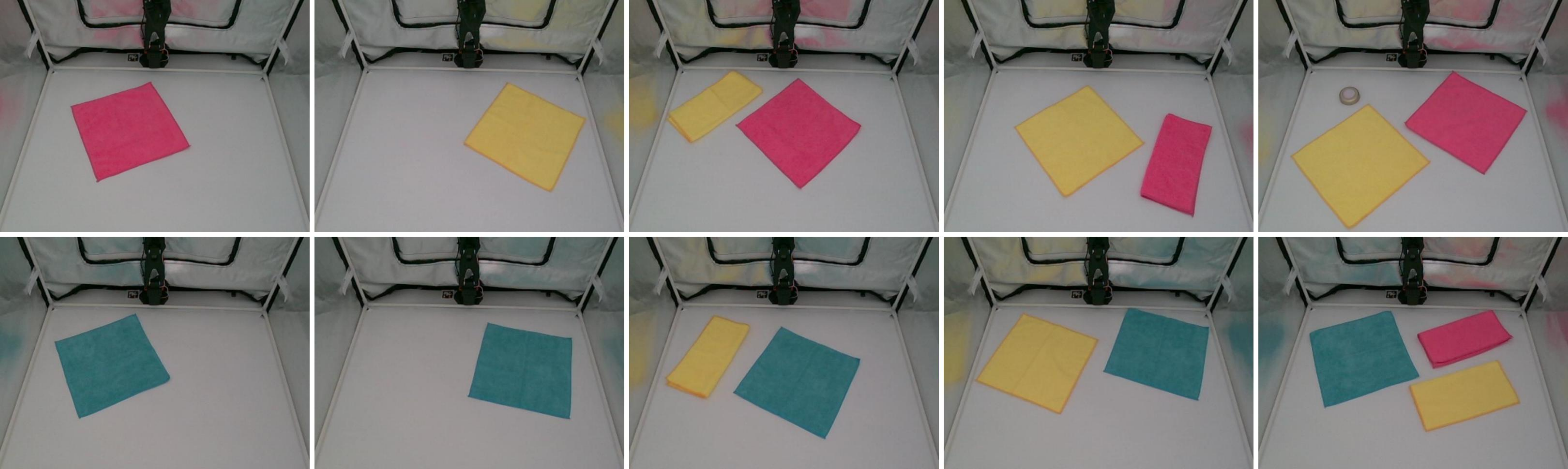}
    \caption{Reference images for Task 4: Fold Towel (ID\# 31-40).}
    \label{fig:task04}
\end{figure}

\begin{figure}[H]
    \centering
    \includegraphics[width=\linewidth]{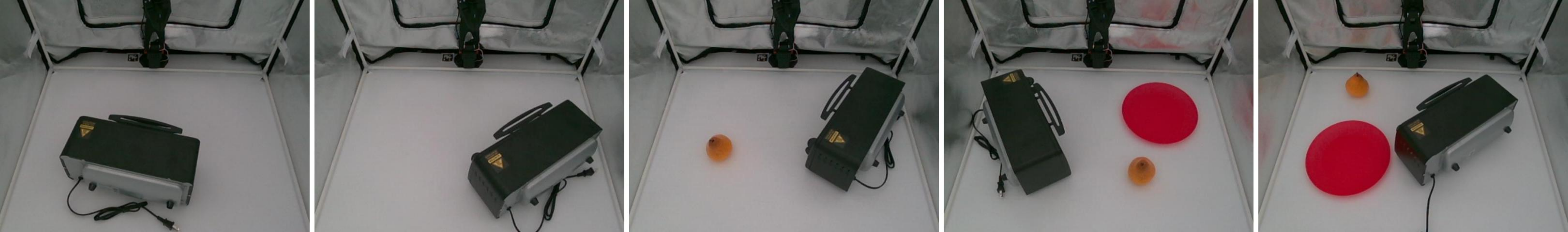}
    \caption{Reference images for Task 5: Open Oven (ID\# 41-45).}
    \label{fig:task05}
\end{figure}

\begin{figure}[H]
    \centering
    \includegraphics[width=\linewidth]{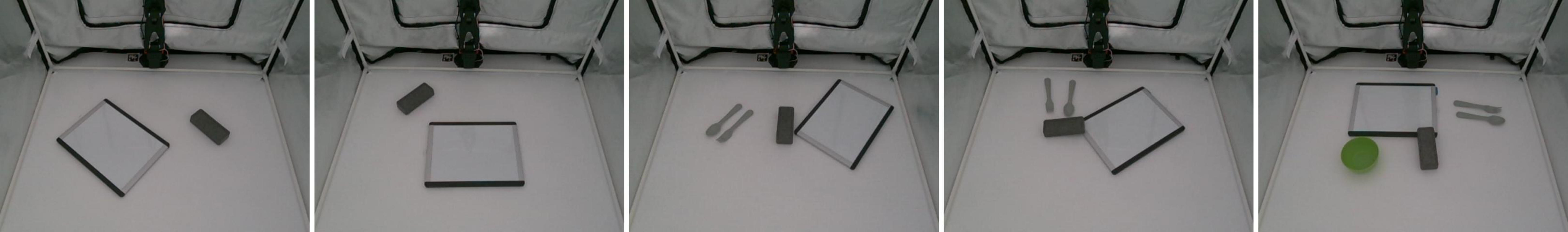}
    \caption{Reference images for Task 6: Clean Whiteboard (ID\# 46-50).}
    \label{fig:task06}
\end{figure}

\begin{figure}[H]
    \centering
    \includegraphics[width=\linewidth]{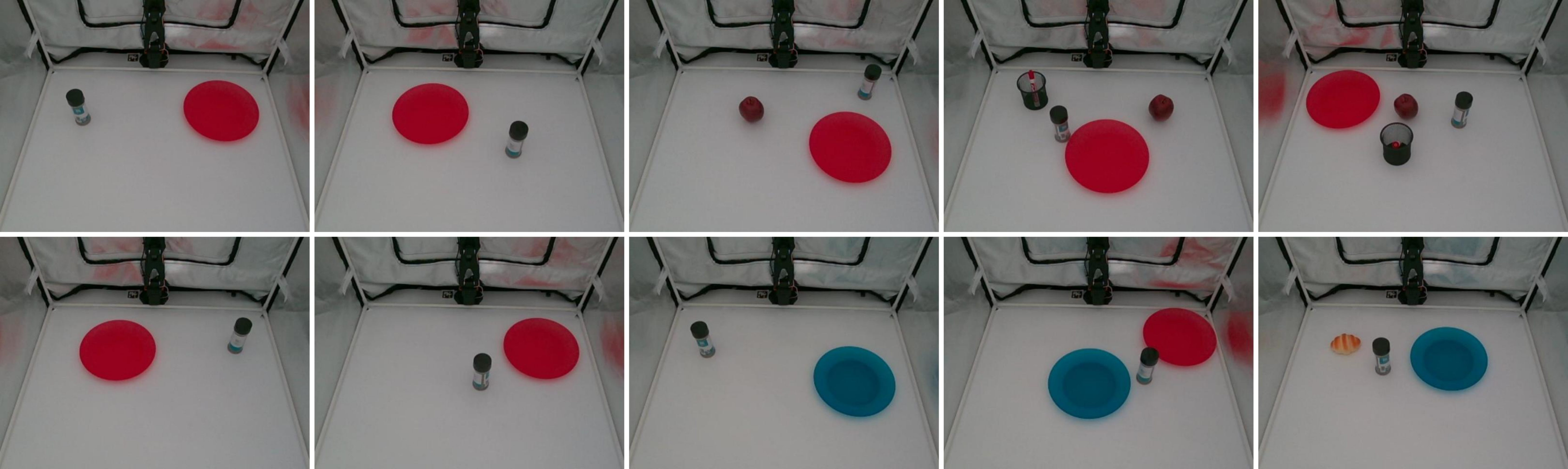}
    \caption{Reference images for Task 7: Pour Pepper (ID\# 51-60).}
    \label{fig:task07}
\end{figure}

\begin{figure}[H]
    \centering
    \includegraphics[width=\linewidth]{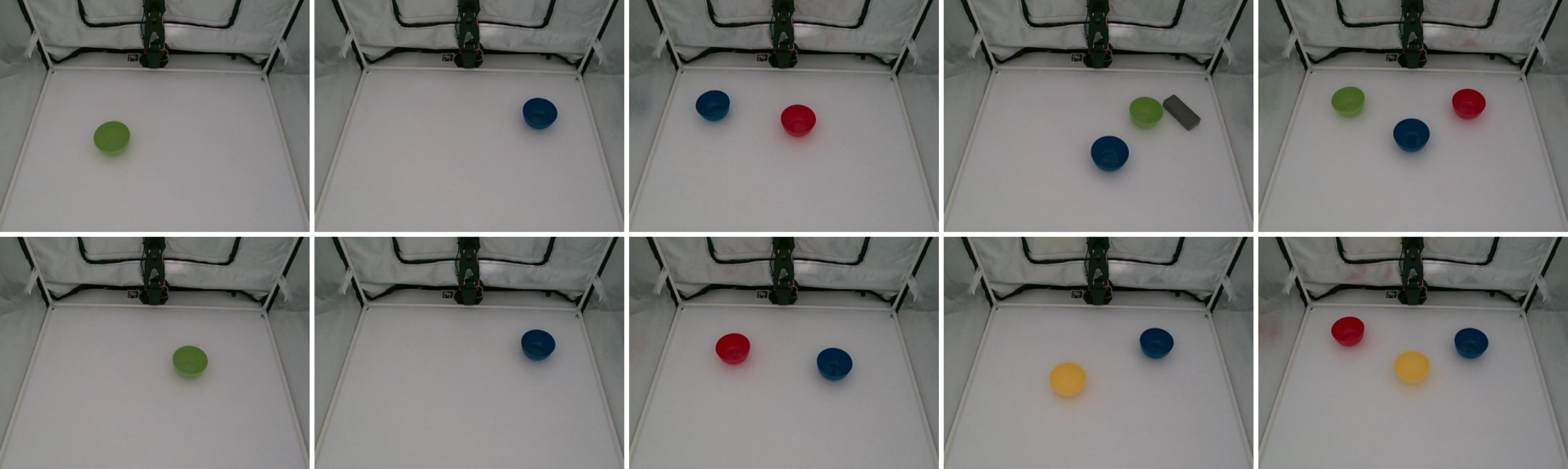}
    \caption{Reference images for Task 8: Lift Bowl (ID\# 61-70).}
    \label{fig:task08}
    \vspace{-3mm}
\end{figure}

\begin{figure}[H]
    \centering
    \includegraphics[width=\linewidth]{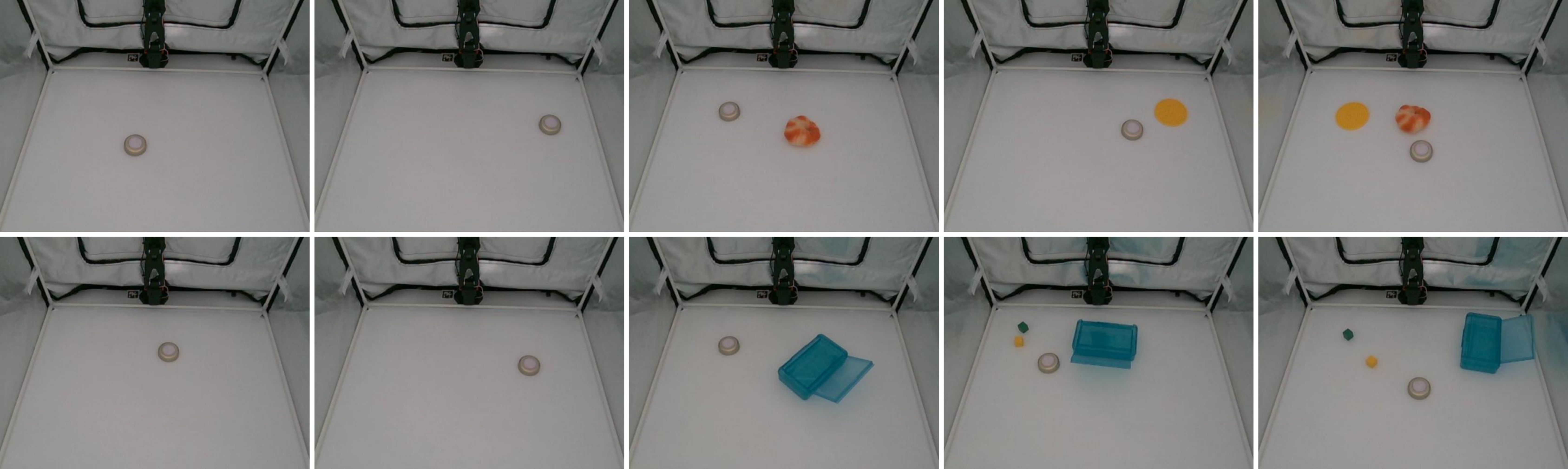}
    \caption{Reference images for Task 9: Press Button (ID\# 71-80).}
    \label{fig:task09}
    \vspace{-3mm}
\end{figure}

\begin{figure}[H]
    \centering
    \includegraphics[width=\linewidth]{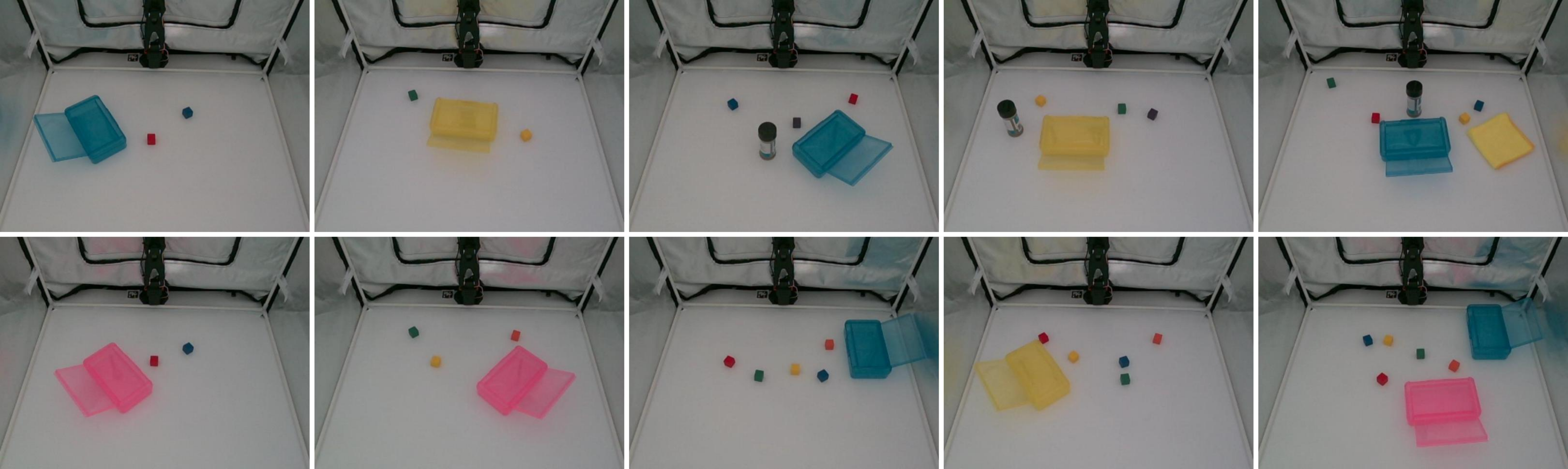}
    \caption{Reference images for Task 10: Collect Blocks (ID\# 81-90).}
    \label{fig:task10}
\end{figure}

\section{Training Details}
\label{app:training}

All imitation learning and VLA fine-tuning experiments are implemented using the official LeRobot codebase~\citep{cadene2024lerobot}. Training configurations for the nine policies are reported in Tables~\ref{tab:training_details} and~\ref{tab:training_details_pt2}, respectively. For all methods, we keep the default optimizer, learning rate, and weight decay settings from the official implementation.
\vspace{-3mm}
\begin{table}[h]
    \centering
    \scriptsize
    \setlength{\tabcolsep}{3pt}
    \renewcommand{\arraystretch}{0.8}
    \caption{Training and fine-tuning details for the evaluated policies.}
    \label{tab:training_details}
    \begin{tabular}{lccccccc}
        \toprule
        \textbf{Parameter} 
        & \textbf{ACT}
        & \textbf{DiT-D}
        & \textbf{DiT-F}
        & \textbf{SmolVLA}
        & \textbf{X-VLA}
        & $\bm{\pi}_0$
        & $\bm{\pi}_{0.5}$ \\
        \midrule
        Training type 
        & From scratch & From scratch & From scratch & Fine-tuning & Fine-tuning & Fine-tuning & Fine-tuning \\
        Dataset size 
        & 500 demos & 500 demos & 500 demos & 500 demos & 500 demos & 500 demos & 500 demos \\
        Batch size 
        & 128 & 128 & 128 & 128 & 128 & 16 & 16 \\
        Training steps 
        & 40K & 40K & 40K & 40K & 40K & 40K & 40K \\
        GPUs 
        & 2 & 1 & 1 & 4 & 4 & 1 & 1 \\
        GPU type 
        & A6000 Ada & H200 & H200 & A6000 Ada & A6000 Ada & H200 & H200 \\
        Action chunk size 
        & 32 & 32 & 32 & 32 & 32 & 32 & 32 \\
        Number of action steps 
        & 32 & 24 & 24 & 32 & 32 & 32 & 32 \\
        Vision encoder 
        & Default & Default & Default & Default & Trainable & Frozen & Default \\
        Implementation 
        & LeRobot & LeRobot & LeRobot & LeRobot & LeRobot & LeRobot & LeRobot \\
        Learning Rate
        & Default & Default & Default & Default & Default & Default & Default \\
        \bottomrule
    \end{tabular}
\end{table}

\vspace{-3mm}
\begin{table}[h]
    \centering
    \scriptsize
    \setlength{\tabcolsep}{3pt}
    \renewcommand{\arraystretch}{0.8}
    \caption{Training and fine-tuning details for additional evaluated policies.}
    \label{tab:training_details_pt2}
    \begin{tabular}{lcc}
        \toprule
        \textbf{Parameter} 
        & \textbf{MolmoAct2-SO100\_101} & \textbf{GR00T N1.7} \\
        \midrule
        Training type 
        & Fine-tuning
        & Fine-tuning
        \\
        Dataset size 
        & 500 demos
        & 500 demos
        \\
        Batch size 
        & 16 & 32 \\
        Training steps 
        & 40K
        & 40K
        \\
        GPUs 
        & 4 & 4 \\
        GPU type 
        & A6000 Ada
        & A6000 Ada \\
        Action chunk size 
        & 32
        & 32
        \\
        Number of action steps 
        & 32
        & 32 \\
        Vision encoder 
        & Default
        & Default
        \\
        Implementation 
        & LeRobot
        & LeRobot
        \\
        Learning Rate
        & Default
        & Default
        \\
        \bottomrule
    \end{tabular}
\end{table}

\section{Detailed Evaluation Results}
\label{app:detailed_evaluation_results}
Tables~\ref{tab:ID_success_rate} and~\ref{tab:OOD_success_rate} report the updated three-round evaluation of nine policies. The project leaderboard provides separate Run 1, Run 2, and Run 3 video links for each policy and split at \url{https://irvlutd.github.io/VLAReplica/\#leaderboard}. Across the 50 ID and 40 OOD scenes, this protocol comprises 2,430 evaluation trials (nine policies, three rounds, 90 scenes), excluding setup tests and the separate reproducibility study.

Tables~\ref{tab:ID_success_rate} and~\ref{tab:OOD_success_rate} summarize the benchmark results. Here we report scene-level binary outcomes for nine policies on the original environment, as well as the archived separate reproduced-setup study. The project leaderboard provides rollout videos at \url{https://irvlutd.github.io/VLAReplica/\#leaderboard}.

\begin{table}[htbp]
\centering
\footnotesize
\caption{ACT --- scene-level binary outcomes in the original and reproduced environments.}
\label{tab:act}
\begin{subtable}[t]{0.495\linewidth}
\centering
\caption{ID, original env.\ (10 tasks)}
\label{tab:act_id_orig}
\resizebox{\linewidth}{!}{%
\begin{tabular}{lcccccc}
\toprule
\textbf{Task} & \textbf{\#1} & \textbf{\#2} & \textbf{\#3} & \textbf{\#4} & \textbf{\#5} & \textbf{Avg.} \\
\midrule
Bread on Plate & \texttt{111} & \texttt{100} & \texttt{000} & \texttt{000} & \texttt{010} & 0.33 \\
Bowl on Coaster & \texttt{000} & \texttt{000} & \texttt{010} & \texttt{000} & \texttt{000} & 0.07 \\
Stack Block & \texttt{000} & \texttt{000} & \texttt{000} & \texttt{000} & \texttt{000} & 0.00 \\
Fold Towel & \texttt{111} & \texttt{111} & \texttt{011} & \texttt{010} & \texttt{000} & 0.60 \\
Open Oven & \texttt{110} & \texttt{000} & \texttt{000} & \texttt{000} & \texttt{111} & 0.33 \\
Erase Whiteboard & \texttt{000} & \texttt{101} & \texttt{010} & \texttt{000} & \texttt{000} & 0.20 \\
Pepper $N$ Times & \texttt{000} & \texttt{000} & \texttt{110} & \texttt{011} & \texttt{000} & 0.27 \\
Lift Bowl $N$ Times & \texttt{000} & \texttt{011} & \texttt{100} & \texttt{010} & \texttt{010} & 0.33 \\
Press Button $N$ Times & \texttt{000} & \texttt{000} & \texttt{000} & \texttt{000} & \texttt{000} & 0.00 \\
$N$ Blocks into Box & \texttt{000} & \texttt{000} & \texttt{000} & \texttt{000} & \texttt{000} & 0.00 \\
\midrule
\textbf{Overall} & \multicolumn{5}{c}{\scriptsize Run 1/2/3: 0.18 / 0.30 / 0.16} & \textbf{0.21} \\
\bottomrule
\end{tabular}%
}
\end{subtable}
\hfill
\begin{subtable}[t]{0.495\linewidth}
\centering
\caption{OOD, original env.\ (8 tasks)}
\label{tab:act_ood_orig}
\resizebox{\linewidth}{!}{%
\begin{tabular}{lcccccc}
\toprule
\textbf{Task} & \textbf{\#1} & \textbf{\#2} & \textbf{\#3} & \textbf{\#4} & \textbf{\#5} & \textbf{Avg.} \\
\midrule
Bread on Plate & \texttt{100} & \texttt{011} & \texttt{001} & \texttt{001} & \texttt{100} & 0.40 \\
Bowl on Coaster & \texttt{111} & \texttt{000} & \texttt{000} & \texttt{000} & \texttt{010} & 0.27 \\
Stack Block & \texttt{000} & \texttt{000} & \texttt{000} & \texttt{000} & \texttt{000} & 0.00 \\
Fold Towel & \texttt{000} & \texttt{000} & \texttt{000} & \texttt{000} & \texttt{000} & 0.00 \\
Open Oven & \texttt{{\char45}{\char45}{\char45}} & \texttt{{\char45}{\char45}{\char45}} & \texttt{{\char45}{\char45}{\char45}} & \texttt{{\char45}{\char45}{\char45}} & \texttt{{\char45}{\char45}{\char45}} & -- \\
Erase Whiteboard & \texttt{{\char45}{\char45}{\char45}} & \texttt{{\char45}{\char45}{\char45}} & \texttt{{\char45}{\char45}{\char45}} & \texttt{{\char45}{\char45}{\char45}} & \texttt{{\char45}{\char45}{\char45}} & -- \\
Pepper $N$ Times & \texttt{000} & \texttt{000} & \texttt{000} & \texttt{000} & \texttt{000} & 0.00 \\
Lift Bowl $N$ Times & \texttt{000} & \texttt{000} & \texttt{000} & \texttt{000} & \texttt{000} & 0.00 \\
Press Button $N$ Times & \texttt{000} & \texttt{000} & \texttt{000} & \texttt{000} & \texttt{000} & 0.00 \\
$N$ Blocks into Box & \texttt{000} & \texttt{000} & \texttt{000} & \texttt{000} & \texttt{000} & 0.00 \\
\midrule
\textbf{Overall} & \multicolumn{5}{c}{\scriptsize Run 1/2/3: 0.07 / 0.07 / 0.10} & \textbf{0.08} \\
\bottomrule
\end{tabular}%
}
\end{subtable}
\\[0.8em]
\begin{subtable}[t]{0.495\linewidth}
\centering
\caption{ID, reproduced env.\ (10 tasks)}
\label{tab:act_id_repro}
\resizebox{\linewidth}{!}{%
\begin{tabular}{lcccccc}
\toprule
\textbf{Task} & \textbf{\#1} & \textbf{\#2} & \textbf{\#3} & \textbf{\#4} & \textbf{\#5} & \textbf{Avg.} \\
\midrule
Bread on Plate & 1 & 0 & 0 & 0 & 1 & 0.40 \\
Bowl on Coaster & 0 & 0 & 0 & 0 & 0 & 0.00 \\
Stack Block & 0 & 0 & 0 & 0 & 0 & 0.00 \\
$N$ Blocks into Box & 0 & 0 & 0 & 0 & 0 & 0.00 \\
Fold Towel & 1 & 1 & 1 & 0 & 0 & 0.60 \\
Open Oven & 0 & 0 & 1 & 0 & 1 & 0.40 \\
Erase Whiteboard & 0 & 0 & 1 & 0 & 0 & 0.20 \\
Pepper $N$ Times & 0 & 1 & 0 & 1 & 0 & 0.40 \\
Lift Bowl $N$ Times & 0 & 1 & 0 & 0 & 0 & 0.20 \\
Press Button $N$ Times & 0 & 0 & 0 & 0 & 0 & 0.00 \\
\midrule
\textbf{Overall} & \multicolumn{5}{c}{} & \textbf{0.22} \\
\bottomrule
\end{tabular}%
}
\end{subtable}
\hfill
\begin{subtable}[t]{0.495\linewidth}
\centering
\caption{OOD, reproduced env.\ (8 tasks)}
\label{tab:act_ood_repro}
\resizebox{\linewidth}{!}{%
\begin{tabular}{lcccccc}
\toprule
\textbf{Task} & \textbf{\#1} & \textbf{\#2} & \textbf{\#3} & \textbf{\#4} & \textbf{\#5} & \textbf{Avg.} \\
\midrule
Bread on Plate & 0 & 1 & 0 & 0 & 1 & 0.40 \\
Bowl on Coaster & 1 & 0 & 0 & 0 & 1 & 0.40 \\
Stack Block & 0 & 0 & 0 & 0 & 0 & 0.00 \\
$N$ Blocks into Box & 0 & 0 & 0 & 0 & 0 & 0.00 \\
Fold Towel & 0 & 0 & 0 & 0 & 0 & 0.00 \\
Open Oven & -- & -- & -- & -- & -- & -- \\
Erase Whiteboard & -- & -- & -- & -- & -- & -- \\
Pepper $N$ Times & 0 & 0 & 0 & 0 & 0 & 0.00 \\
Lift Bowl $N$ Times & 0 & 0 & 0 & 0 & 0 & 0.00 \\
Press Button $N$ Times & 0 & 0 & 0 & 0 & 0 & 0.00 \\
\midrule
\textbf{Overall} & \multicolumn{5}{c}{} & \textbf{0.10} \\
\bottomrule
\end{tabular}%
}
\end{subtable}
\end{table}

\begin{table}[htbp]
\centering
\footnotesize
\caption{$\pi_0$ --- scene-level binary outcomes in the original and reproduced environments.}
\label{tab:pi0}
\begin{subtable}[t]{0.495\linewidth}
\centering
\caption{ID, original env.\ (10 tasks)}
\label{tab:pi0_id_orig}
\resizebox{\linewidth}{!}{%
\begin{tabular}{lcccccc}
\toprule
\textbf{Task} & \textbf{\#1} & \textbf{\#2} & \textbf{\#3} & \textbf{\#4} & \textbf{\#5} & \textbf{Avg.} \\
\midrule
Bread on Plate & \texttt{111} & \texttt{111} & \texttt{100} & \texttt{111} & \texttt{000} & 0.67 \\
Bowl on Coaster & \texttt{011} & \texttt{110} & \texttt{110} & \texttt{111} & \texttt{010} & 0.67 \\
Stack Block & \texttt{000} & \texttt{000} & \texttt{000} & \texttt{000} & \texttt{000} & 0.00 \\
Fold Towel & \texttt{110} & \texttt{110} & \texttt{111} & \texttt{100} & \texttt{011} & 0.67 \\
Open Oven & \texttt{000} & \texttt{100} & \texttt{000} & \texttt{011} & \texttt{000} & 0.20 \\
Erase Whiteboard & \texttt{101} & \texttt{110} & \texttt{000} & \texttt{001} & \texttt{000} & 0.33 \\
Pepper $N$ Times & \texttt{111} & \texttt{001} & \texttt{000} & \texttt{000} & \texttt{000} & 0.27 \\
Lift Bowl $N$ Times & \texttt{000} & \texttt{011} & \texttt{100} & \texttt{000} & \texttt{000} & 0.20 \\
Press Button $N$ Times & \texttt{000} & \texttt{111} & \texttt{000} & \texttt{000} & \texttt{000} & 0.20 \\
$N$ Blocks into Box & \texttt{000} & \texttt{000} & \texttt{000} & \texttt{000} & \texttt{000} & 0.00 \\
\midrule
\textbf{Overall} & \multicolumn{5}{c}{\scriptsize Run 1/2/3: 0.34 / 0.34 / 0.28} & \textbf{0.32} \\
\bottomrule
\end{tabular}%
}
\end{subtable}
\hfill
\begin{subtable}[t]{0.495\linewidth}
\centering
\caption{OOD, original env.\ (8 tasks)}
\label{tab:pi0_ood_orig}
\resizebox{\linewidth}{!}{%
\begin{tabular}{lcccccc}
\toprule
\textbf{Task} & \textbf{\#1} & \textbf{\#2} & \textbf{\#3} & \textbf{\#4} & \textbf{\#5} & \textbf{Avg.} \\
\midrule
Bread on Plate & \texttt{000} & \texttt{111} & \texttt{110} & \texttt{111} & \texttt{111} & 0.73 \\
Bowl on Coaster & \texttt{111} & \texttt{101} & \texttt{001} & \texttt{010} & \texttt{111} & 0.67 \\
Stack Block & \texttt{000} & \texttt{101} & \texttt{000} & \texttt{001} & \texttt{000} & 0.20 \\
Fold Towel & \texttt{111} & \texttt{111} & \texttt{000} & \texttt{100} & \texttt{0{\char45}{\char45}} & 0.54 \\
Open Oven & \texttt{{\char45}{\char45}{\char45}} & \texttt{{\char45}{\char45}{\char45}} & \texttt{{\char45}{\char45}{\char45}} & \texttt{{\char45}{\char45}{\char45}} & \texttt{{\char45}{\char45}{\char45}} & -- \\
Erase Whiteboard & \texttt{{\char45}{\char45}{\char45}} & \texttt{{\char45}{\char45}{\char45}} & \texttt{{\char45}{\char45}{\char45}} & \texttt{{\char45}{\char45}{\char45}} & \texttt{{\char45}{\char45}{\char45}} & -- \\
Pepper $N$ Times & \texttt{010} & \texttt{000} & \texttt{101} & \texttt{000} & \texttt{00{\char45}} & 0.21 \\
Lift Bowl $N$ Times & \texttt{000} & \texttt{000} & \texttt{000} & \texttt{000} & \texttt{000} & 0.00 \\
Press Button $N$ Times & \texttt{010} & \texttt{000} & \texttt{000} & \texttt{000} & \texttt{000} & 0.07 \\
$N$ Blocks into Box & \texttt{000} & \texttt{000} & \texttt{000} & \texttt{000} & \texttt{000} & 0.00 \\
\midrule
\textbf{Overall} & \multicolumn{5}{c}{\scriptsize Run 1/2/3: 0.30 / 0.28 / 0.32} & \textbf{0.30} \\
\bottomrule
\end{tabular}%
}
\end{subtable}
\\[0.8em]
\begin{subtable}[t]{0.495\linewidth}
\centering
\caption{ID, reproduced env.\ (10 tasks)}
\label{tab:pi0_id_repro}
\resizebox{\linewidth}{!}{%
\begin{tabular}{lcccccc}
\toprule
\textbf{Task} & \textbf{\#1} & \textbf{\#2} & \textbf{\#3} & \textbf{\#4} & \textbf{\#5} & \textbf{Avg.} \\
\midrule
Bread on Plate & 1 & 1 & 0 & 1 & 1 & 0.80 \\
Bowl on Coaster & 1 & 0 & 1 & 1 & 0 & 0.60 \\
Stack Block & 1 & 0 & 0 & 0 & 0 & 0.20 \\
$N$ Blocks into Box & 0 & 0 & 0 & 0 & 0 & 0.00 \\
Fold Towel & 1 & 1 & 1 & 0 & 0 & 0.60 \\
Open Oven & 0 & 0 & 0 & 0 & 1 & 0.20 \\
Erase Whiteboard & 0 & 1 & 0 & 0 & 0 & 0.20 \\
Pepper $N$ Times & 1 & 0 & 0 & 0 & 0 & 0.20 \\
Lift Bowl $N$ Times & 0 & 1 & 0 & 0 & 0 & 0.20 \\
Press Button $N$ Times & 0 & 1 & 0 & 0 & 0 & 0.20 \\
\midrule
\textbf{Overall} & \multicolumn{5}{c}{} & \textbf{0.32} \\
\bottomrule
\end{tabular}%
}
\end{subtable}
\hfill
\begin{subtable}[t]{0.495\linewidth}
\centering
\caption{OOD, reproduced env.\ (8 tasks)}
\label{tab:pi0_ood_repro}
\resizebox{\linewidth}{!}{%
\begin{tabular}{lcccccc}
\toprule
\textbf{Task} & \textbf{\#1} & \textbf{\#2} & \textbf{\#3} & \textbf{\#4} & \textbf{\#5} & \textbf{Avg.} \\
\midrule
Bread on Plate & 1 & 1 & 0 & 1 & 1 & 0.80 \\
Bowl on Coaster & 0 & 1 & 0 & 0 & 1 & 0.40 \\
Stack Block & 0 & 1 & 0 & 0 & 0 & 0.20 \\
$N$ Blocks into Box & 0 & 0 & 0 & 0 & 0 & 0.00 \\
Fold Towel & 0 & 0 & 1 & 1 & 1 & 0.60 \\
Open Oven & -- & -- & -- & -- & -- & -- \\
Erase Whiteboard & -- & -- & -- & -- & -- & -- \\
Pepper $N$ Times & 0 & 0 & 1 & 0 & 0 & 0.20 \\
Lift Bowl $N$ Times & 0 & 0 & 0 & 1 & 0 & 0.20 \\
Press Button $N$ Times & 0 & 1 & 0 & 0 & 0 & 0.20 \\
\midrule
\textbf{Overall} & \multicolumn{5}{c}{} & \textbf{0.325} \\
\bottomrule
\end{tabular}%
}
\end{subtable}
\end{table}

\begin{table}[htbp]
\centering
\footnotesize
\caption{$\pi_{0.5}$ --- scene-level binary outcomes in the original and reproduced environments.}
\label{tab:pi05}
\begin{subtable}[t]{0.495\linewidth}
\centering
\caption{ID, original env.\ (10 tasks)}
\label{tab:pi05_id_orig}
\resizebox{\linewidth}{!}{%
\begin{tabular}{lcccccc}
\toprule
\textbf{Task} & \textbf{\#1} & \textbf{\#2} & \textbf{\#3} & \textbf{\#4} & \textbf{\#5} & \textbf{Avg.} \\
\midrule
Bread on Plate & \texttt{110} & \texttt{111} & \texttt{010} & \texttt{111} & \texttt{111} & 0.80 \\
Bowl on Coaster & \texttt{111} & \texttt{111} & \texttt{111} & \texttt{110} & \texttt{011} & 0.87 \\
Stack Block & \texttt{101} & \texttt{010} & \texttt{000} & \texttt{110} & \texttt{000} & 0.33 \\
Fold Towel & \texttt{111} & \texttt{111} & \texttt{111} & \texttt{111} & \texttt{111} & 1.00 \\
Open Oven & \texttt{000} & \texttt{100} & \texttt{011} & \texttt{110} & \texttt{111} & 0.53 \\
Erase Whiteboard & \texttt{101} & \texttt{111} & \texttt{000} & \texttt{000} & \texttt{000} & 0.33 \\
Pepper $N$ Times & \texttt{000} & \texttt{001} & \texttt{000} & \texttt{100} & \texttt{111} & 0.33 \\
Lift Bowl $N$ Times & \texttt{000} & \texttt{111} & \texttt{000} & \texttt{010} & \texttt{101} & 0.40 \\
Press Button $N$ Times & \texttt{100} & \texttt{010} & \texttt{010} & \texttt{010} & \texttt{000} & 0.27 \\
$N$ Blocks into Box & \texttt{111} & \texttt{000} & \texttt{100} & \texttt{011} & \texttt{001} & 0.47 \\
\midrule
\textbf{Overall} & \multicolumn{5}{c}{\scriptsize Run 1/2/3: 0.54 / 0.58 / 0.48} & \textbf{0.53} \\
\bottomrule
\end{tabular}%
}
\end{subtable}
\hfill
\begin{subtable}[t]{0.495\linewidth}
\centering
\caption{OOD, original env.\ (8 tasks)}
\label{tab:pi05_ood_orig}
\resizebox{\linewidth}{!}{%
\begin{tabular}{lcccccc}
\toprule
\textbf{Task} & \textbf{\#1} & \textbf{\#2} & \textbf{\#3} & \textbf{\#4} & \textbf{\#5} & \textbf{Avg.} \\
\midrule
Bread on Plate & \texttt{111} & \texttt{111} & \texttt{110} & \texttt{111} & \texttt{111} & 0.93 \\
Bowl on Coaster & \texttt{111} & \texttt{010} & \texttt{111} & \texttt{000} & \texttt{011} & 0.60 \\
Stack Block & \texttt{110} & \texttt{000} & \texttt{010} & \texttt{000} & \texttt{000} & 0.20 \\
Fold Towel & \texttt{111} & \texttt{111} & \texttt{111} & \texttt{100} & \texttt{011} & 0.80 \\
Open Oven & \texttt{{\char45}{\char45}{\char45}} & \texttt{{\char45}{\char45}{\char45}} & \texttt{{\char45}{\char45}{\char45}} & \texttt{{\char45}{\char45}{\char45}} & \texttt{{\char45}{\char45}{\char45}} & -- \\
Erase Whiteboard & \texttt{{\char45}{\char45}{\char45}} & \texttt{{\char45}{\char45}{\char45}} & \texttt{{\char45}{\char45}{\char45}} & \texttt{{\char45}{\char45}{\char45}} & \texttt{{\char45}{\char45}{\char45}} & -- \\
Pepper $N$ Times & \texttt{000} & \texttt{000} & \texttt{110} & \texttt{110} & \texttt{000} & 0.27 \\
Lift Bowl $N$ Times & \texttt{000} & \texttt{000} & \texttt{000} & \texttt{000} & \texttt{000} & 0.00 \\
Press Button $N$ Times & \texttt{010} & \texttt{000} & \texttt{000} & \texttt{000} & \texttt{000} & 0.07 \\
$N$ Blocks into Box & \texttt{111} & \texttt{010} & \texttt{000} & \texttt{000} & \texttt{000} & 0.27 \\
\midrule
\textbf{Overall} & \multicolumn{5}{c}{\scriptsize Run 1/2/3: 0.38 / 0.50 / 0.30} & \textbf{0.39} \\
\bottomrule
\end{tabular}%
}
\end{subtable}
\\[0.8em]
\begin{subtable}[t]{0.495\linewidth}
\centering
\caption{ID, reproduced env.\ (10 tasks)}
\label{tab:pi05_id_repro}
\resizebox{\linewidth}{!}{%
\begin{tabular}{lcccccc}
\toprule
\textbf{Task} & \textbf{\#1} & \textbf{\#2} & \textbf{\#3} & \textbf{\#4} & \textbf{\#5} & \textbf{Avg.} \\
\midrule
Bread on Plate & 1 & 0 & 1 & 1 & 1 & 0.80 \\
Bowl on Coaster & 1 & 1 & 1 & 1 & 1 & 1.00 \\
Stack Block & 1 & 0 & 1 & 1 & 0 & 0.60 \\
$N$ Blocks into Box & 1 & 0 & 1 & 0 & 0 & 0.40 \\
Fold Towel & 1 & 1 & 1 & 0 & 1 & 0.80 \\
Open Oven & 1 & 0 & 1 & 0 & 0 & 0.40 \\
Erase Whiteboard & 1 & 1 & 0 & 0 & 0 & 0.40 \\
Pepper $N$ Times & 0 & 1 & 0 & 0 & 1 & 0.40 \\
Lift Bowl $N$ Times & 0 & 1 & 1 & 0 & 0 & 0.40 \\
Press Button $N$ Times & 1 & 0 & 0 & 0 & 0 & 0.20 \\
\midrule
\textbf{Overall} & \multicolumn{5}{c}{} & \textbf{0.54} \\
\bottomrule
\end{tabular}%
}
\end{subtable}
\hfill
\begin{subtable}[t]{0.495\linewidth}
\centering
\caption{OOD, reproduced env.\ (8 tasks)}
\label{tab:pi05_ood_repro}
\resizebox{\linewidth}{!}{%
\begin{tabular}{lcccccc}
\toprule
\textbf{Task} & \textbf{\#1} & \textbf{\#2} & \textbf{\#3} & \textbf{\#4} & \textbf{\#5} & \textbf{Avg.} \\
\midrule
Bread on Plate & 1 & 1 & 0 & 1 & 1 & 0.80 \\
Bowl on Coaster & 0 & 0 & 1 & 1 & 0 & 0.40 \\
Stack Block & 0 & 0 & 1 & 0 & 0 & 0.20 \\
$N$ Blocks into Box & 1 & 1 & 0 & 0 & 0 & 0.40 \\
Fold Towel & 1 & 1 & 1 & 1 & 0 & 0.80 \\
Open Oven & -- & -- & -- & -- & -- & -- \\
Erase Whiteboard & -- & -- & -- & -- & -- & -- \\
Pepper $N$ Times & 0 & 0 & 0 & 0 & 0 & 0.00 \\
Lift Bowl $N$ Times & 0 & 1 & 0 & 0 & 0 & 0.20 \\
Press Button $N$ Times & 0 & 0 & 1 & 0 & 0 & 0.20 \\
\midrule
\textbf{Overall} & \multicolumn{5}{c}{} & \textbf{0.375} \\
\bottomrule
\end{tabular}%
}
\end{subtable}
\end{table}

\begin{table}[htbp]
\centering
\footnotesize
\caption{SmolVLA --- scene-level binary outcomes in the original and reproduced environments.}
\label{tab:smolvla}
\begin{subtable}[t]{0.495\linewidth}
\centering
\caption{ID, original env.\ (10 tasks)}
\label{tab:smolvla_id_orig}
\resizebox{\linewidth}{!}{%
\begin{tabular}{lcccccc}
\toprule
\textbf{Task} & \textbf{\#1} & \textbf{\#2} & \textbf{\#3} & \textbf{\#4} & \textbf{\#5} & \textbf{Avg.} \\
\midrule
Bread on Plate & \texttt{111} & \texttt{111} & \texttt{101} & \texttt{011} & \texttt{011} & 0.80 \\
Bowl on Coaster & \texttt{001} & \texttt{001} & \texttt{100} & \texttt{000} & \texttt{000} & 0.20 \\
Stack Block & \texttt{101} & \texttt{010} & \texttt{001} & \texttt{000} & \texttt{000} & 0.27 \\
Fold Towel & \texttt{111} & \texttt{111} & \texttt{111} & \texttt{011} & \texttt{000} & 0.73 \\
Open Oven & \texttt{101} & \texttt{000} & \texttt{110} & \texttt{000} & \texttt{010} & 0.33 \\
Erase Whiteboard & \texttt{000} & \texttt{111} & \texttt{000} & \texttt{000} & \texttt{000} & 0.20 \\
Pepper $N$ Times & \texttt{000} & \texttt{000} & \texttt{000} & \texttt{000} & \texttt{000} & 0.00 \\
Lift Bowl $N$ Times & \texttt{000} & \texttt{011} & \texttt{000} & \texttt{100} & \texttt{000} & 0.20 \\
Press Button $N$ Times & \texttt{000} & \texttt{111} & \texttt{000} & \texttt{000} & \texttt{000} & 0.20 \\
$N$ Blocks into Box & \texttt{000} & \texttt{000} & \texttt{000} & \texttt{000} & \texttt{000} & 0.00 \\
\midrule
\textbf{Overall} & \multicolumn{5}{c}{\scriptsize Run 1/2/3: 0.26 / 0.28 / 0.34} & \textbf{0.29} \\
\bottomrule
\end{tabular}%
}
\end{subtable}
\hfill
\begin{subtable}[t]{0.495\linewidth}
\centering
\caption{OOD, original env.\ (8 tasks)}
\label{tab:smolvla_ood_orig}
\resizebox{\linewidth}{!}{%
\begin{tabular}{lcccccc}
\toprule
\textbf{Task} & \textbf{\#1} & \textbf{\#2} & \textbf{\#3} & \textbf{\#4} & \textbf{\#5} & \textbf{Avg.} \\
\midrule
Bread on Plate & \texttt{101} & \texttt{111} & \texttt{111} & \texttt{000} & \texttt{111} & 0.73 \\
Bowl on Coaster & \texttt{111} & \texttt{000} & \texttt{101} & \texttt{000} & \texttt{000} & 0.33 \\
Stack Block & \texttt{000} & \texttt{100} & \texttt{001} & \texttt{001} & \texttt{000} & 0.20 \\
Fold Towel & \texttt{111} & \texttt{100} & \texttt{011} & \texttt{100} & \texttt{000} & 0.47 \\
Open Oven & \texttt{{\char45}{\char45}{\char45}} & \texttt{{\char45}{\char45}{\char45}} & \texttt{{\char45}{\char45}{\char45}} & \texttt{{\char45}{\char45}{\char45}} & \texttt{{\char45}{\char45}{\char45}} & -- \\
Erase Whiteboard & \texttt{{\char45}{\char45}{\char45}} & \texttt{{\char45}{\char45}{\char45}} & \texttt{{\char45}{\char45}{\char45}} & \texttt{{\char45}{\char45}{\char45}} & \texttt{{\char45}{\char45}{\char45}} & -- \\
Pepper $N$ Times & \texttt{000} & \texttt{000} & \texttt{000} & \texttt{000} & \texttt{000} & 0.00 \\
Lift Bowl $N$ Times & \texttt{110} & \texttt{001} & \texttt{000} & \texttt{000} & \texttt{000} & 0.20 \\
Press Button $N$ Times & \texttt{000} & \texttt{000} & \texttt{010} & \texttt{000} & \texttt{000} & 0.07 \\
$N$ Blocks into Box & \texttt{110} & \texttt{000} & \texttt{000} & \texttt{000} & \texttt{000} & 0.13 \\
\midrule
\textbf{Overall} & \multicolumn{5}{c}{\scriptsize Run 1/2/3: 0.30 / 0.23 / 0.28} & \textbf{0.27} \\
\bottomrule
\end{tabular}%
}
\end{subtable}
\\[0.8em]
\begin{subtable}[t]{0.495\linewidth}
\centering
\caption{ID, reproduced env.\ (10 tasks)}
\label{tab:smolvla_id_repro}
\resizebox{\linewidth}{!}{%
\begin{tabular}{lcccccc}
\toprule
\textbf{Task} & \textbf{\#1} & \textbf{\#2} & \textbf{\#3} & \textbf{\#4} & \textbf{\#5} & \textbf{Avg.} \\
\midrule
Bread on Plate & 1 & 1 & 1 & 0 & 0 & 0.60 \\
Bowl on Coaster & 0 & 1 & 1 & 0 & 0 & 0.40 \\
Stack Block & 0 & 0 & 0 & 0 & 0 & 0.00 \\
$N$ Blocks into Box & 0 & 0 & 0 & 0 & 0 & 0.00 \\
Fold Towel & 1 & 1 & 1 & 0 & 1 & 0.80 \\
Open Oven & 0 & 0 & 1 & 0 & 1 & 0.40 \\
Erase Whiteboard & 0 & 1 & 0 & 0 & 0 & 0.20 \\
Pepper $N$ Times & 0 & 0 & 0 & 0 & 0 & 0.00 \\
Lift Bowl $N$ Times & 0 & 1 & 0 & 0 & 0 & 0.20 \\
Press Button $N$ Times & 0 & 1 & 0 & 0 & 0 & 0.20 \\
\midrule
\textbf{Overall} & \multicolumn{5}{c}{} & \textbf{0.28} \\
\bottomrule
\end{tabular}%
}
\end{subtable}
\hfill
\begin{subtable}[t]{0.495\linewidth}
\centering
\caption{OOD, reproduced env.\ (8 tasks)}
\label{tab:smolvla_ood_repro}
\resizebox{\linewidth}{!}{%
\begin{tabular}{lcccccc}
\toprule
\textbf{Task} & \textbf{\#1} & \textbf{\#2} & \textbf{\#3} & \textbf{\#4} & \textbf{\#5} & \textbf{Avg.} \\
\midrule
Bread on Plate & 0 & 1 & 1 & 0 & 1 & 0.60 \\
Bowl on Coaster & 1 & 0 & 0 & 1 & 0 & 0.40 \\
Stack Block & 0 & 0 & 0 & 0 & 0 & 0.00 \\
$N$ Blocks into Box & 1 & 0 & 0 & 0 & 0 & 0.20 \\
Fold Towel & 1 & 0 & 1 & 1 & 0 & 0.60 \\
Open Oven & -- & -- & -- & -- & -- & -- \\
Erase Whiteboard & -- & -- & -- & -- & -- & -- \\
Pepper $N$ Times & 0 & 0 & 0 & 0 & 0 & 0.00 \\
Lift Bowl $N$ Times & 0 & 1 & 0 & 0 & 0 & 0.20 \\
Press Button $N$ Times & 0 & 0 & 0 & 0 & 0 & 0.00 \\
\midrule
\textbf{Overall} & \multicolumn{5}{c}{} & \textbf{0.25} \\
\bottomrule
\end{tabular}%
}
\end{subtable}
\end{table}

\begin{table}[htbp]
\centering
\footnotesize
\caption{DiT-Multitask --- scene-level binary outcomes in the original environment.}
\label{tab:dit_multitask}
\begin{subtable}[t]{0.495\linewidth}
\centering
\caption{ID, original env.\ (10 tasks)}
\label{tab:dit_multitask_id}
\resizebox{\linewidth}{!}{%
\begin{tabular}{lcccccc}
\toprule
\textbf{Task} & \textbf{\#1} & \textbf{\#2} & \textbf{\#3} & \textbf{\#4} & \textbf{\#5} & \textbf{Avg.} \\
\midrule
Bread on Plate & \texttt{100} & \texttt{010} & \texttt{000} & \texttt{100} & \texttt{000} & 0.20 \\
Bowl on Coaster & \texttt{011} & \texttt{000} & \texttt{000} & \texttt{000} & \texttt{000} & 0.13 \\
Stack Block & \texttt{000} & \texttt{000} & \texttt{000} & \texttt{000} & \texttt{000} & 0.00 \\
Fold Towel & \texttt{010} & \texttt{101} & \texttt{000} & \texttt{000} & \texttt{000} & 0.20 \\
Open Oven & \texttt{100} & \texttt{000} & \texttt{000} & \texttt{100} & \texttt{100} & 0.20 \\
Erase Whiteboard & \texttt{100} & \texttt{011} & \texttt{000} & \texttt{000} & \texttt{000} & 0.20 \\
Pepper $N$ Times & \texttt{000} & \texttt{010} & \texttt{000} & \texttt{000} & \texttt{000} & 0.07 \\
Lift Bowl $N$ Times & \texttt{000} & \texttt{000} & \texttt{000} & \texttt{000} & \texttt{000} & 0.00 \\
Press Button $N$ Times & \texttt{000} & \texttt{000} & \texttt{000} & \texttt{000} & \texttt{000} & 0.00 \\
$N$ Blocks into Box & \texttt{000} & \texttt{100} & \texttt{000} & \texttt{000} & \texttt{000} & 0.07 \\
\midrule
\textbf{Overall} & \multicolumn{5}{c}{\scriptsize Run 1/2/3: 0.16 / 0.10 / 0.06} & \textbf{0.11} \\
\bottomrule
\end{tabular}%
}
\end{subtable}
\hfill
\begin{subtable}[t]{0.495\linewidth}
\centering
\caption{OOD, original env.\ (8 tasks)}
\label{tab:dit_multitask_ood}
\resizebox{\linewidth}{!}{%
\begin{tabular}{lcccccc}
\toprule
\textbf{Task} & \textbf{\#1} & \textbf{\#2} & \textbf{\#3} & \textbf{\#4} & \textbf{\#5} & \textbf{Avg.} \\
\midrule
Bread on Plate & \texttt{000} & \texttt{001} & \texttt{000} & \texttt{000} & \texttt{000} & 0.07 \\
Bowl on Coaster & \texttt{111} & \texttt{001} & \texttt{001} & \texttt{000} & \texttt{000} & 0.33 \\
Stack Block & \texttt{000} & \texttt{000} & \texttt{000} & \texttt{000} & \texttt{000} & 0.00 \\
Fold Towel & \texttt{000} & \texttt{000} & \texttt{100} & \texttt{000} & \texttt{000} & 0.07 \\
Open Oven & \texttt{{\char45}{\char45}{\char45}} & \texttt{{\char45}{\char45}{\char45}} & \texttt{{\char45}{\char45}{\char45}} & \texttt{{\char45}{\char45}{\char45}} & \texttt{{\char45}{\char45}{\char45}} & -- \\
Erase Whiteboard & \texttt{{\char45}{\char45}{\char45}} & \texttt{{\char45}{\char45}{\char45}} & \texttt{{\char45}{\char45}{\char45}} & \texttt{{\char45}{\char45}{\char45}} & \texttt{{\char45}{\char45}{\char45}} & -- \\
Pepper $N$ Times & \texttt{000} & \texttt{001} & \texttt{000} & \texttt{000} & \texttt{000} & 0.07 \\
Lift Bowl $N$ Times & \texttt{000} & \texttt{000} & \texttt{000} & \texttt{000} & \texttt{000} & 0.00 \\
Press Button $N$ Times & \texttt{010} & \texttt{000} & \texttt{000} & \texttt{000} & \texttt{000} & 0.07 \\
$N$ Blocks into Box & \texttt{000} & \texttt{000} & \texttt{000} & \texttt{000} & \texttt{000} & 0.00 \\
\midrule
\textbf{Overall} & \multicolumn{5}{c}{\scriptsize Run 1/2/3: 0.05 / 0.05 / 0.12} & \textbf{0.07} \\
\bottomrule
\end{tabular}%
}
\end{subtable}
\end{table}

\begin{table}[htbp]
\centering
\footnotesize
\caption{DiT-FlowMatching --- scene-level binary outcomes in the original environment.}
\label{tab:dit_flow}
\begin{subtable}[t]{0.495\linewidth}
\centering
\caption{ID, original env.\ (10 tasks)}
\label{tab:dit_flow_id}
\resizebox{\linewidth}{!}{%
\begin{tabular}{lcccccc}
\toprule
\textbf{Task} & \textbf{\#1} & \textbf{\#2} & \textbf{\#3} & \textbf{\#4} & \textbf{\#5} & \textbf{Avg.} \\
\midrule
Bread on Plate & \texttt{100} & \texttt{000} & \texttt{100} & \texttt{000} & \texttt{000} & 0.13 \\
Bowl on Coaster & \texttt{000} & \texttt{000} & \texttt{000} & \texttt{000} & \texttt{000} & 0.00 \\
Stack Block & \texttt{000} & \texttt{000} & \texttt{000} & \texttt{000} & \texttt{000} & 0.00 \\
Fold Towel & \texttt{000} & \texttt{001} & \texttt{110} & \texttt{000} & \texttt{010} & 0.27 \\
Open Oven & \texttt{100} & \texttt{000} & \texttt{100} & \texttt{000} & \texttt{000} & 0.13 \\
Erase Whiteboard & \texttt{001} & \texttt{100} & \texttt{000} & \texttt{000} & \texttt{000} & 0.13 \\
Pepper $N$ Times & \texttt{000} & \texttt{000} & \texttt{000} & \texttt{000} & \texttt{000} & 0.00 \\
Lift Bowl $N$ Times & \texttt{000} & \texttt{000} & \texttt{000} & \texttt{000} & \texttt{000} & 0.00 \\
Press Button $N$ Times & \texttt{000} & \texttt{000} & \texttt{000} & \texttt{000} & \texttt{000} & 0.00 \\
$N$ Blocks into Box & \texttt{000} & \texttt{000} & \texttt{000} & \texttt{000} & \texttt{000} & 0.00 \\
\midrule
\textbf{Overall} & \multicolumn{5}{c}{\scriptsize Run 1/2/3: 0.12 / 0.04 / 0.04} & \textbf{0.07} \\
\bottomrule
\end{tabular}%
}
\end{subtable}
\hfill
\begin{subtable}[t]{0.495\linewidth}
\centering
\caption{OOD, original env.\ (8 tasks)}
\label{tab:dit_flow_ood}
\resizebox{\linewidth}{!}{%
\begin{tabular}{lcccccc}
\toprule
\textbf{Task} & \textbf{\#1} & \textbf{\#2} & \textbf{\#3} & \textbf{\#4} & \textbf{\#5} & \textbf{Avg.} \\
\midrule
Bread on Plate & \texttt{000} & \texttt{000} & \texttt{001} & \texttt{100} & \texttt{000} & 0.13 \\
Bowl on Coaster & \texttt{001} & \texttt{000} & \texttt{000} & \texttt{000} & \texttt{000} & 0.07 \\
Stack Block & \texttt{000} & \texttt{000} & \texttt{000} & \texttt{000} & \texttt{000} & 0.00 \\
Fold Towel & \texttt{000} & \texttt{000} & \texttt{010} & \texttt{000} & \texttt{000} & 0.07 \\
Open Oven & \texttt{{\char45}{\char45}{\char45}} & \texttt{{\char45}{\char45}{\char45}} & \texttt{{\char45}{\char45}{\char45}} & \texttt{{\char45}{\char45}{\char45}} & \texttt{{\char45}{\char45}{\char45}} & -- \\
Erase Whiteboard & \texttt{{\char45}{\char45}{\char45}} & \texttt{{\char45}{\char45}{\char45}} & \texttt{{\char45}{\char45}{\char45}} & \texttt{{\char45}{\char45}{\char45}} & \texttt{{\char45}{\char45}{\char45}} & -- \\
Pepper $N$ Times & \texttt{000} & \texttt{000} & \texttt{000} & \texttt{000} & \texttt{000} & 0.00 \\
Lift Bowl $N$ Times & \texttt{000} & \texttt{000} & \texttt{000} & \texttt{000} & \texttt{000} & 0.00 \\
Press Button $N$ Times & \texttt{000} & \texttt{000} & \texttt{000} & \texttt{000} & \texttt{000} & 0.00 \\
$N$ Blocks into Box & \texttt{010} & \texttt{000} & \texttt{000} & \texttt{000} & \texttt{000} & 0.07 \\
\midrule
\textbf{Overall} & \multicolumn{5}{c}{\scriptsize Run 1/2/3: 0.03 / 0.05 / 0.05} & \textbf{0.04} \\
\bottomrule
\end{tabular}%
}
\end{subtable}
\end{table}

\begin{table}[htbp]
\centering
\footnotesize
\caption{X-VLA --- scene-level binary outcomes in the original environment.}
\label{tab:xvla}
\begin{subtable}[t]{0.495\linewidth}
\centering
\caption{ID, original env.\ (10 tasks)}
\label{tab:xvla_id}
\resizebox{\linewidth}{!}{%
\begin{tabular}{lcccccc}
\toprule
\textbf{Task} & \textbf{\#1} & \textbf{\#2} & \textbf{\#3} & \textbf{\#4} & \textbf{\#5} & \textbf{Avg.} \\
\midrule
Bread on Plate & \texttt{111} & \texttt{100} & \texttt{000} & \texttt{000} & \texttt{000} & 0.27 \\
Bowl on Coaster & \texttt{101} & \texttt{000} & \texttt{000} & \texttt{000} & \texttt{000} & 0.13 \\
Stack Block & \texttt{000} & \texttt{000} & \texttt{000} & \texttt{000} & \texttt{000} & 0.00 \\
Fold Towel & \texttt{111} & \texttt{001} & \texttt{101} & \texttt{110} & \texttt{000} & 0.53 \\
Open Oven & \texttt{010} & \texttt{000} & \texttt{000} & \texttt{000} & \texttt{000} & 0.07 \\
Erase Whiteboard & \texttt{000} & \texttt{000} & \texttt{000} & \texttt{000} & \texttt{000} & 0.00 \\
Pepper $N$ Times & \texttt{000} & \texttt{110} & \texttt{000} & \texttt{010} & \texttt{000} & 0.20 \\
Lift Bowl $N$ Times & \texttt{000} & \texttt{010} & \texttt{000} & \texttt{000} & \texttt{000} & 0.07 \\
Press Button $N$ Times & \texttt{000} & \texttt{000} & \texttt{000} & \texttt{000} & \texttt{000} & 0.00 \\
$N$ Blocks into Box & \texttt{000} & \texttt{000} & \texttt{000} & \texttt{000} & \texttt{000} & 0.00 \\
\midrule
\textbf{Overall} & \multicolumn{5}{c}{\scriptsize Run 1/2/3: 0.14 / 0.14 / 0.10} & \textbf{0.13} \\
\bottomrule
\end{tabular}%
}
\end{subtable}
\hfill
\begin{subtable}[t]{0.495\linewidth}
\centering
\caption{OOD, original env.\ (8 tasks)}
\label{tab:xvla_ood}
\resizebox{\linewidth}{!}{%
\begin{tabular}{lcccccc}
\toprule
\textbf{Task} & \textbf{\#1} & \textbf{\#2} & \textbf{\#3} & \textbf{\#4} & \textbf{\#5} & \textbf{Avg.} \\
\midrule
Bread on Plate & \texttt{110} & \texttt{110} & \texttt{100} & \texttt{000} & \texttt{000} & 0.33 \\
Bowl on Coaster & \texttt{000} & \texttt{000} & \texttt{000} & \texttt{000} & \texttt{000} & 0.00 \\
Stack Block & \texttt{000} & \texttt{000} & \texttt{000} & \texttt{000} & \texttt{000} & 0.00 \\
Fold Towel & \texttt{000} & \texttt{000} & \texttt{000} & \texttt{010} & \texttt{000} & 0.07 \\
Open Oven & \texttt{{\char45}{\char45}{\char45}} & \texttt{{\char45}{\char45}{\char45}} & \texttt{{\char45}{\char45}{\char45}} & \texttt{{\char45}{\char45}{\char45}} & \texttt{{\char45}{\char45}{\char45}} & -- \\
Erase Whiteboard & \texttt{{\char45}{\char45}{\char45}} & \texttt{{\char45}{\char45}{\char45}} & \texttt{{\char45}{\char45}{\char45}} & \texttt{{\char45}{\char45}{\char45}} & \texttt{{\char45}{\char45}{\char45}} & -- \\
Pepper $N$ Times & \texttt{000} & \texttt{000} & \texttt{000} & \texttt{000} & \texttt{000} & 0.00 \\
Lift Bowl $N$ Times & \texttt{000} & \texttt{000} & \texttt{000} & \texttt{000} & \texttt{000} & 0.00 \\
Press Button $N$ Times & \texttt{000} & \texttt{000} & \texttt{000} & \texttt{000} & \texttt{000} & 0.00 \\
$N$ Blocks into Box & \texttt{000} & \texttt{000} & \texttt{000} & \texttt{000} & \texttt{000} & 0.00 \\
\midrule
\textbf{Overall} & \multicolumn{5}{c}{\scriptsize Run 1/2/3: 0.07 / 0.07 / 0.00} & \textbf{0.05} \\
\bottomrule
\end{tabular}%
}
\end{subtable}
\end{table}

\begin{table}[htbp]
\centering
\footnotesize
\caption{MolmoAct2\_SO100\_101 --- scene-level binary outcomes
in the original environment.}
\label{tab:molmoact2}

\begin{subtable}[t]{0.495\linewidth}
\centering
\caption{ID, original env.\ (10 tasks)}
\label{tab:molmoact2_id_orig}
\resizebox{\linewidth}{!}{%
\begin{tabular}{lcccccc}
\toprule
\textbf{Task} & \textbf{\#1} & \textbf{\#2} &
\textbf{\#3} & \textbf{\#4} & \textbf{\#5} &
\textbf{Avg.} \\
\midrule
Bread on Plate & \texttt{111} & \texttt{111} & \texttt{101} & \texttt{111} & \texttt{111} & 0.93 \\
Bowl on Coaster & \texttt{111} & \texttt{001} & \texttt{111} & \texttt{101} & \texttt{110} & 0.73 \\
Stack Block & \texttt{010} & \texttt{101} & \texttt{100} & \texttt{000} & \texttt{000} & 0.27 \\
Fold Towel & \texttt{111} & \texttt{111} & \texttt{010} & \texttt{111} & \texttt{001} & 0.73 \\
Open Oven & \texttt{000} & \texttt{000} & \texttt{100} & \texttt{000} & \texttt{100} & 0.13 \\
Erase Whiteboard & \texttt{111} & \texttt{110} & \texttt{111} & \texttt{000} & \texttt{000} & 0.53 \\
Pepper $N$ Times & \texttt{000} & \texttt{010} & \texttt{000} & \texttt{100} & \texttt{000} & 0.13 \\
Lift Bowl $N$ Times & \texttt{100} & \texttt{000} & \texttt{000} & \texttt{001} & \texttt{000} & 0.13 \\
Press Button $N$ Times & \texttt{000} & \texttt{000} & \texttt{000} & \texttt{000} & \texttt{000} & 0.00 \\
$N$ Blocks into Box & \texttt{101} & \texttt{100} & \texttt{010} & \texttt{000} & \texttt{000} & 0.27 \\
\midrule
\textbf{Overall} &
\multicolumn{5}{c}{\scriptsize Run 1/2/3: 0.460 / 0.340 / 0.360} &
\textbf{0.39} \\
\bottomrule
\end{tabular}%
}
\end{subtable}
\hfill
\begin{subtable}[t]{0.495\linewidth}
\centering
\caption{OOD, original env.\ (8 tasks)}
\label{tab:molmoact2_ood_orig}
\resizebox{\linewidth}{!}{%
\begin{tabular}{lcccccc}
\toprule
\textbf{Task} & \textbf{\#1} & \textbf{\#2} &
\textbf{\#3} & \textbf{\#4} & \textbf{\#5} &
\textbf{Avg.} \\
\midrule
Bread on Plate & \texttt{111} & \texttt{111} & \texttt{111} & \texttt{010} & \texttt{111} & 0.87 \\
Bowl on Coaster & \texttt{111} & \texttt{111} & \texttt{111} & \texttt{111} & \texttt{100} & 0.87 \\
Stack Block & \texttt{110} & \texttt{001} & \texttt{000} & \texttt{111} & \texttt{111} & 0.60 \\
Fold Towel & \texttt{111} & \texttt{111} & \texttt{111} & \texttt{010} & \texttt{111} & 0.87 \\
Open Oven & \texttt{---} & \texttt{---} & \texttt{---} & \texttt{---} & \texttt{---} & -- \\
Erase Whiteboard & \texttt{---} & \texttt{---} & \texttt{---} & \texttt{---} & \texttt{---} & -- \\
Pepper $N$ Times & \texttt{001} & \texttt{000} & \texttt{010} & \texttt{000} & \texttt{000} & 0.13 \\
Lift Bowl $N$ Times & \texttt{000} & \texttt{000} & \texttt{100} & \texttt{001} & \texttt{000} & 0.13 \\
Press Button $N$ Times & \texttt{000} & \texttt{000} & \texttt{000} & \texttt{010} & \texttt{000} & 0.07 \\
$N$ Blocks into Box & \texttt{110} & \texttt{010} & \texttt{000} & \texttt{000} & \texttt{000} & 0.20 \\
\midrule
\textbf{Overall} &
\multicolumn{5}{c}{\scriptsize Run 1/2/3: 0.450 / 0.525 / 0.425} &
\textbf{0.47} \\
\bottomrule
\end{tabular}%
}
\end{subtable}
\end{table}

\begin{table}[htbp]
\centering
\footnotesize
\caption{GR00T N1.7 --- scene-level binary outcomes in the original environment.}
\label{tab:gr00t}
\begin{subtable}[t]{0.495\linewidth}
\centering
\caption{ID, original env.\ (10 tasks)}
\label{tab:gr00t_id_orig}
\resizebox{\linewidth}{!}{%
\begin{tabular}{lcccccc}
\toprule
\textbf{Task} & \textbf{\#1} & \textbf{\#2} & \textbf{\#3} & \textbf{\#4} & \textbf{\#5} & \textbf{Avg.} \\
\midrule
Bread on Plate & \texttt{111} & \texttt{111} & \texttt{000} & \texttt{111} & \texttt{111} & 0.80 \\
Bowl on Coaster & \texttt{111} & \texttt{101} & \texttt{111} & \texttt{111} & \texttt{011} & 0.87 \\
Stack Block & \texttt{000} & \texttt{000} & \texttt{000} & \texttt{000} & \texttt{000} & 0.00 \\
Fold Towel & \texttt{111} & \texttt{111} & \texttt{111} & \texttt{111} & \texttt{111} & 1.00 \\
Open Oven & \texttt{001} & \texttt{000} & \texttt{011} & \texttt{101} & \texttt{011} & 0.47 \\
Erase Whiteboard & \texttt{011} & \texttt{110} & \texttt{011} & \texttt{000} & \texttt{000} & 0.40 \\
Pepper $N$ Times & \texttt{111} & \texttt{000} & \texttt{000} & \texttt{000} & \texttt{000} & 0.20 \\
Lift Bowl $N$ Times & \texttt{000} & \texttt{000} & \texttt{000} & \texttt{010} & \texttt{000} & 0.07 \\
Press Button $N$ Times & \texttt{101} & \texttt{000} & \texttt{000} & \texttt{000} & \texttt{000} & 0.13 \\
$N$ Blocks into Box & \texttt{011} & \texttt{000} & \texttt{000} & \texttt{100} & \texttt{000} & 0.29\\
\midrule
\textbf{Overall} & \multicolumn{5}{c}{\scriptsize Run 1/2/3: 0.36 / 0.42 / 0.46} & \textbf{0.41} \\
\bottomrule
\end{tabular}%
}
\end{subtable}
\hfill
\begin{subtable}[t]{0.495\linewidth}
\centering
\caption{OOD, original env.\ (8 tasks)}
\label{tab:gr00t_ood_orig}
\resizebox{\linewidth}{!}{%
\begin{tabular}{lcccccc}
\toprule
\textbf{Task} & \textbf{\#1} & \textbf{\#2} & \textbf{\#3} & \textbf{\#4} & \textbf{\#5} & \textbf{Avg.} \\
\midrule
Bread on Plate & \texttt{111} & \texttt{111} & \texttt{111} & \texttt{111} & \texttt{011} & 0.93 \\
Bowl on Coaster & \texttt{111} & \texttt{111} & \texttt{100} & \texttt{111} & \texttt{000} & 0.67 \\
Stack Block & \texttt{000} & \texttt{000} & \texttt{000} & \texttt{000} & \texttt{000} & 0.00 \\
Fold Towel & \texttt{111} & \texttt{111} & \texttt{111} & \texttt{111} & \texttt{111} & 1.00 \\
Open Oven & \texttt{{\char45}{\char45}{\char45}} & \texttt{{\char45}{\char45}{\char45}} & \texttt{{\char45}{\char45}{\char45}} & \texttt{{\char45}{\char45}{\char45}} & \texttt{{\char45}{\char45}{\char45}} & -- \\
Erase Whiteboard & \texttt{{\char45}{\char45}{\char45}} & \texttt{{\char45}{\char45}{\char45}} & \texttt{{\char45}{\char45}{\char45}} & \texttt{{\char45}{\char45}{\char45}} & \texttt{{\char45}{\char45}{\char45}} & -- \\
Pepper $N$ Times & \texttt{000} & \texttt{000} & \texttt{000} & \texttt{000} & \texttt{000} & 0.00 \\
Lift Bowl $N$ Times & \texttt{000} & \texttt{000} & \texttt{100} & \texttt{000} & \texttt{000} & 0.07 \\
Press Button $N$ Times & \texttt{000} & \texttt{000} & \texttt{000} & \texttt{000} & \texttt{000} & 0.00 \\
$N$ Blocks into Box & \texttt{011} & \texttt{000} & \texttt{000} & \texttt{000} & \texttt{000} & 0.13 \\
\midrule
\textbf{Overall} & \multicolumn{5}{c}{\scriptsize Run 1/2/3: 0.35 / 0.35 / 0.35} & \textbf{0.35} \\
\bottomrule
\end{tabular}%
}
\end{subtable}
\end{table}

\pagebreak

\section{License}
\label{app:license}
Our code, checkpoints, and dataset are released under the Creative Commons Attribution 4.0 International (CC BY 4.0) license, which permits use, sharing, adaptation, and redistribution with appropriate attribution.

%% file: checklist.tex
\section*{NeurIPS Paper Checklist}

\begin{enumerate}

\item {\bf Claims}
    \item[] Question: Do the main claims made in the abstract and introduction accurately reflect the paper's contributions and scope?
    \item[] Answer: \answerYes{}
    \item[] Justification: Our claims in the abstract accurately reflect our contributions, which includes a novel, low-cost, reproducible evaluation benchmark for imitation learning and Vision-Language-Action (VLA) models, and we provide experiments for policy evaluations with the benchmark. 

\item {\bf Limitations}
    \item[] Question: Does the paper discuss the limitations of the work performed by the authors?
    \item[] Answer: \answerYes{} 
    \item[] Justification: Yes, we mentioned in the \textbf{Conclusion \& Limitation} Section.

\item {\bf Theory assumptions and proofs}
    \item[] Question: For each theoretical result, does the paper provide the full set of assumptions and a complete (and correct) proof?
    \item[] Answer: \answerYes{} 
    \item[] Justification: Yes, we provide mathematical proof to show the actions predicted by the policy can be transfer to a new SO101 arm. We also provide experiments to prove that a new setting won't affect evaluation performance. 

    \item {\bf Experimental result reproducibility}
    \item[] Question: Does the paper fully disclose all the information needed to reproduce the main experimental results of the paper to the extent that it affects the main claims and/or conclusions of the paper (regardless of whether the code and data are provided or not)?
    \item[] Answer: \answerYes{} 
    \item[] Justification: We build a reproducible platform, so everything can be reproduced. We also report our model training parameters. Following the platform setup and training details should able to reproduce the results. 

\item {\bf Open access to data and code}
    \item[] Question: Does the paper provide open access to the data and code, with sufficient instructions to faithfully reproduce the main experimental results, as described in supplemental material?
    \item[] Answer: \answerYes{} 
    \item[] Justification: Yes we provided instructions for benchmark construction and benchmark evaluation in the paper and the appendix.

\item {\bf Experimental setting/details}
    \item[] Question: Does the paper specify all the training and test details (e.g., data splits, hyperparameters, how they were chosen, type of optimizer) necessary to understand the results?
    \item[] Answer: \answerYes{} 
    \item[] Justification: Yes we provide our training details and test setup in the main paper and appendix. 

\item {\bf Experiment statistical significance}
    \item[] Question: Does the paper report error bars suitably and correctly defined or other appropriate information about the statistical significance of the experiments?
    \item[] Answer: \answerYes{} 
    \item[] Justification: Tables~\ref{tab:ID_success_rate} and~\ref{tab:OOD_success_rate} report means and sample standard deviations across three evaluation rounds. Section~\ref{sec:algorithm} defines both task-level and overall aggregation and states the source of variability. The separate single-round reproducibility study is explicitly identified; no training-seed uncertainty or statistical significance is claimed. 

\item {\bf Experiments compute resources}
    \item[] Question: For each experiment, does the paper provide sufficient information on the computer resources (type of compute workers, memory, time of execution) needed to reproduce the experiments?
    \item[] Answer: \answerYes{} 
    \item[] Justification: Yes we provide training details in the appendix.
    
\item {\bf Code of ethics}
    \item[] Question: Does the research conducted in the paper conform, in every respect, with the NeurIPS Code of Ethics \url{https://neurips.cc/public/EthicsGuidelines}?
    \item[] Answer: \answerYes{} 
    \item[] Justification: Yes we followed Neurips Code of Ethics.

\item {\bf Broader impacts}
    \item[] Question: Does the paper discuss both potential positive societal impacts and negative societal impacts of the work performed?
    \item[] Answer: \answerYes{} 
    \item[] Justification: Yes we talked about the social impacts in the introductions and conclusion and limitation section. 
    
\item {\bf Safeguards}
    \item[] Question: Does the paper describe safeguards that have been put in place for responsible release of data or models that have a high risk for misuse (e.g., pre-trained language models, image generators, or scraped datasets)?
    \item[] Answer: \answerNA{} 
    \item[] Justification: We only include a human-collected (teleoperated) robot dataset, which includes around 500 trajectories. 

\item {\bf Licenses for existing assets}
    \item[] Question: Are the creators or original owners of assets (e.g., code, data, models), used in the paper, properly credited and are the license and terms of use explicitly mentioned and properly respected?
    \item[] Answer: \answerYes{}
    \item[] Justification: We credits the creators and original owners of all used assets and explicitly states the licenses and terms of use.

\item {\bf New assets}
    \item[] Question: Are new assets introduced in the paper well documented and is the documentation provided alongside the assets?
    \item[] Answer: \answerYes{} 
    \item[] Justification: Details regarding the data collection protocol, benchmark set up licensing and limitations are provided in Appendix. Additional documentation, including the dataset structure and usage guideline, is available on the project page linked in paper.

\item {\bf Crowdsourcing and research with human subjects}
    \item[] Question: For crowdsourcing experiments and research with human subjects, does the paper include the full text of instructions given to participants and screenshots, if applicable, as well as details about compensation (if any)? 
    \item[] Answer: \answerNA{} 
    \item[] Justification: The paper does not involve crowdsourcing experiments or human-subjects research. Human involvement was limited to teleoperating the robot to collect task demonstrations in a controlled laboratory setup. No personal, behavioral, or sensitive human-subject data were collected.

\item {\bf Institutional review board (IRB) approvals or equivalent for research with human subjects}
    \item[] Question: Does the paper describe potential risks incurred by study participants, whether such risks were disclosed to the subjects, and whether Institutional Review Board (IRB) approvals (or an equivalent approval/review based on the requirements of your country or institution) were obtained?
    \item[] Answer: \answerNA{} 
    \item[] Justification: The paper does not involve human-subjects research. Human involvement was limited to teleoperating the robot for demonstration collection, and no personal or sensitive human-subject data were collected. Therefore, no need for IRB. 

\item {\bf Declaration of LLM usage}
    \item[] Question: Does the paper describe the usage of LLMs if it is an important, original, or non-standard component of the core methods in this research? Note that if the LLM is used only for writing, editing, or formatting purposes and does \emph{not} impact the core methodology, scientific rigor, or originality of the research, declaration is not required.
    \item[] Answer: \answerNA{} 
    \item[] Justification: The core idea, methodology, and protocol were designed and developed by the authors. LLMs were not involved in the original intellectual contributions or key technical components of the project.

\end{enumerate}